\documentclass{article}

\usepackage{fr}
\usepackage{graphicx}

\usepackage[]{apacite}
\usepackage{setspace}
\usepackage[labelfont=bf]{caption}
\usepackage{listings}
\usepackage{xcolor}
\usepackage[noabbrev,capitalize]{cleveref}
\usepackage[title]{appendix}
\usepackage{tabularx}
\usepackage[export]{adjustbox}
\usepackage{booktabs}
\usepackage{multirow}
\usepackage{placeins}
\usepackage{bytefield} \usepackage{lscape}
\usepackage{url}
\usepackage{textcomp}
\usepackage{array}
\usepackage{float}
\usepackage{environ}
\usepackage[binary-units=true]{siunitx}
\usepackage{csquotes}

\definecolor{codegreen}{rgb}{0,0.6,0}
\definecolor{codegray}{rgb}{0.5,0.5,0.5}
\definecolor{codepurple}{rgb}{0.58,0,0.82}
\definecolor{backcolour}{rgb}{0.95,0.95,0.92}

\lstdefinestyle{mystyle}{
        commentstyle=\color{codegreen},
    keywordstyle=\color{blue},
    numberstyle=\tiny\color{codegray},
    stringstyle=\color{codepurple},
    basicstyle=\ttfamily\footnotesize,
    breakatwhitespace=false,         
    breaklines=true,                 
    captionpos=b,                    
    keepspaces=true,                 
    numbers=left,                    
    numbersep=5pt,                  
    showspaces=false,                
    showstringspaces=false,
    showtabs=false,                  
    tabsize=2
}

\title{From Warfighting Needs to Robot Actuation: \\A Complete Rapid Integration Swarming Solution}

\author{
	Eugene M. Taranta II \\
		Northrop Grumman Corporation \\
	Mission Systems \\
	Orlando, FL 32817 \\
	\texttt{eugene.taranta@ngc.com} \\
	\And
	Adam Seiwert \\
		Northrop Grumman Corporation \\
	Mission Systems \\
	Aurora, CO 80017 \\
	\texttt{adam.seiwert@ngc.com} \\
\And
	Anthony Goeckner \\
		Northrop Grumman Corporation \\
	Mission Systems \\
	Annapolis, MD 21401 \\
	\texttt{anthony.goeckner@ngc.com} \\
	\And
	Khiem Nguyen \\
		Northrop Grumman Corporation \\
	Mission Systems \\
	Aurora, CO 80017 \\
	\texttt{khiem.d.nguyen@ngc.com} \\
\And
	Erin Cherry \\
		Northrop Grumman Corporation \\
	Mission Systems \\
	McLean, VA 22102 \\
	\texttt{erin.cherry@ngc.com} \\
}

\begin{document}

\maketitle

\begin{abstract}
Swarm robotics systems have the potential to transform warfighting in urban environments, but until now have not seen large-scale field testing. We present the Rapid Integration Swarming Ecosystem (RISE), a platform for future multi-agent research and deployment. RISE enables rapid integration of third-party swarm tactics and behaviors, which was demonstrated using both physical and simulated swarms. Our physical testbed is composed of more than 250 networked heterogeneous agents and has been extensively tested in mock warfare scenarios at five urban combat training ranges. RISE implements live, virtual, constructive simulation capabilities to allow the use of both virtual and physical agents simultaneously, while our ``fluid fidelity'' simulation enables adaptive scaling between low and high fidelity simulation levels based on dynamic runtime requirements. Both virtual and physical agents are controlled with a unified gesture-based interface that enables a greater than 150:1 agent-to-operator ratio. Through this interface, we enable efficient swarm-based mission execution. RISE translates mission needs to robot actuation with rapid tactic integration, a reliable testbed, and efficient operation.
\end{abstract}

\begin{figure}
    \centering
    \includegraphics[width=1\textwidth]{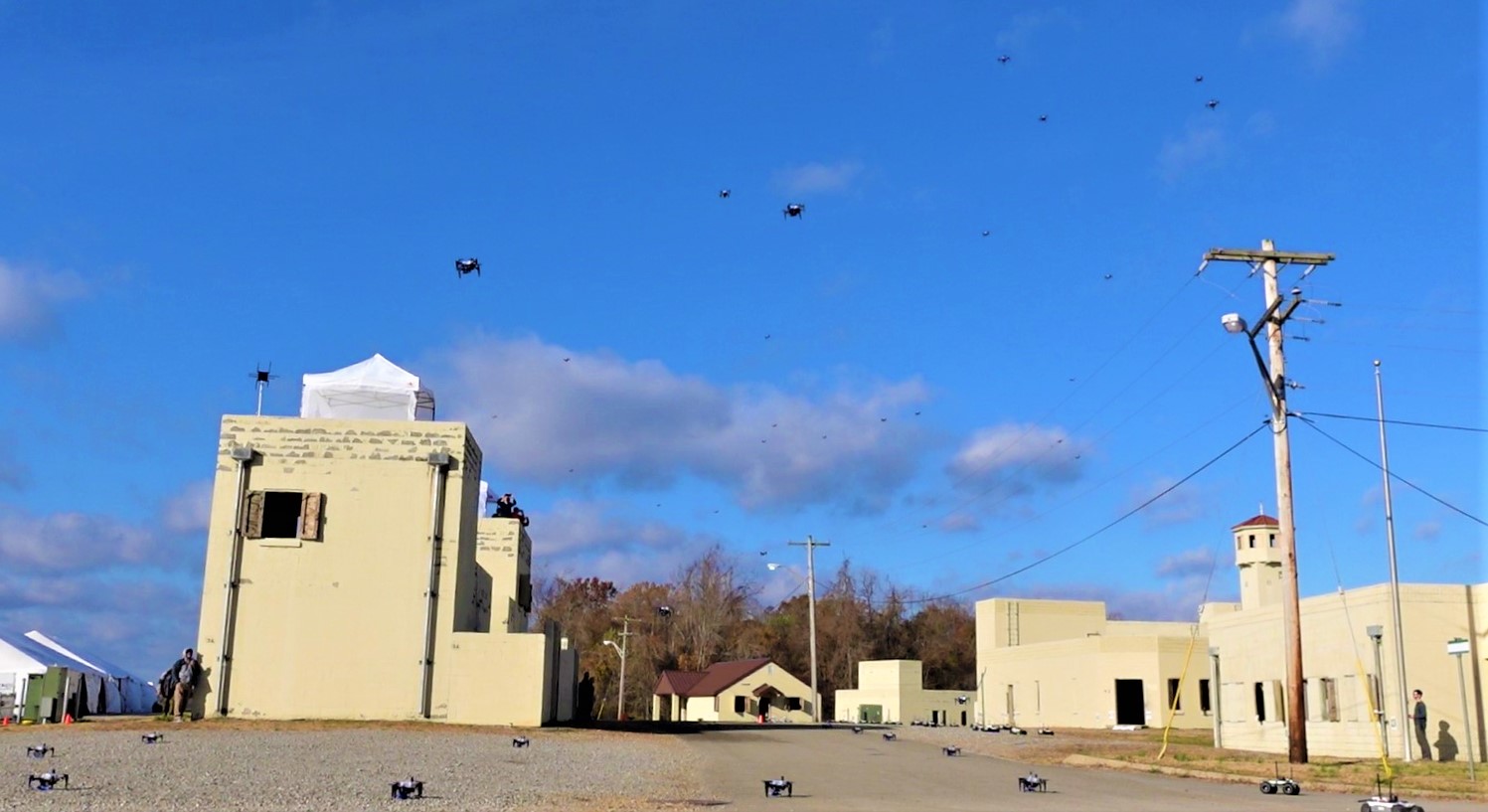}
    \captionof{figure}{Large-scale unmanned aerial vehicle (UAV) launch for field experiment FX-6 at Ft. Campbell, TN.}
    \label{fig:fx6_masslaunch}
\end{figure}

\section{Introduction}

Coordinated, collaborative robotic swarms have the potential to offer transformational societal impacts. For example, swarms are expected to significantly improve disaster response activities such as search and rescue and damage assessment, in addition to providing a communication network and even delivering supplies. Swarming may also benefit industries like agriculture and farming through autonomous operations that reduce cost, save time, and improve precision~\cite{farming}. In our work, we focus on the military application of large-scale swarms through the Defense Advanced Research Projects Agency (DARPA) OFFensive Swarm-Enabled Tactics (OFFSET) program. The employment of a robotic swarm increases a military unit’s span of control from 1:1 (Soldier:Platform) to 1:150+. In this way, a swarm may increase mission effectiveness by several orders of magnitude. Swarming is considered a non-linear dispersed (NLD) operation with a RAND report by~\citeA{edwards2005swarming} positing that swarming is the most aggressive form of NLD and that it will play a central role in future military operations when heavy forces are unavailable. 

Advances in low-cost robotics hardware, network solutions, sensing algorithms, and artificial intelligence help bring practical large scale robot swarms closer to reality~\cite{chung_live-fly_2016}. These advancements are typically fragmented across unique and isolated problem domains. However, robot swarming is interdisciplinary, spanning several interrelated fields. Autonomy, distributed perception, logistics, mobile ad hoc networking, and human-swarm teaming ~\cite{chung_live-fly_2016, edwards2000swarming} are example domains where advances in one area may have positive compounding effects in another. Consequently, to understand the impact of new capabilities on swarm mission outcomes, one must integrate and evaluate their technologies into a swarming ecosystem. This in itself is problematic as few holistic robotic swarming systems exist for large scale swarm research, development, and deployment, especially those that have been fielded and proven to operate with a sufficient technical readiness. Our technology, the Rapid Integration Swarm Ecosystem (RISE), is one such system that addresses this problem. 

\begin{figure}[t]
    \centering
    \includegraphics[width=.7\textwidth]{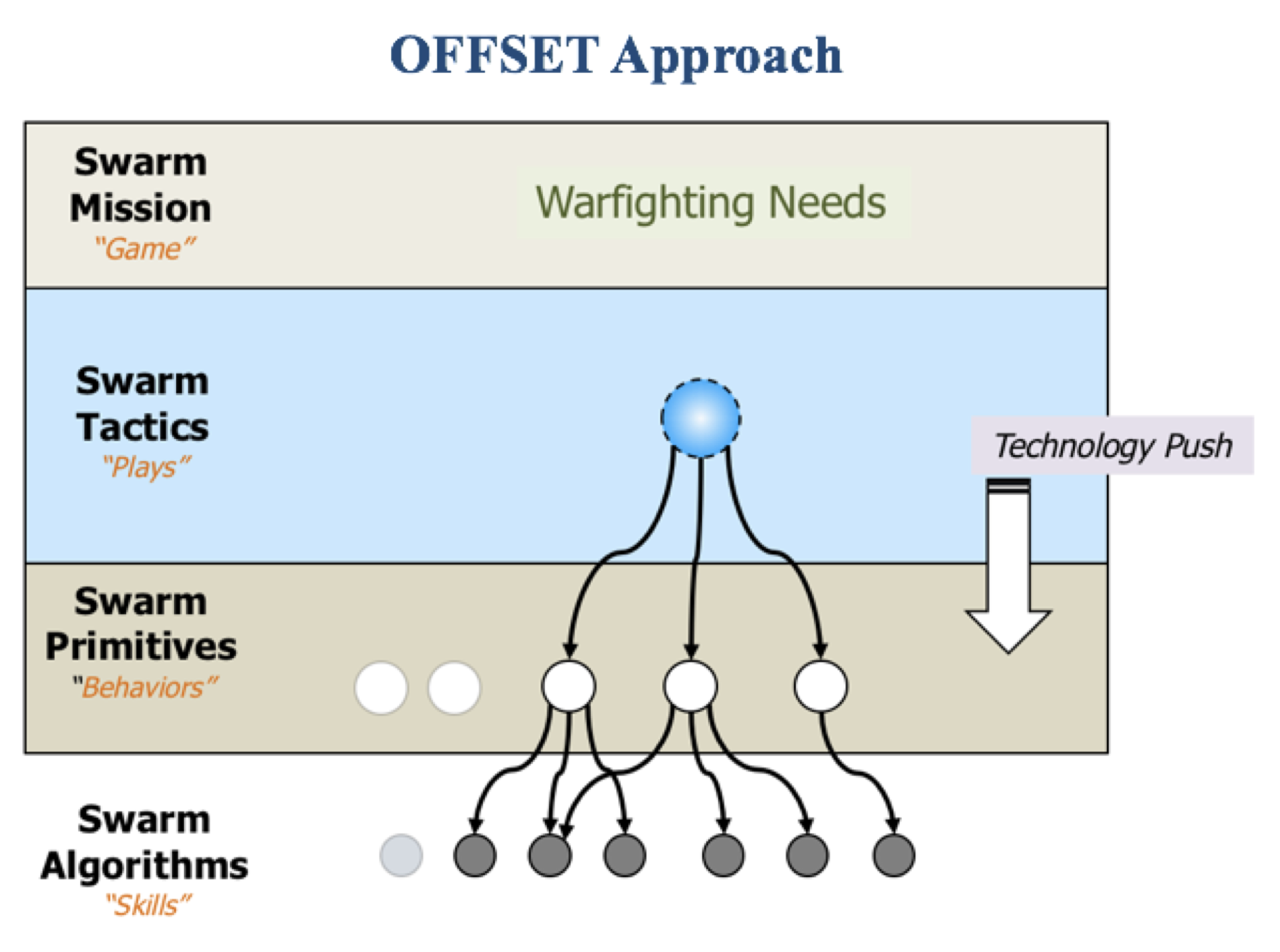}
    \captionof{figure}{The DARPA OFFSET view on how warfighting needs may be hierarchically decomposed into algorithms that interact with robot hardware. A human operator translates mission objectives into tactical swarm maneuvers that in turn decompose into robot primitives utilizing algorithms. Each layer is present in RISE via a unique software architecture designed to facilitate rapid development and integration. Image adopted from \cite{TacticTree}.}
    \label{fig:tactic_tree}
\end{figure}

More importantly, we address the question: can a multi-robot ecosystem designed for rapid integration and one-to-many control can be fielded, where a single operator commands 150+ autonomous vehicles in tactical maneuvers? Large scale swarms have yet to be deployed outside of simulation, and it remains unclear what software and hardware architecture can support single operator command at this scale. Earlier efforts in this direction have been made by \cite{chung_live-fly_2016} and \cite{clark_ccast_2021}, though they are limited in either scope or scale. For the first time, we demonstrate with RISE that large scale swarm control for offensive tactical operations is possible. Further, in this work we discuss our architecture design choices and the requirements that informed our design, which practitioners can use to guide their own work. 

In general, a swarming ecosystem requires a specialized command and control (C2) interface for swarm control. When we designed RISE, it was our goal that our C2 system would enable a single operator to control hundreds of platforms by specifying only high level commands, and enable the operator to build tactics without having to program robot behaviors. To this end, we were inspired by playbook frameworks and decided to leverage the hierarchical swarm framework of Tactics, Primitives, and Algorithms that were established for the DARPA OFFSET program shown in \cref{fig:tactic_tree}. This framework is similar to the work of \citeA{masc}, and both drew inspiration from swarm experiments at the Naval Postgraduate School.

The main contribution of this work is RISE\footnote{RISE software is maintained by Northrop Grumman Corporation and the US Government. Contact Erin Cherry (erin.cherry@ngc.com) for more information on procedures for gaining access.}, an end-to-end robot swarming ecosystem that enables rapid development, research, testing, and deployment of large scale heterogeneous robot swarm technologies. The RISE contribution further comprises:
\begin{enumerate}
    \item A human-swarm team interface that enables a single operator to command large scale swarms,
    \item A platform for swarm tactic development and execution that bridges the operator to robots and enables practitioners unfamiliar with robotics software to author tactics, and
        \item A lightweight, mobile ad hoc network solution that enables reliable communication at scale between hundreds of agents and other devices. 
\end{enumerate} 

In addition to the RISE software, we customized, configured, and maintained a swarm testbed of over 250 physical, small, unmanned air and ground vehicles. We demonstrated RISE on this swarm testbed at five military ranges over the last four years. The most recent event in November 2021 was the culminating OFFSET swarm field experiment, and we demonstrated one user controlling 174 platforms (84 air and 90 ground).

\section{Related Work}

In this section, we discuss existing work in several fields relevant to RISE. Where possible, we describe the major papers in those fields. Throughout this section, we also refer the curious reader to existing surveys for further information. We also note that swarms consist of many intelligent ``agents,'' individual entities which are capable of autonomous sensing, decision-making, and actuation \cite{van_der_hoek_multi-agent_2008}. We use the words ``agent'' or ``swarm agent'' to refer to the robotic hardware platform together with the software which runs upon it. Further, we sometimes make a distinction between \textit{simulated agents} and \textit{physical agents}, where the former uses simulated hardware and the latter uses real hardware.

\subsection{Multi-Robot Systems}
In the literature, ``swarm robotics'' typically refers to homogeneous systems that have a biological inspiration and which communicate using pheromones or other non-switched networking methods. While RISE is a ``swarming ecosystem,'' it is also more accurately a ``multi-robot system'' (MRS) or more generally a ``multi-agent system'' (MAS). This distinction is highlighted by \cite{rizk_cooperative_2019}, who perform an extensive and high-quality survey of multi-agent and multi-robot systems. Additional surveys of interest include \citeA{dorigo_swarm_2021}, which focuses on homogeneous swarm robotics, and \citeA{schranz_swarm_2020}, which provides a high-level overview of swarm and multi-robot projects. Rather than reiterate the results of these surveys here, we will only highlight especially relevant works. Note that while we refer to RISE as both a ``swarm robotics'' and a ``multi-robot'' system in this paper, we limit our discussion to that of multi-robot systems, since they are most similar to RISE.

In particular, we first focus on multi-robot systems which function as testbeds for further research. A survey of multi-robot testbeds is given by \citeA{jimenez-gonzalez_testbeds_2013}, although several advances have occurred since its publication. Significant early testbeds included the Mobile Emulab \cite{johnson_mobile_2006}, HoTDeC \cite{stubbs_multivehicle_2006}, and RAVEN \cite{how_real-time_2008}. The Mobile Emulab and HoTDeC were characterized by homogeneous robotic platforms and constrained operating environments, only capable of operation in a lab. Both are designed to provide access to remote users who may control the systems' operations via the internet. In both testbeds, robots are extremely simple with limited localization or sensing capabilities of their own, instead relying on centralized camera systems. Importantly, neither study develops abstract swarm concepts such as tactics or primitives, instead providing only a basic computing platform on each agent without a swarm software framework.

Of these early multi-robot testbeds, perhaps the most similar to RISE is RAVEN, a heterogeneous air and ground swarm testbed designed for experimentation with a variety of multi-robot scenarios. While the software architecture is similar to that of RISE, RAVEN is limited to ten vehicles and may only be used in a lab with some external sensing apparatus. Additionally, RAVEN's C2 paradigm is mostly focused on direct control of individual vehicles, whereas RISE focuses on command of the swarm as a whole to maximize the agent-to-operator ratio.

In \citeA{pickem_gritsbot_2015} and later in \citeA{pickem_robotarium_2017}, the authors showcase a multi-robot testbed created by their GRITSBot platforms. More recently, the ``ARGroHBotS'' testbed was published by \citeA{ospina_argrohbots_2021} and ``Crazyswarm'' was published by \citeA{preiss_crazyswarm_2017}. Again, these testbeds are limited to lab use and have no or limited command and control functionality.

The Swarmanoid project developed a heterogeneous swarm or multi-robot system consisting of both unmanned aerial vehicles (UAVs) and unmanned ground vehicles (UGVs) \cite{dorigo_swarmanoid_2013}. As with other projects, the Swarmanoid authors do not clearly define a framework for swarm tactics and overall behavior, and do not clearly state how additional behaviors may be integrated with their system. However, they provide an advanced multi-robot simulator, ARGoS, which remains in use for multi-agent research to this day \cite{dorigo_swarmanoid_2013}.

The COMRADE multi-robot system is an advanced multi-robot system intended for clearing mines and explosive devices in conflict zones. COMRADE provides multi-robot search, data fusion, task allocation, and a basic C2 interface, and is designed for real-world use \cite{dasgupta_comrade_2015}. However, the authors do not expand on the concept of tactics or other abstractions for the design of new swarm behaviors; rather, the system is single-purpose. Additionally, the C2 system is extremely rudimentary, providing only the barest of controls to direct individual agents and apparently lacking the ability to manually command the swarm as a whole by issuing new tasks for allocation.

The RUTA heterogeneous swarm \cite{abukhalil_coordinating_2016} is developed primarily for research on dynamic task allocation. However, it takes a similar approach to robotic hardware abstraction as RISE, allowing for extreme heterogeneity. Some definition of the different layers of software execution for swarm behaviors is provided, with RUTA's ``actions'' being similar to RISE's agent-level ``primitives'' or ``algorithms'' \cite{abukhalil_coordinating_2016}. However, the swarm is small and not designed for scale with only five robots. Additionally, the system primarily focuses on task allocation and does not include any concepts related to C2.

More recently, \citeA{chamanbaz_swarm-enabling_2017} developed a hardware and software package to enable simple integration of robotic platforms into a multi-robot system. It consists of computing and networking hardware which can be integrated with existing platforms, along with a software library to enable creation of new behaviors using those platforms. This platform is tested on two separate swarm systems. However, the authors do not differentiate between different layers of the behavior stack (tactics, primitives, and algorithms) and do not consider any human-computer interaction factors. In fact, the authors make no mention of any type of user interface or control.

A significant multi-robot system is presented by \citeA{chung_live-fly_2016}, in which fifty UAVs are simultaneously controlled by a team of two. This work may be thought of as a direct predecessor of OFFSET with the same program manager, Dr. Timothy Chung, with similar goals of enabling multi-robot systems to perform complex coordinated actions while under the control of a minimal number of operators. This work is expanded upon by \citeA{gros_multi-swarm_2018}, in which two of the UAV swarms created by \citeA{chung_live-fly_2016} battle for air dominance. RISE expands upon this work with a far greater number of agents, innovative new C2 concepts, a software architecture designed for integration of new tactics and behaviors, the introduction of heterogeneity, and more.

Recently, a number of tools which ease the development of multi-robot systems have been created. One such tool is ROS2swarm, a hardware and package for the Robot Operating System 2 framework (ROS 2) that can be retrofitted to existing platforms. ROS2swarm provides multi-robot programming capabilities using a similar breakdown of swarm software to our tactics, primitives, and algorithms concept \cite{kaiser_ros2swarm_2022}. However, ROS2swarm does not enable dynamic transitions between tactics. In other words, the agents may only perform one tactic in a single execution run without a complete reboot of the robot software. ROS2swarm also does not provide any command and control capabilities, unlike RISE. Additionally, ROS2swarm currently uses a centralized network (access point-based) topology, while RISE uses a decentralized topology. While the ROS2swarm is tested with just six agents, we are quite certain that the ROS2swarm system is actually incapable of supporting large numbers of agents (150+), given its use of ROS 2's built-in networking mechanism. RISE replaces ROS 2's default networking mechanism with a custom solution for inter-agent communication. See \cref{sec:networking} for further details on RISE's networking solutions.

An alternate platform, SwarmUS, is provided by \citeA{villemure_swarmus_2022}, which includes both a hardware retrofit board and a software package, similar to \citeA{kaiser_ros2swarm_2022}. This provides necessary services for a multi-robot system such as communication, coordination, and localization. An Android-based C2 interface is also provided, but appears to be rudimentary and is not designed for control of a large-scale swarm by a single operator. In fact, SwarmUS was tested on no more than six robots, as opposed to RISE's hundreds \cite{villemure_swarmus_2022}. Importantly though, SwarmUS appears to focus primarily on hardware design, whereas RISE primarily focuses on software design. Thus, the two are difficult to compare further.

RISE is contemporaneous with ``CCAST'', developed by our counterparts on the DARPA OFFSET program \cite{clark_ccast_2021}. As CCAST and RISE solve the same problems and are designed for the same program and experiment scenarios, most of their capabilities are quite similar. However, RISE differs from CCAST in its use a decentralized mesh network for communication, as opposed to CCAST's access point approach. Additionally, RISE makes use of innovative common gesture commands for easy cross-platform use on laptops, touchscreen devices, and virtual \& augmented reality, whereas CCAST uses a point-and-click virtual reality interface. CCAST shares the behavior decomposition of tactics, primitives, and algorithms with RISE.

\subsection{Robotic Platforms}
Since RISE is a complex system of software which may be used with a variety of hardware platforms (see \cref{fig:robots}), we do not elaborate on prior hardware platforms used for swarm robotics or multi-agent research. Surveys on this area may be found in the literature, such as the one by \citeA{schranz_swarm_2020} which lists a number of platforms.

\subsection{Swarm Networking}
Swarm robotics systems in the literature may be broadly divided between those which use biologically-inspired ``pheromone'' systems, directional communications such as in \citeA{kornienkoInfraredComms}, and nondirectional radio-based networks. We will focus here on undirected digital radio-based networks, since we believe that the former methods are so different from our approach as to be nearly incomparable.

Three types of radio network architecture are commonly used: central access points, local wireless broadcasts, and mobile ad hoc networks (MANETs). In traditional wireless networks, agents communicate only with fixed infrastructure (known as an access point). Direct agent-to-agent communication is impossible. Instead, agents must first send data to the access point, which then retransmits the data to the destination agent. By contrast, a MANET allows both traditional agent-to-infrastructure communication and direct agent-to-agent communication \cite{hoebeke2004overview}.

A central access point approach is used by the CCAST system \cite{clark_ccast_2021}. The authors of that study found that agent connection to the access point became a major concern because agents were unable to communicate directly with their neighbors, thus increasing collision risk. For two neighboring agents to communicate, each needed an active connection to the access point, something not always possible in urban environments. This approach is also found in \citeA{stubbs_multivehicle_2006} and numerous other implementations.

Local wireless broadcast architectures are more common. While they often do not require a central access point, these architectures typically do not support relaying of data, distinguishing them from true ``MANETs'' by the definition in \citeA{hoebeke2004overview}. In \citeA{centibots}, an agent-to-agent local wireless broadcast is used. A similar system is employed by \citeA{rubensteinThousandBots}, where data is not relayed to agents beyond reception range of the transmitting agent, and only ``local'' communication with nearby neighbors is possible. Similar methods are implemented in \citeA{caprariAliceBot,dorigo_swarmanoid_2013}.

\citeA{cianciEPuck} implement a true MANET. Their approach makes use of the ZigBee mesh protocol. ZigBee is also used to implement a MANET by \citeA{fernandesZigbee}, \citeA{jevticRoboBees}, and \citeA{zahugiLibot}, among others. By far, the ZigBee system is the most popular MANET implementation in the literature. A MANET solution by Rajant Corporation is also used with success in an unmanned aerial vehicle swarm application by \citeA{engebraaten2019uav}, and in a mine-exploration swarm in \citeA{darpaSubT}.

\subsection{Simulation}
\label{sec:lit_sim}
As with any robotic development, simulation is a critical tool, but especially so with multi-agent and swarm systems. Simulations dealing with multiple agents vary widely in implementation, but even more so in their ability to translate from simulation to actual hardware. A large majority of swarm simulations typically consist of custom-built simulation environments to do evaluation on particular algorithms \cite{hamer2010simulation}. While this is appropriate for those particular evaluations, a setup of this nature would not suffice for a rapid integration swarming ecosystem such as RISE. Given this, we focused more on simulation environments that emphasize one-to-one translation to hardware, such as Gazebo \cite{Aguero-2015-VRC}. We leveraged lessons learned and best practices of simulations of this nature and then focused on making the necessary adjustments to maintain the real world translation while still reaching the necessary number of simulated agents (250+ per the OFFSET program). In \cref{sec:Simulation}, we go into more detail of this approach and additional acknowledgments of other relevant simulations.

\subsection{Human-Swarm Interface}
\label{sec:lit_hsi}
Human-swarm interaction specifically and human-robot interaction generally is an active area of research that includes interface design, control, communications, autonomy, and human factors such as situation awareness and cognition load, among others \cite{drew2021multi,chen2021human,kolling2015human,hocraffer2017meta}. However, in this subsection we focus on work that enables a one-to-many operator control over an offensive swarm, beginning with command complexity and ending with user interface customization.

Command complexity draws parallels with computational complexity in that total effort is bound by the number of decisions and actions an operator must take to complete a task \cite{lewis2006teamwork}. When robots work independently, effort grows linearly with the number of robots $n$ an operator must command, $\mathcal{O}(n)$. Conversely, tasks requiring careful coordination between robotics may result in superlinear complexity $\mathcal{O}(>n)$. Lewis provides by way of example the coordination of two unmanned vehicles pushing on the corners of a box, where the operator oscillates between each robot to move and straighten the box. 

Although command complexity enables practitioners to compare the bounded efficiency of different command strategies, fan-out is another useful measure proposed by \citeA{olsen2003metrics} that one uses to estimate how many homogeneous robots an operator can effectively operate at one time. Fan-out is defined mathematically as $(NF + IT) / (IT)$, where neglect time $NT$ is the expected duration of time a robot can be ignored without degrading beyond a minimum performance threshold, and interaction time $IT$ is the duration of time required to interact with a robot \cite{crandall2005validating}. Intuitively, the more a robot can do without requiring operator attention and the faster an operator can set up commands, the more total robots an operator can command without falling below a required level of performance. Prior work has shown an operator can command between eight and twelve robots simultaneously \cite{wang2009search}, depending on a variety of factors. This implies we must move away from direct robot control toward a higher level of abstraction in order to support large scale offensive swarming. 

One solution is to increase swarm autonomy \cite{mi2013human}. Command complexities moving toward order $\mathcal{O}(1)$ can be achieved in part through the use of planners \cite{lewis2013human}. Consider that without path planners and obstacle avoidance, it has been found that operators are unable to supervise more than a few UAVs \cite{cummings2007predicting}. Delegating certain tasks to an autonomy can further improve command complexity and fan-out. For example, \citeA{mclurkin2006speaking} designed a swarm interface inspired by the poplar real-time strategy games WarCraft and StarCraft. These games enable a player to control individual units, groups of units, or an entire army with simple high level commands such as mine, build, and attack. By distributing a single command to multiple robots, amortized interaction time can be significantly reduced and fan-out improved. Similarly, command complexity becomes sublinear when the operator can command multiple robots with only a single command when he or she is able to rely on the underlying automation. Another example called Playbook \cite{miller2000tasking, calhoun2018human} provides operators with a library of play templates that the operator can modify and which automation adapts to the situation. Playbook is analogous to a sports team’s playbook, where a leader selects a play that the team executes, and several studies have shown the benefits of templated plays.

We adopt a similar approach to these delegation strategies. In RISE, an operator controls the swarm by using mission level tactic commands that appropriate resources and coordinate the actions of multiple robots. An operator may combine tactics to form a course of action (COA) and modify the plan in real time as the situation changes. During mission planning with human warfighters, commanders communicate tactics through COA diagrams that include tactical control measures to establish responsibility and constrain operations that prevent units from interfering with each other \cite{united2015field}. This concept is easily extended to swarm command using a sketch and gesture based interface to simulate the natural use of pen and paper. \citeA{hammond2010sketch}, for example, developed a system for free-hand COA diagram sketch recognition with support for military symbology, achieving high accuracy on 485 symbols. 

One issue is that COA diagrams are drawn with standardized military symbology, such as those defined by MIL-STD 2525D \cite{army2020field}. However, RISE swarm tactics and control measures are constantly evolving and for some time will remain unstandardized. For this reason, practitioners who develop new tactics and control measures require rapid interface integration for development, test, and deployment. This can be achieved through gesture customization using rapid prototyping gesture recognizers \cite{jackknife}. In this way, we are able to expand our tactics library without requiring direct interface developer support or cluttering the interface, all while supporting a one-to-many low command complexity relationship.

\section{Background}

\subsection{On the Decomposition of Warfighting Needs into Robot Actuation}

We illustrate the hierarchical relationship between warfighting needs and robotics algorithms as envisioned by DARPA's OFFSET program in \cref{fig:tactic_tree}. This decomposition leads to a multilevel classification system that maps warfighting into operator, swarm, robot, and hardware component objectives. Each level corresponds to a unique perspective, world view, and development approach. Although we did not intentionally design RISE's architecture to reflect this hierarchy, it nevertheless organically evolved into a distributed system comprising four components that directly correlate to the hierarchy: a user interface for translating warfighting needs into tactics, a system for tactics development, a platform composition of primitives, and a framework for algorithm design and integration. In further detail, top-down:

\begin{itemize}

\item \textbf{Warfighting Needs}: Based on mission objectives, warfighters generate mission plans that evolve through time and drive swarm command. One must therefore translate mission plans into swarm tactics that satisfy mission objectives. However, one must maintain situation awareness (SA) in order to adapt the mission plan to new information. This requires that the swarm reports relevant information to the command and control (C2) team. Our C2 interface for mission planning, tactics execution, and SA is described in \cref{sec:rise_c2}.

\item \textbf{Tactics}: Tactics encompass the ordered arrangement and maneuver of forces on or near the battlefield \cite{tactics}. Whereas primitives are robot-centric, tactics herein are swarm-centric, being software that organizes and employs agents to achieve mission level objectives. An ``overhead scan" is an example tactic that partitions an aerial space into regions, each of which requires reconnaissance. Overhead scan may then utilize multiple agents by issuing multiple move-to primitives to complete its objective. Like primitives, one may construct tactics hierarchically. A ``breach" tactic may incorporate tactics for persistent surveillance and building entry, followed by exploration and securement. In RISE, we enable tactics development through a Python-based software application called PyC2 (see \cref{sec:rise_tactics}).

\item \textbf{Primitives}: Primitives are software components that utilize algorithms to achieve robot-centric objectives. One example primitive is the ``move-to" behavior, where an agent travels to a specific battlefield location while avoiding threats. Move-to employs algorithms for localization, path planning, and obstacle avoidance. In parallel, a second continuously running primitive enables the agent to maintain situation awareness via sensor input analysis and communication with other agents. Although move-to works to avoid threats, a third primitive may recognize an immediate inescapable danger, preempt move-to, and engage the threat. Primitives are therefore compositional and hierarchical. Our interface for primitive development and integration is described in \cref{sec:rise_primitives}.

\item \textbf{Algorithms}: Algorithms are the software components that interface with hardware by analyzing sensor input and driving actuator output. Example algorithms include computer vision techniques like YOLO \cite{yolov1} that recognize objects embedded in video frame data, localization techniques that perform dead-reckoning from continuous inertial measurement unit (IMU) data \cite{brossard2020ai}, and actuation of motor control systems from velocity commands. Single algorithms enable platform capabilities; they are also known as skills. Our interface for algorithm development and integration is described in \cref{sec:rise_algorithms}.

\end{itemize}

\subsection{The Human-Swarm Team}

Over the course of five large-scale field experiments and several intermediate integration tests, we self organized our human team into five disparate roles that we refer to as the swarm commander, swarm operator, swarm health engineer, field operations officer, and field support personnel. Each role not only accounts for one aspect of running a successful mission, but also correlates with the amount of responsibility we found one person could handle without overloading the individual. We envision that these roles will further evolve as robotics technologies, swarming, and RISE continue to mature, and that they can be adopted to a variety of organizational structures. The roles in detail are as follows:

\begin{itemize}
    \item \textbf{Swarm Commander}: A commander leads the C2 team by defining mission objectives, converting those objectives into a mission plan, modifying the plan as the mission progresses, and coordinating with C2 team members. A commander interacts with C2 software to maintain situation awareness, but does not directly command the swarm. In this way, the swarm commander's interaction with C2 software correlates with how a platoon or company leader would use the tool.
    
    \item \textbf{Swarm Operator}: An operator interfaces with the swarm by converting the mission plan into swarm tactics using C2 software. An operator is also responsible for maintaining situation awareness, reporting intelligence information to the commander, and making tactical recommendations based on evolving circumstances and swarm capabilities. The operator may also request support from the swarm health engineer to command individual agents or investigate error conditions. While there are currently no military occupational specialties (MOS) for swarm operators, we expect this role will map to several MOS categories, including the 15W, which is a single UAV operator.
    
    \item \textbf{Swarm Health Engineer}: A health engineer monitors the communications network as well as individual agents' health. An engineer is a technician who is able to diagnose and remotely resolve robotics issues. In an operational setting, this role could fit into a variety of engineering MOS specialties.
    
    \item \textbf{Field Operations Officer}: The field operations officer is responsible for logistics involving agent deployment, recovery, physical maintenance, field personnel coordination, and operational safety. An operations officer coordinates with the commander and may lead a small ground team of field support personnel. The field operations officer role is an artifact of the logistical construct of the field experiment events, the swarm hardware utilized (\textit{e.g.}, limited battery life), and safety requirements for system testing at scale.
    
    \item \textbf{Field Support}:         Field support personnel report to the field operations officer and provide the logistical power needed to deploy and maintain the swarm.
    Given the known limitation of hardware and the need for sustained operations over extended periods of time, this team provides the means to recover vehicles and prepare them for redeployment.     We expect that as the technology continues to mature, the need for this role will subside. 
\end{itemize}

In line with DARPA OFFSET program objectives, we employed a heterogeneous swarm of low-cost ground and aerial vehicles. For aerial operations, we leveraged quadcopter platforms and vertical take-off and landing (VTOL) fixed wing aircraft. Each platform type offers unique operational benefits, but also comes with its own limitations. Further details on the platforms we leverage and their tradeoffs are presented in \cref{sec:fx6swarm}.

\subsection{Live, Virtual, \& Constructive}
Time, logistics, and cost constraints prohibit frequent testing of single agent and large-scale swarm solutions. To support rapid development and integration, it is therefore critical that engineers are able to iterate their software designs without having access to physical hardware. Simulation facilitates this goal, but only when the simulation and deployment environments are uniform. Otherwise, engineers lose time supporting multiple related but incompatible interfaces. In other words, software developed in simulation should transition directly to the real world without concern over what is real or virtual. For this reason, we take care to ensure our software is abstracted from an underlying reality, such that user interface features, tactics, and primitives are unaware of the underlying platform's true nature. We further discuss our Live, Virtual, \& Constructive (LVC) approach in \cref{sec:Simulation}.

\section{The Rapid Integration Swarming Ecosystem (RISE)}

\subsection{Requirements}
\label{sec:requirements}
\newcounter{risereq}
\setcounter{risereq}{0}

\NewEnviron{requirement}{
  \noindent
  \begingroup
    \refstepcounter{risereq}
    \textbf{Req.~\arabic{risereq}}
    \label{req:\arabic{risereq}}
    \it
    \BODY
  \endgroup
}

The design of our swarming architecture is informed by program requirements as well as our experiences as an OFFSET systems integrator. We present some high-level system requirements below:

\begin{requirement}
  Capable of one-to-many (150+) operator-to-agent ratio for control of swarm agents.
\end{requirement}

In reviewing prior systems, we found that operators were unable to effectively command more than a handful of individual platforms (see \cref{sec:lit_hsi}). Thus, one of our primary goals for RISE is to enable the operation of hundreds of vehicles simultaneously by a single operator. This necessitates the concept of ``swarm operation'' as opposed to individual agent operation, whereby the operator directs the swarm as a single unit rather than providing commands to individual agents.

\begin{requirement}
  Capable of supporting 250 networked agents.
\end{requirement}

As part of the DARPA OFFSET program, RISE is required to support at least 250 networked agents in simultaneous operation, although this number was not reached in any official field experiments (see \cref{sec:field_experimentation} due to logistical (not technical) limitations. This requirement necessitated the creation of advanced networking protocols and techniques (see \cref{sec:networking}).

\begin{requirement}
  Capable of operation in a real-world, uncertain environment.
\end{requirement}

The RISE swarm must be capable of operating in real-world, non-lab environments. To enable intelligent decisions and effective C2 in the field, this requires agent-level sensing and autonomy along with robust networking and communication.

\begin{requirement}
\label{req:tacticsAbstraction}
	Tactics development abstracted from robotics software.
\end{requirement}

We found that many involved in swarm-based research are not roboticists, especially those focused on tactics development. These individuals prefer to view robots as entities capable of autonomous navigation through the environment with which they have only occasional interactions. Their concerns primarily lie in the organization and coordination of independent agents that serve an operational objective, such as to intelligently acquire and maintain situation awareness. In other words, tactical engineers typically take an exocentric world view, solving problems at a level of abstraction above the robot. An architecture serving tactical engineering needs will then separate tactics from behaviors, enabling the engineer to focus on converting a mission level objective into a set of robot commands. Therefore, RISE must provide such an abstraction to allow these engineers to easily develop new tactics.

\begin{requirement}
\label{req:robot_dev}
	Robotic development for primitives and algorithms to support tactic creation.
\end{requirement}

While tactic development was abstracted in such a way that robotics expertise is not needed, there is still the need for capable robotic platforms. Most off-the-shelf platforms do not come setup with the capabilities required to perform the tactics that were created within RISE. The tactics have a dependency upon the robots being capable of autonomous navigation through the environment and environmental interactions. These dependencies drive the development described in \cref{sec:rise_primitives} and \cref{sec:rise_algorithms}.

\begin{requirement}
    Flexible swarm command interface for tactical coordination. 
\end{requirement}

Tactics vary in complexity and oversight. While a tactical engineer may not be concerned with the underlying task management system, support for allocating robot cohorts, initiating direct and indirect command, synchronizing command execution, canceling outstanding tasks, and receiving command status is required. 

\begin{requirement}
\label{req:tacticsInterface}
  Interface for exposing new tactics, tactic parameters, and tactical control measures.
\end{requirement}

To facilitate rapid development and because tactical engineers often do not use end-user interface software, an engineer must be able to easily define a new tactic description, means of invocation, associated parameters, and required tactical control measures, which are then immediately available to the operator.

\begin{requirement}
  Support for virtual hardware in real experimentation.
\end{requirement}

While the RISE software is hardware-agnostic, research programs such as DARPA OFFSET often have funding limitations and use off-the-shelf platforms (see \cref{fig:robots}) with limited capabilities. Thus, not all desired experiment scenarios may be feasible with the physical hardware. For this reason, RISE is required to provide support for simultaneous operation of simulated and physical agents.

\begin{requirement}
\label{req:sim_low}
  Low fidelity simulation for rapid development.
\end{requirement}

While high-fidelity simulations are useful for development of robotic behaviors and for testing, we found that tactic developers often did not have the computing resources necessary for a full simulation. Therefore, one of RISE's earliest requirements is to enable low-fidelity simulation that can be run on a basic laptop computer. This allows rapid iteration on tactics development by our third-party collaborators (see \cref{sec:integrations}).

\begin{requirement}
\label{req:sim_high}
  Simulation for robotic development and evaluation.
\end{requirement}

While the low fidelity simulation is primarily utilized for tactic development, there is still a need for a simulation that supports primitive and algorithm development. Without a simulation of this nature, robotic development is tedious and always requires on platform testing. Therefore the simulation must support both mock sensor feeds and actual hardware endpoints.

\subsection{The RISE Architecture}
\label{sec:architecture}

\begin{figure}[h]
    \centering
    \includegraphics[width=.8\textwidth]{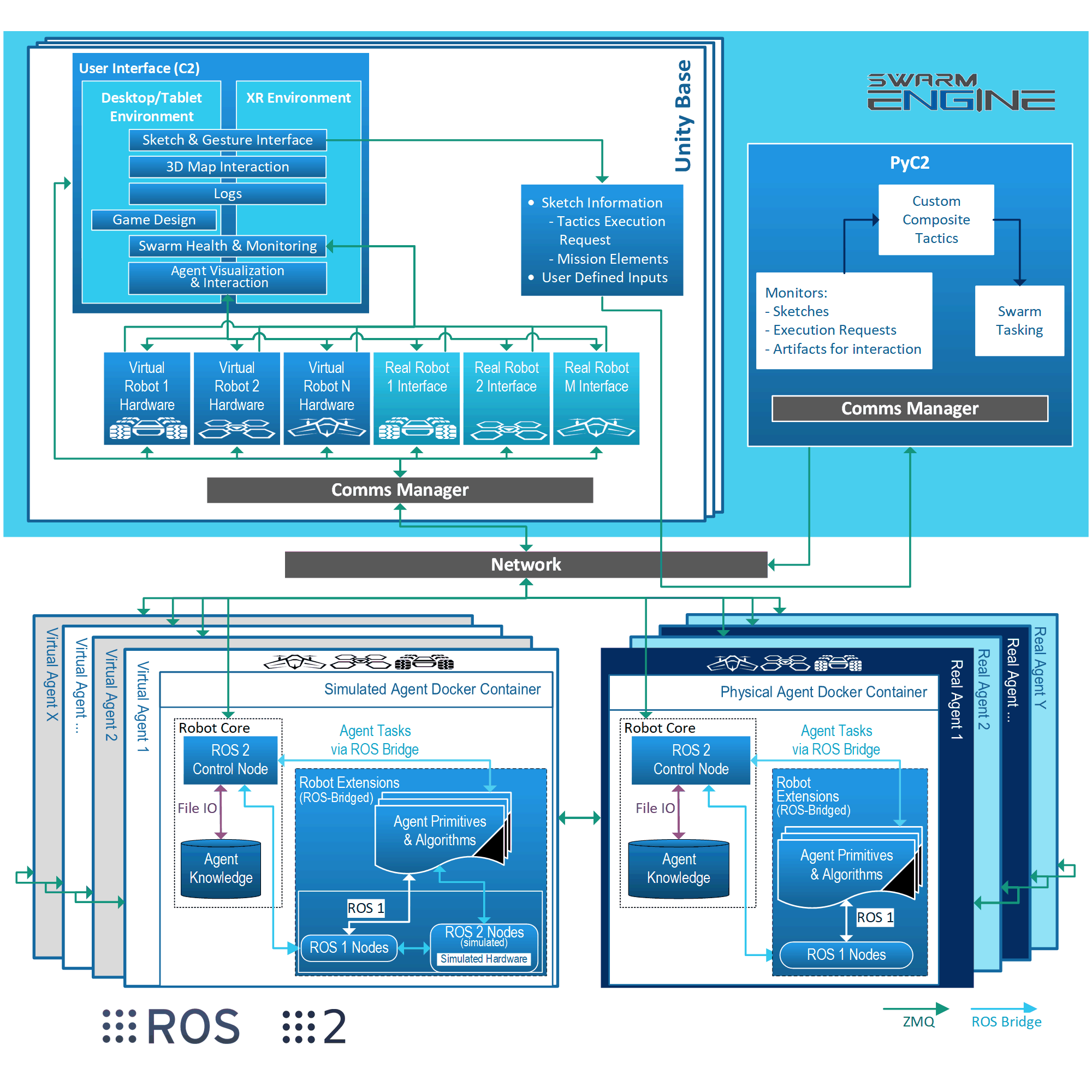}
    \captionof{figure}{The RISE Architecture. Consisting of four main components split between agent side computation and command and control computation. }
    \label{fig:architecture}
\end{figure}

The RISE architecture presented in \cref{fig:architecture} conforms with the decomposition of warfighting needs into robot actuation, as discussed in \cref{fig:tactic_tree}. One of the major components is called Swarm Engine\texttrademark{} and houses the command and control solution, which includes C2 and PyC2. The former (C2; referred to as ``Unity Base'' in \cref{fig:architecture}) implements our user interface software for environment visualization, situation awareness, and sketch-based swarm command. We implemented C2 using the Unity game engine\footnote{\url{https://unity.com}} because of its cross-platform compatibility; ability to support advanced desktop, VR, and AR interactions; and built-in simulation support, in addition to having a large active community. PyC2 is a Python-based software application designed for rapid tactics development. It comprises a number of tools designed to assist developers with writing tactics that interact with sketch input, query the environment, and generate robot primitive commands. PyC2 implements a task allocation system, and provides low fidelity simulation support so that engineers can iterate tactic designs without having to run robot software, as specified by Requirement \ref{req:tacticsAbstraction}. We chose to build PyC2 in Python because of the language's ease of use, extensive library, and wide use, thus fulfilling Requirement \ref{req:tacticsInterface}.

Agent software is divided into ``Robot Core'' and ``Robot Extensions'' modules, both encapsulated by a Docker container. Together, these components fulfill Requirement \ref{req:robot_dev}. Robot Core is a platform-agnostic compilation of ROS/ROS 2 nodes that handle key functions for task bidding, task execution, health monitoring, swarm communication, and situational reasoning and reporting. All external communication from agent to agent or agent to swarm is also facilitated through the Robot Core system. This creates a layer of abstraction that enables any agent running the Robot Core software stack to become part of the swarm. Robot Core interacts with agents via their given Robot Extensions codebase. The Robot Extensions codebase contains all the ROS nodes responsible for sensor or hardware interaction, as well as onboard agent algorithms (see \cref{sec:rise_algorithms}). Interaction between Robot Core and Robot Extensions is facilitated by standard ROS messaging, using a bidirectional ROS 1 to ROS 2 bridge\footnote{https://index.ros.org/p/ros1\_bridge/} when appropriate.

Note that in the architecture diagram (\cref{fig:architecture}), there are two variants of the agent software listed---one representing the software running on a physical agent and the other representing a simulated agent. The only differences between the two are the Docker images used and the ROS 1 sensor data sources. The simulation environment, which is the same Swarm Engine environment used for commanding real platforms, generates sensor information in place of real sensors (see \cref{sec:level3}). This distinct abstraction of sensor feeds works to facilitate simulation to hardware translation and fulfill Requirement \ref{req:sim_high}. All aspects of the agent software are contained within Docker containers for easy deployment and simulation testing. We also make use of Ansible\footnote{https://www.ansible.com/} for fleet management and deployment using this Docker based system (see \cref{sec:fx6swarm}).

\subsubsection{ROS}
\label{sec:ROS}
ROS \footnote{https://ros.org/} is a key component of the RISE architecture. Everything was developed with ROS in mind for ease of integration and development. The agent software consists of a mixture of ROS and ROS 2 nodes. Due to a lack of hardware driver support in ROS 2 at development time, the Robot Core modules use ROS 2 and the Robot Extensions modules currently use ROS, although Robot Extensions may be upgraded to ROS 2 at a later date when all required drivers are available in ROS 2. All inter- and intra-agent messages are transmitted using either native ROS 1/2 messages using standard ROS networking middleware or ROS 2 messages using a custom ZeroMQ middleware (see \cref{sec:networking}). All communication between components of Swarm Engine and varying instances of the Unity environment uses ROS messaging. The simulation environment broadcasts ROS sensor messages to integrate with the agent software for development and testing. ROS messaging is how all algorithms and primitives communicate, and is used at every level of the system, from tactic tasking down  to agent-level actuation. ROS's MAVLink/MAVROS package handles all sensor and actuator interaction on each agent. During the course of development for RISE, several releases of ROS and ROS 2 were utilized, but the program currently uses ROS Melodic and ROS 2 Eloquent.

\subsection{Swarm Networking}
\label{sec:networking}
This section describes the network used to enable communication amongst agents and between agents and the operators. We start by providing a summary of early prototypes and challenges and then describe the final topology and protocols.

\subsubsection{Initial Prototypes}
\label{sec:netprototypes}
In the early stages of RISE development, extremely simple off-the-shelf mechanisms were used for swarm networking. Due to availability of components, RISE initially used standard 802.11 Wi-Fi with all agents connected to centralized access points. However, field experiments (see \cref{sec:field_experimentation}) quickly exposed the downsides of this approach. Operators needed to pre-position the communication infrastructure before mission execution could begin. This was a difficult process requiring careful site surveys to ensure maximum coverage and establishment of a power supply and network backhaul link at the chosen access point location. Obviously, such preparatory work was deemed impractical for real-world tactical exercises. Additionally, even with site surveys and careful access point placement, the dense concrete urban environment used for field experiments (see \cref{sec:field_experimentation}) yielded large ``dead zones'' in which agents were unable to communicate with the pre-installed access points.

To alleviate these issues, and with the addition of further funding, we conducted an evaluation of commercial off-the-shelf networking solutions and purchased equipment for a mobile ad hoc network (MANET) from Rajant Corporation. See \cref{sec:manet} for more information on the benefits of this approach.

However, after switching to a MANET architecture, we encountered further problems. While initially we had been using the default ROS 2 messaging mechanism for inter-agent communications, we found that this did not scale well to more than a few tens of agents when using a MANET. ROS 2 messaging is based on the Data Distribution Service (DDS). Thanks to the MANET's shared medium (2.4 GHz and 5 GHz radio frequency bands), large numbers of radios attempting to transmit simultaneously caused significant data loss and retransmission, rendering the network practically unusable. We worked extensively with our DDS vendor's engineers to resolve this problem, but were still unable to operate hundreds of agents simultaneously. At that point, we began implementation of our own  protocol solution for inter-agent communications. See \cref{sec:protocols} for more information.

\subsubsection{Mobile Ad hoc Network}
\label{sec:manet}
One of the critical challenges in fielding a multi-agent robotic system is in the design of a robust communication network. While several architectures were initially considered, we employed a mobile ad hoc network (MANET), commonly referred to as a mesh network, in order to facilitate all communication between agents and all non-emergency communication with C2 operators.

We selected the MANET architecture for a variety of reasons, which we list below:
\begin{itemize}
    \item Urban environments typically comprise structures of varying materials and geometric complexity that obstruct radio communications, preventing access-point-based networks from effectively communicating with agents inside of buildings or agents in-between structures. We found this to be especially true in the environments used for our field testing (see \cref{sec:field_experimentation}). A MANET alleviates this problem by allowing other agents in the vicinity to act as network relays, a capability that RISE's tactics exploit. For example, see the building clearing capability developed by SoarTech in \cref{sec:soartech_integration}.
    \item By using a MANET, we make optional the pre-positioning of network equipment inside the operational area, a task that is required when using networks designed around central access-points.
    \item A MANET architecture eliminates any single-points-of-failure in the system, allowing agents to continue operation even if one or more of the ground radios fail. The agents may communicate directly with any other devices in the network without routing messages through any centralized infrastructure.
    \item The MANET architecture allows for easy swarm deployment, a critical capability when operating large-scale swarm systems. Agents do not need to be reconfigured for different deployment strategies, and the ground infrastructure can be changed on-the-fly without any network reconfiguration.
\end{itemize}

\begin{figure}[t]
    \centering
    \includegraphics[width=.6\textwidth]{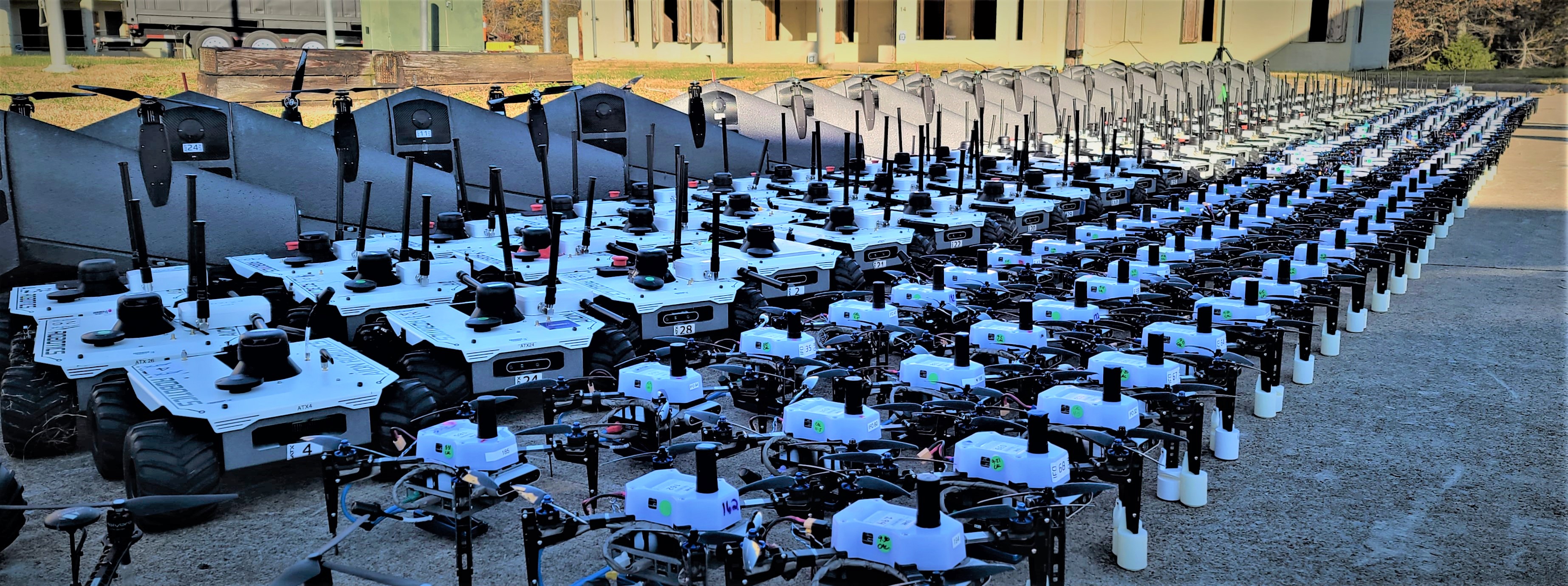}
    \includegraphics[width=.35\textwidth]{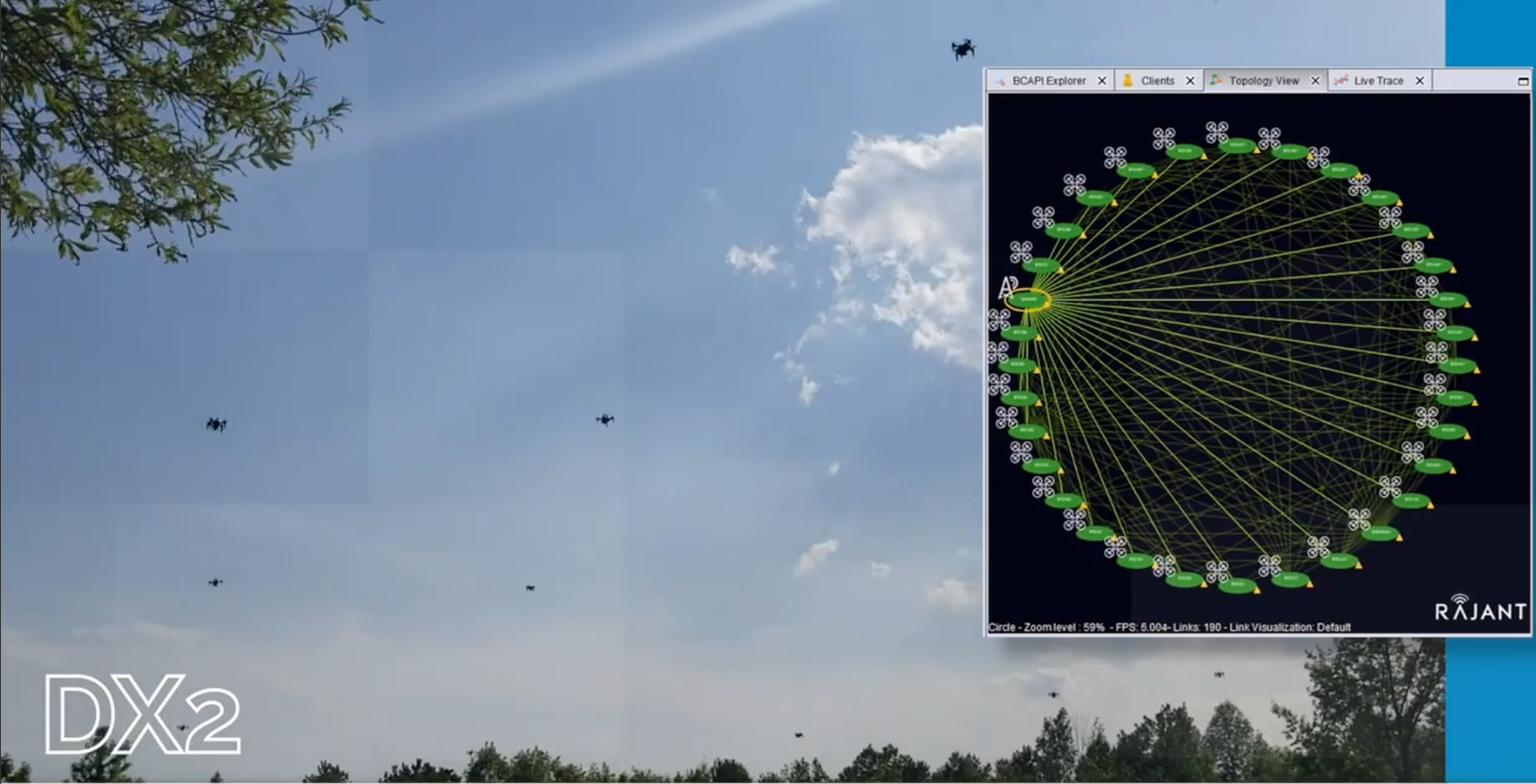}
    \captionof{figure}{At left, Rajant ES1 radios are seen mounted to ground vehicles. Aerial vehicles have Rajant DX2 radios mounted underneath. At right, mesh radios are shown communicating between IFO platforms during a small-scale field test, as highlighted by Rajant's network management software.}
    \label{fig:radiosMounted}
\end{figure}

For these reasons, we chose a commercial off-the-shelf MANET solution from Rajant Corporation, utilizing a number of different radios from their product line in our system. Air vehicles are equipped with Rajant DX2 or DX4-series radios, while ground vehicles are equipped with the ES1. The ME4 and Peregrine-series radios are used for ground/base communications, connecting the operators' Ethernet network to the MANET. The mounting of these radios are shown in \cref{fig:radiosMounted}. All radios operate on the 2.4 GHz band. Except for the DX2s, all radios also utilize a second channel on the 5 GHz band. Rajant software manages the use of these bands to allow for efficient transmission with many radios in close vicinity.

\subsubsection{Network Overview}

\begin{figure}[t]
    \centering
    \includegraphics[width=.8\textwidth]{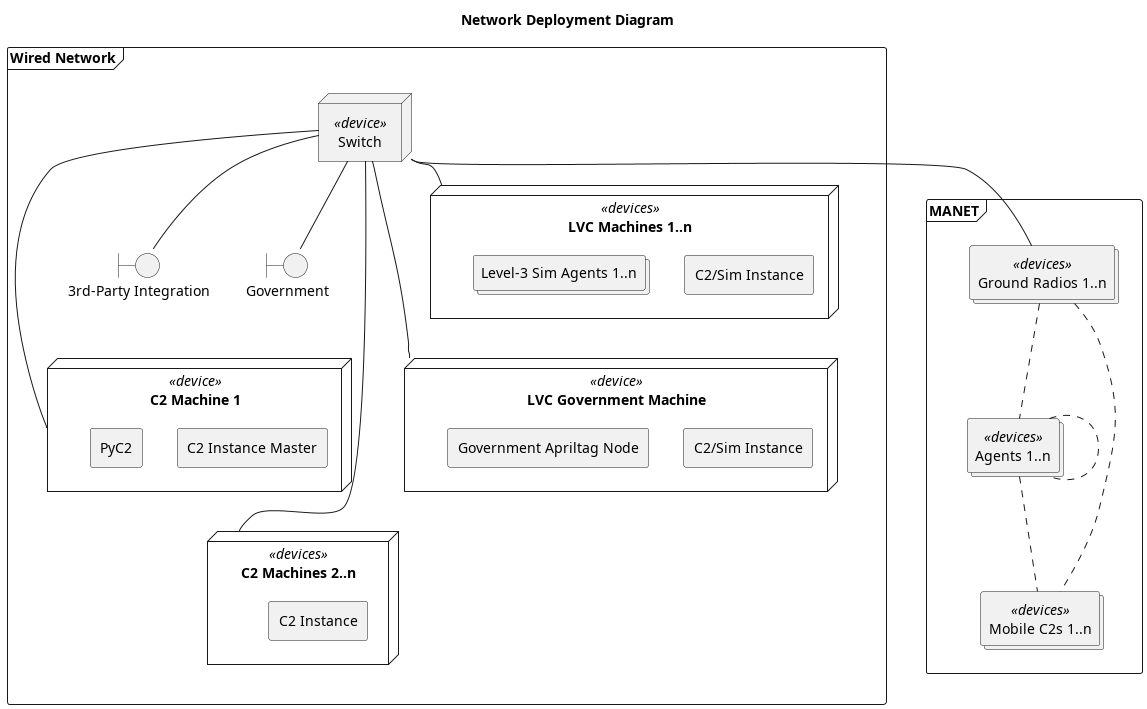}
    \includegraphics[width=.5\textwidth]{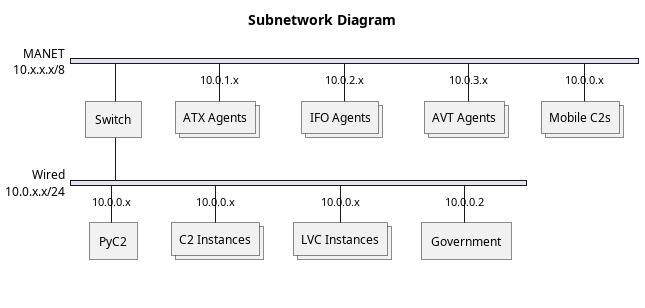}
    \captionof{figure}{Above, a deployment diagram of the overall RISE network. Items suffixed by ``1..n'' may be of arbitrary quantity. Below, a subnet diagram of the RISE network. In both diagrams, ``Government'' refers to the government experimentation infrastructure (see \cref{sec:field_experimentation}).}
    \label{fig:netDeploymentDiagram}
\end{figure}

While the MANET described above is critical for agent communication, it cannot handle the high volumes of data necessary for LVC sensor simulation (see \cref{sec:Simulation}). For this reason, we add a 1000BASE-T Ethernet network, connected to the MANET via a switch. This Ethernet network serves multiple LVC simulation machines, plus various C2 computers, and facilitates connections with 3rd-party ground systems (see \cref{sec:integrations}). The overall network is shown in \cref{fig:netDeploymentDiagram}.

\subsubsection{Network and Transport Protocols}
\label{sec:protocols}
Initially, we attempted to use ROS 2's built-in network stack, which is based on the Data Distribution Service (DDS). However, we quickly found that DDS was not suitable for our large-scale MANET due to DDS's discovery phase overhead \cite{dds}. We instead implemented a custom solution using ZeroMQ \cite{zmq} as a socket library to allow for reliable multicast communication between all networked systems.

We use the Internet Protocol (IP) \cite{rfc791} as the basis of our swarm network. Every agent has an IP address and is joined to one or more IP multicast groups using the Internet Group Management Protocol (IGMP). In multicast, a device must only transmit one message, which is addressed to a group and will be received by all members of the group \cite{rfc1112}. This is advantageous in our architecture, since it reduces the amount of air time that a vehicle's radio must use for message transmission. Transmitting a unicast message once for every agent that must receive it (often 200+) requires a significantly larger amount of airtime than transmitting a single multicast message. In a swarm with potentially hundreds of radios in proximity, this is a critical consideration. We take advantage of Rajant's ``Tactical Multicast'' feature to prevent multicast messages from ``echoing'' around the network. This feature allows multicast messages to reach all vehicles with minimal retransmission throughout the MANET \cite{rajantMulticastPatent}.

We divide our messages into two transport classes: reliable and unreliable. Both use the IP network protocol, but have different transport protocols. For unreliable messages, standard User Datagram Protocol (UDP) datagrams are employed to transport messages. This protocol ensures message correctness, but does not guarantee delivery or reception order \cite{rfc768}. Such a protocol is ideal in a dense transmitter environment, since it minimizes transmissions and does not waste air time by attempting to retransmit unimportant data.

For messages that require delivery guarantees, we use the Pragmatic General Multicast (PGM) protocol. PGM uses negative acknowledgments to enable retransmission of lost or corrupted messages. That is, each device maintains a sequence number, which is periodically transmitted in a heartbeat message (SPM packet) to other devices. This allows receiving devices to recognize message loss and request retransmission of those messages using a negative acknowledgment (NAK) \cite{rfc3208}. We changed several PGM parameters from their default values to allow for more efficient use of our limited network resources. Information on these parameter changes is included in \cref{app:network}.

\subsubsection{Application Protocols}

\begin{figure}[t]
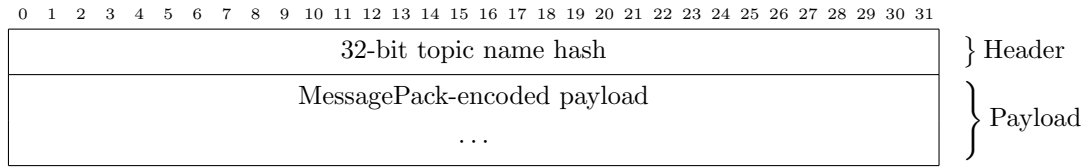

    \centering
    \begin{bytefield}[bitwidth=1.1em]{32}
        \bitheader{0-31} \\
        \begin{rightwordgroup}{Header}
            \bitbox{32}{32-bit topic name hash}
        \end{rightwordgroup} \\
        \begin{rightwordgroup}{Payload}
            \wordbox[lr]{1}{MessagePack-encoded payload} \\
            \wordbox[blr]{1}{$\cdots$}
        \end{rightwordgroup}
    \end{bytefield}
    \captionof{figure}{The network application protocol.}
    \label{fig:netAppProto}
\end{figure}

As all agents in the system use ROS 2 for their high-level logic and control, integration of the network stack with ROS 2 was a crucial requirement. We created a ``Communication Handler'' ROS 2 node and package for this purpose. The node accepts a user-specified list of ROS 2 topics (see \cref{sec:ROS}) to be transmitted to or received from the network. 
The application protocol was designed to be straightforward and easy to implement. We first encode the topic name as a 32-bit value using the DJB2 hash algorithm \cite{djb2}. Then, we use the MessagePack library to encode each field of the message using MessagePack's efficient encoding scheme \cite{msgPack}. Each field of the ROS 2 message is automatically decoded and then encoded as a MessagePack field, and the whole message is packed into a structure described by \cref{fig:netAppProto}. The message is then transmitted using the ZeroMQ library over one of the transport mechanisms described above.

To receive a message, the reverse occurs. The message is received by ZeroMQ, the topic hash compared against a dictionary of known topics (and the message dropped if unknown), and then each field is decoded by MessagePack and re-encoded as a ROS 2 message. The ROS 2 message is then published using the standard ROS 2 publishing mechanism.

ROS 2 publications, subscriptions, services, and service clients are supported with automatic message translation in both directions. See \cref{sec:api} for more information on topics used in the swarm. The ROS 2 project originally investigated the use of ZeroMQ, but chose DDS for ease of development at the cost of messaging customization \cite{rosZmq}. We have implemented ROS 2 messaging over ZeroMQ and have realized many of those performance and customization gains. While we have lost some of ROS 2/DDS's flexibility (such as automated topic discovery), these changes allow the system to function under the heavy constraints imposed by our 200+ vehicle mesh network.

\subsubsection{Emergency Control Channel}

We also maintain an emergency out-of-band control channel in the 2.4 GHz range. The Swarm Operator may command UAV emergency stop via this channel, which will immediately cause all UAVs to land and disarm.

\subsection{Swarm API}
\label{sec:api}

\begin{figure}[h]
    \centering
    \setlength{\fboxsep}{0pt}
    \fbox{\includegraphics[width=.5\textwidth]{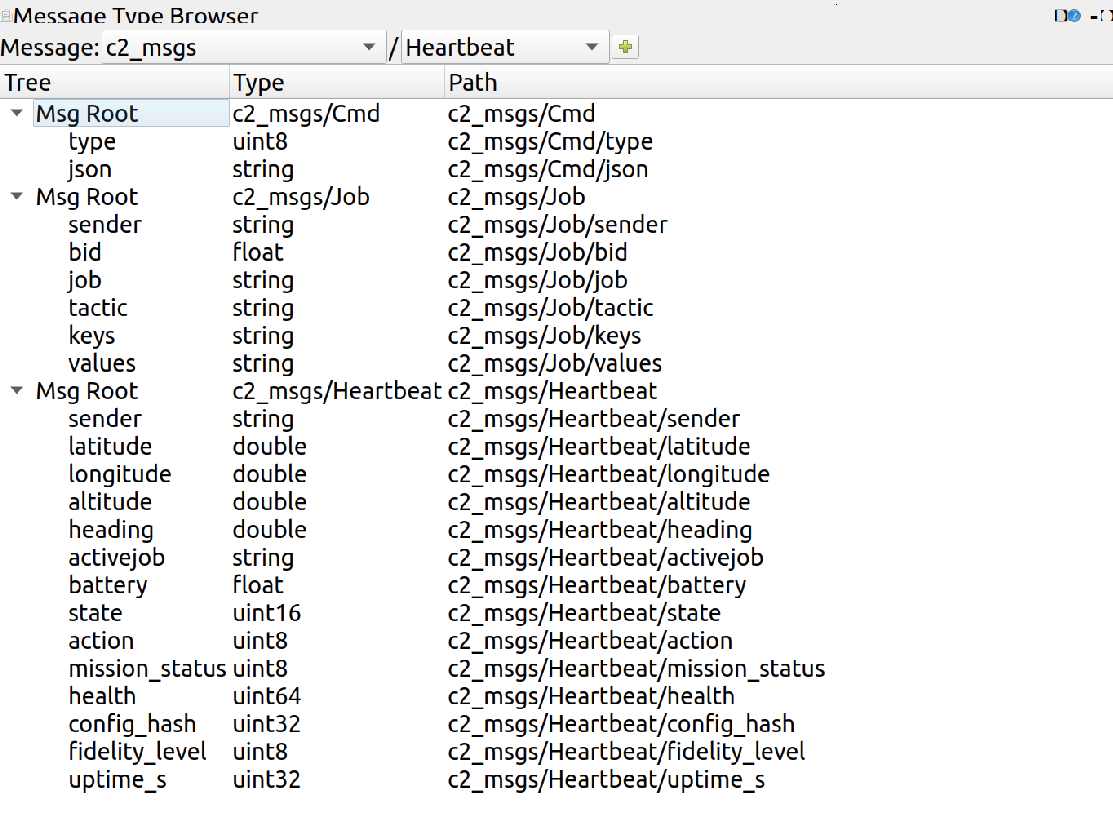}}
    \captionof{figure}{RQT Message view showing some of the primary API messages. Heartbeat, Cmd, and Job }
    \label{fig:rqt}~\\~\\
\end{figure}

As mentioned in \cref{sec:ROS}, ROS is utilized throughout RISE and is the common API interface. As with any ROS-based system, the interfaces between nodes are well defined by ROS messages, publishers, and subscribers.  Agents in the swarm primarily interact through three ROS 2 messages which we define: the Heartbeat, Cmd, and Job messages. \cref{fig:rqt} shows the contents of these three primary messages utilized by swarm agents and Swarm Engine. Individual agents of the swarm must populate the elements of the Heartbeat message and send that information repeatedly on the /a2c/heartbeat topic. As agents receive tasking or other factors occur, they must update the respective fields in the message. All Swarm Engine entities listen to that message topic to automatically discover swarm agents and add them to their agent table. Likewise, Swarm Engine instances will broadcast the same Heartbeat message information on a /heartbeat topic so that swarm agents are aware of all the Swarm Engine instances that are online and other Swarm Engine instances are also aware of each other.

\begin{figure}[h]
    \centering
    \includegraphics[width=.5\textwidth]{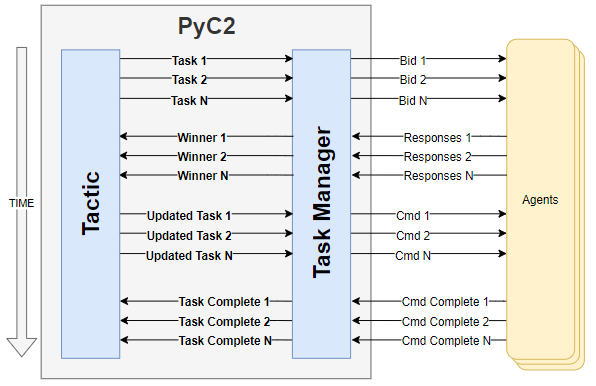}
    \captionof{figure}{Bidding command flow.}
    \label{fig:bidding}
\end{figure}

\begin{figure}[tph]
    \centering
    \includegraphics[height=.9\textheight]{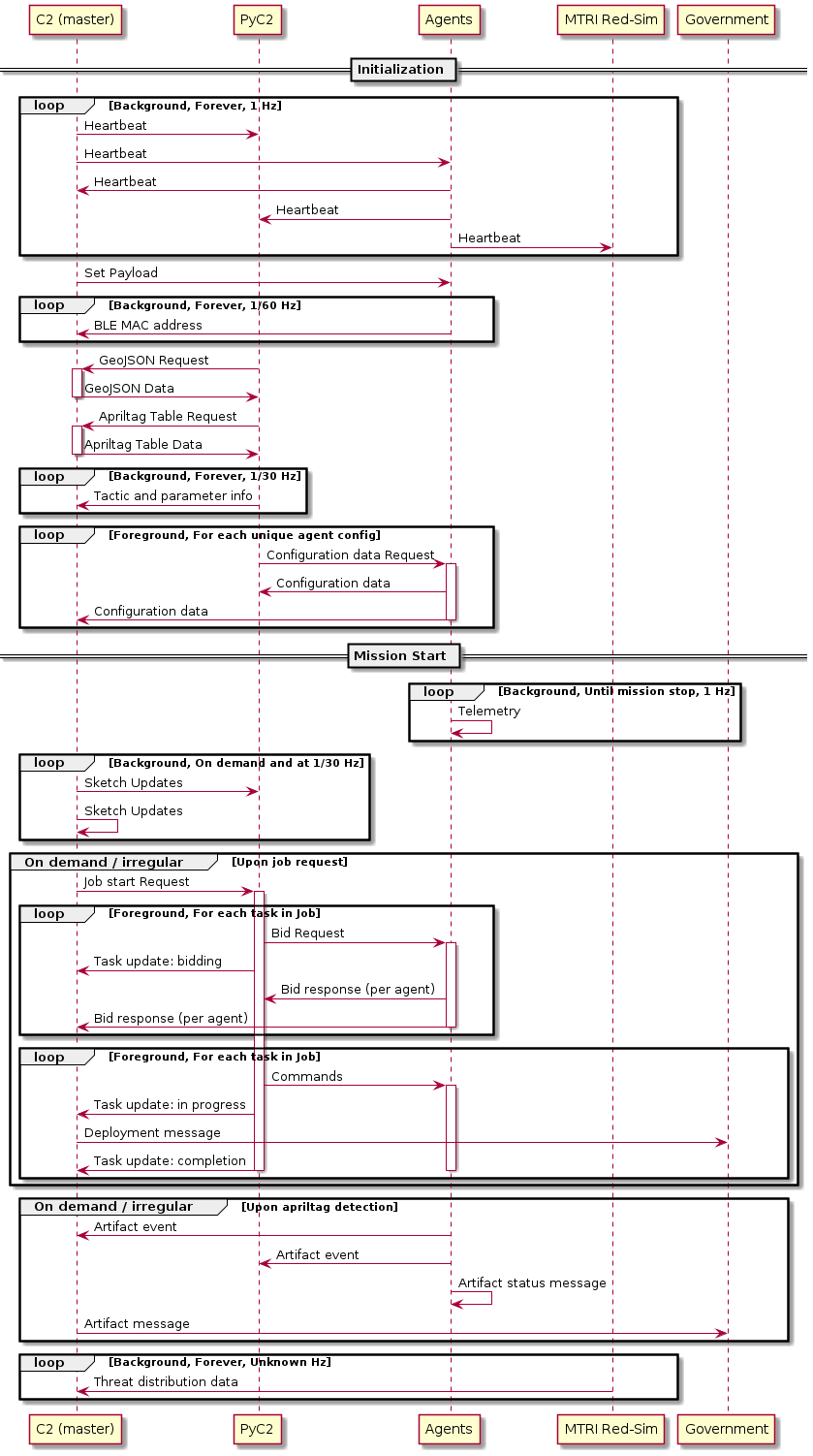}
    \captionof{figure}{Sequence diagram showing messages sent over the network during swarm operation.}
    \label{fig:netSequence}
\end{figure}

The Job and Cmd message are used as part of the bidding process that makes up the swarm task allocation. \cref{fig:bidding} demonstrates how tasks are generated from tactic execution and are used to prompt swarm agents to bid via the Job message and bidding topics. After task evaluation has been completed based on bid responses, a command message is sent out to the swarm, instructing agents to complete the requested tactic. This interaction is facilitated by the Cmd message, with agents subscribing to the /command topic to determine if they have had a task assigned to them. If an agent has been tasked, logic flows into the agent primitive part of the architecture. Otherwise, agents continue to respond to task requests on the /bidding topic as additional tactics are executed over the course of the mission. A sequence diagram showing the swarm in operation is available in \cref{fig:netSequence}.

\subsubsection{Command API}
\label{sec:bidding}

Agent task assignment within RISE is handled via a centralized bidding system. While research on decentralized task allocation is quite common \cite{johnson2010improving} and RISE has even been integrated with third parties that have implemented decentralized task allocation (see \cref{sec:heron_integration}), we found centralized tasking to be the ideal solution for RISE's needs. First, centralized task allocation provides a much higher level of control of the swarm. It is abundantly clear when agents will take tasking and who the anticipated agents will be in a centralized system. Another primary reason for the centralized tasking is network load. With the numbers of agents in our use case, we need to limit network traffic wherever possible, and a centralized system allows for this.

The bidding system works through the use of the job and command messages described in \cref{sec:api}. As a swarm operator provides tactic input, a series of tasks or "jobs" are created within our system. These jobs are then sent out via the /bidding topic and agents respond with a heuristic-based bid. The agent heuristic looks at the job being requested, the location of the job relative to itself, battery life, vehicle health, and a few other factors, before it responds with a bid. PyC2 then evaluates all job responses received and eventually assigns tasking for all the jobs in a respective tactic. This final tasking is sent out via the /command topic and message. \cref{fig:bidding} showcases the standard interaction over time for this overall process. Multiple tactics and jobs can be simultaneously bid out.

\subsubsection{Intelligence API}
\label{sec:intel_api}
All agents in the swarm are constantly searching for intel, regardless of the task they are performing. This intelligence is primarily gathered through the various platforms' RGB cameras. While YOLO or some other variation of object detection has been utilized during RISE, intel in DARPA OFFSET field experiments was primarily represented by AprilTags \footnote{https://github.com/AprilRobotics/AprilTag} (See \cref{sec:FX_scenario}). Agents utilize an AprilTag detection algorithm to report to the swarm what they saw via a common detection message which contains information such as who saw the artifact, what does it represent in its current context, where was the artifact, and so forth. All agents and Swarm Engine instances listen for all reported detections from each other. As detections are reported, agents record the events in another subcomponent of Robot Core called the central data store. This information is stored and queried by certain primitives as any prior or dynamic intel is required for that primitive execution. \cref{sec:artifacts} discuses how Swarm Engine visually represents reported intelligence from swarm agents.

\section{RISE: User Interface}
\label{sec:rise_c2}

\begin{figure}[t]
    \centering
    \includegraphics[width=\textwidth]{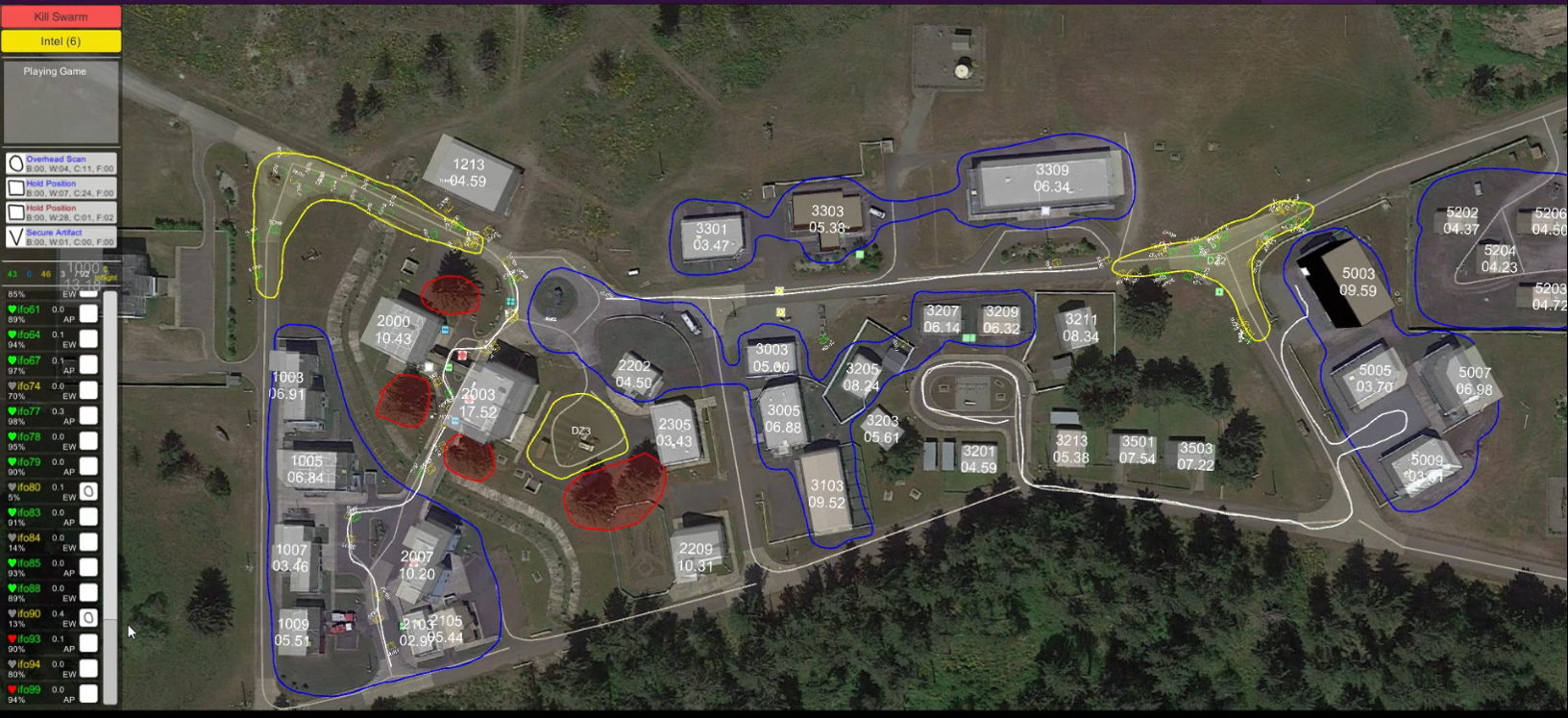}
    \captionof{figure}{Mission plan sketch drawn by the swarm operator during the fourth DARPA OFFSET field experiment at Joint Base Lewis-McChord.}
    \label{fig:fx4_mission_plan}
\end{figure}

We built our C2 user interface to balance the needs of swarm commanders, operators, and health engineers. Swarm commanders are primarily concerned with situation awareness (SA), one aspect of which is ``the perception of elements within a volume of time and space" \cite{endsley1995toward}. Elements of interest include blue and red force positions, terrain layout, and mission plan data. Although operators must similarly maintain SA, they are also responsible for swarm command---the issuance of tactics that effectuate the commander's plan. Given that robotic swarms interact with the environment, tactics are inherently grounded in a temporal and spatial context. The nature of SA and tactic invocation therefore implies that both the commander and operator primarily communicate with the swarm through the environment. On the other hand, swarm health engineers, as do researchers, practitioners, and developers, require access to detailed information that commanders and operators do not necessarily require. Agent task status, logs, component-level health information, and trajectory information are example data that engineers use to diagnosis issues occurring during a mission. Although, health engineers utilize SA and may aid swarm command, being able to quickly locate and interact with specific agents is critical to their success. \cref{fig:fx4_mission_plan} illustrates how we balance these requirements. 

The proposed cross-platform compatible interface was refined through an iterative design approach spanning four years. Our experiences participating in five large-scale field experiments across five Combined Arms Collective Training Facilities (CACTFs), along with periodic field integration testing and regular development, helped inform our design. As shown, the interface is divided into an interactive panel (left) and scene view (right). Given that the environment is our primary communication channel for SA and swarm command, we appropriate significant screen real estate to the scene view. Bandwidth is further maximized by abandoning traditional WIMP (windows, icons, menus, and pointers) design patterns. Instead, we implement a context-sensitive gesture-based user interface, whereby users interact with C2 via sketch commands. This decision also serves our desire for cross-platform capability, as we are able to preserve limited mobile device screen space. In the remainder of this section, we expand on these ideas and describe our interface in detail.

\subsection{Cross Platform Compatibility}
We designed a cross-platform compatible C2 by implementing techniques that utilize only 2D input, which we regard as the least common denominator among all input devices. Mouse, stylus, and touch input are inherently 2D. Three-dimensional controller and hand pose data can be made 2D via planar projections. For this reason, baseline C2 techniques learned on one system transfer to all supported systems. To date, we have ported our software to Windows, Linux, Android, HoloLens 2, and HTC Vive. It is important to note, however, that although we implement a common cross-platform interface, we are still able to exploit affordances offered by more capable systems. Additional information on alternative HMI interfaces supported by RISE can be found in \cite{Williamson2023}.

\subsection{Gesture Interface}
Human motion that intentionally conveys information is a gesture. In the context of this work, gestures are motion patterns captured by an input device that map to software commands---when we recognize a known pattern, we invoke the associated function. To recognize input patterns, we employ Jackknife \cite{jackknife}, a device-agnostic custom gesture recognizer. 

Being device-agnostic means we are able to recognize mouse, touch, stylus, hand, and 3D controller gesture input, among other modalities using the same recognizer, which enables rapid integration of new input modalities. Being customizable means the recognizer learns from a small set of example input patterns. Jackknife specifically achieves high accuracy ($>90\%$) with only one training sample loaded and improves with more training data\footnote{On 2D gestures specifically, Jackknife achieves 95\% accuracy in a writer-independent recognition scenario with two templates loaded.}. Since we only require minimal training data, users can customize the interface according to their preference, which has potential to increase learnability and memorability \cite{nacenta2013memorability}. Further, because Jackknife uses a nearest neighbor pattern matching strategy, we can train the recognizer online in real time, enabling us to define new gesture classes on demand. When RISE implements a new feature, but the associated invocation gesture is unknown, our interface prompts the user for training data, after which the recognizer is retrained and the new feature is immediate accessible. This enables PyC2 to employ new tactic and sketch parameter types without direct C2 support.

All interface functions are accessed via gesture commands, including system commands to start and stop missions, agent commands to access logs and video feeds, and tactic and tactic parameters commands. Besides preserving screen space, an additional advantage in using gestures over WIMP design patterns is that C2 and PyC2 developers can add new functionality without having to reorganize menu hierarchies or toolbars, thereby accelerating development time. More importantly, however, gestures reduce mode switching and allow the user to interleave commands, potentially increasing command throughput.

To aid swarm operators learn the user interface, we developed an interactive training module that walks users through the basic agent, command and control, and navigation interactions. Separately, all available gesture commands can be found through an online help system shows the user how to draw each symbol and informs them on their usage.

\subsection{Agent Interface}

\begin{figure}[t]
    \centering
    \includegraphics[width=.75\textwidth]{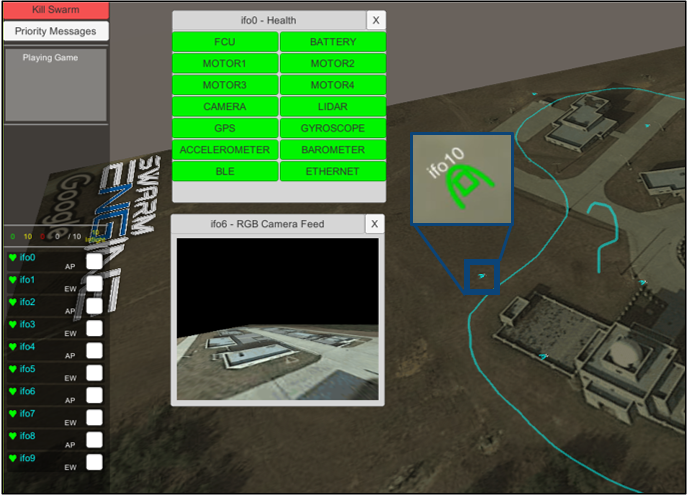}
    \captionof{figure}{User interface elements illustrating the agent list (left), video stream, component health panel, and agent icons within a map view simulation.}
    \label{fig:agent_panel}
\end{figure}

More so than commanders or operators, swarm health engineers are concerned with individual agents. We therefore provide access to essential agent interactions through the left panel shown (\cref{fig:agent_panel}). In this panel, we generate a list of all known agents---those for whom we have received one or more heartbeat messages. Each list entry communicates the agent's name, payload, status, and tactic icon\footnote{Although agents execute primitives, we are able to trace its work back to the tactic that generated it.}. The agent name's color encodes its status, which may be idle (green), killed (orange), manually disabled (red), tasked (blue), or unknown (orange). This later state indicates that C2 has lost communication\footnote{Communication is lost when C2 has not received a heartbeat in seven seconds} with the agent.

A user may access agent-specific data by gesturing on the agent's button, where one's initial contact point selects the agent. A right-swipe causes C2 to position the camera over the agent in the scene view. An \emph{h} gesture opens the agent's health information window wherein all relevant hardware components are listed with color coded status, those being nominal (green), missing (gray), intermediate (yellow), or critical (red). Color and depth camera streams from the agent can be viewed by \emph{h} and \emph{)}, respectively. Finally, \emph{L} displays the agent's logs. Although we are able to provide access to additional features, these were found to be most relevant for initial agent SA and diagnosis. A health swarm engineer will have access to additional information via SSH remote terminal access. 

We present each agent as an icon within the scene view using a platform-specific symbol. Its color reflects the agent's status using the same scheme described above. Within the agent icon, we render a tactic symbol that reflects ongoing or previously completed work in which the agent is or was engaged. In an early user interface iteration, we instead rendered agents as 3D models. However, other than being more aesthetically satisfying, they provided no clear advantage, which we opted for a less resource intensive variant. 

\subsection{Command and Control (C2)}

The heart of swarm command in RISE comes down to specifying tactic commands, their inputs, and their execution order. These three elements are brought together through our sketch-based interface, as described next.

\subsubsection{Tactic Invocation}
\label{sec:tactics_ui}

\begin{figure}
\footnotesize
\begin{tabularx}{\textwidth}{lcllX}
    
		\textbf{Tactic Name} & \textbf{~~~Gesture~~~} &\textbf{Parameter} & \textbf{Type} & \textbf{Description}\\
		\midrule		
		
        \multirow{2}{*}{Examine Object} & 
		\multirow{2}{*}{\includegraphics[width=.055\textwidth]{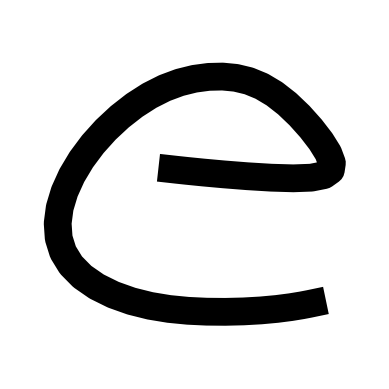}}
		&Context&POI&Use UAV to scan an object of interest.\\
		& & Radius & float & Radius of sphere around object.\\
		& & & & \\
		
		\midrule
        \multirow{4}{*}{Follow Route} & 
		\multirow{4}{*}{\includegraphics[width=.055\textwidth]{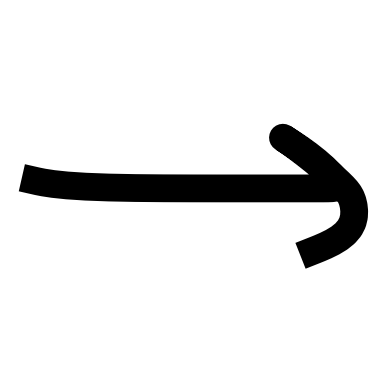}}
		&Context&Path&Request an agent to traverse the nearest path.\\
		& & Altitude & float & Height in meters that UAV will assume.\\
		& & Distance & float & Distance between points. Zero to force simplification.\\
		& & Use Chaining & bool & Issue one point at a time, for testing only.\\
		
		\midrule
        \multirow{5}{*}{Hold Position} & 
		\multirow{5}{*}{\includegraphics[width=.055\textwidth]{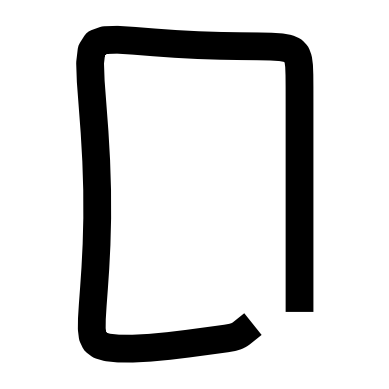}}
		&Context&Path&Move a set of agents to points along the perimeter and hold.\\
		& & Altitude & float & Height in meters that UAV will assume.\\
		& & Duration & float & How long to hold.\\
		& & Agent Count & int & Number of agents to place along perimeter.\\

		\midrule
		\multirow{3}{*}{Overhead Scan} & 
		\multirow{3}{*}{\includegraphics[width=.055\textwidth]{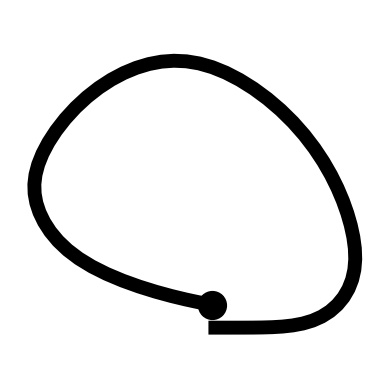}}
		&Context&Explore&Fly UAVs over area to find artifacts.\\
		& & Altitude & float & Height in meters that UAV will assume.\\
		& & Cell Size & float & Minimum linear distance between waypoints in meters.\\
		& & Agent Count & int & Number of agents used to scan area.\\

        \midrule
        \multirow{1}{*}{Safe Land} & 
		\multirow{1}{*}{\includegraphics[width=.055\textwidth]{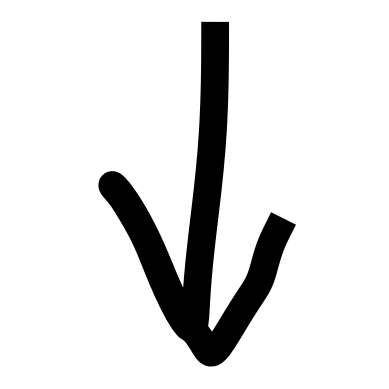}}
		&Context&None&For a given air vehicle, find nearby safe location to land.\\
		& & & & \\
		& & & & \\
		
		\bottomrule
\end{tabularx}
\captionof{table}{Subset of PyC2 tactics an operator may invoke to fulfill mission objectives. The operator draws a tactic's gesture into the environment to access the command. Each tactic requires zero or more parameters, where context is the sketch input type on which the tactic will execute. The operator can modify remaining parameters through a pop-up window before submitting the command to PyC2. Note, this \LaTeX~table was generated with PyC2's automated documentation system.}

\label{tab:commands} \end{figure}

To invoke a specific tactic, an operator must draw its associated gesture into the environment. When the gesture is recognized, C2 inserts an interactive tactic icon at the stroke's centroid. The operator may then click on the tactic to open its associated popup window. This window comprises an input parameter list and two interaction buttons, one that immediately issues the tactic to PyC2 for further processing, and another that simply closes the window to enable mission planning. \cref{tab:commands} shows five example tactics---their invocation gesture and required input parameters. The ``Context'' parameter specifies which sketch input parameter type a tactic will act upon.

The location of a tactic icon within the environment is relevant in that tactics optionally use position information to infer operator intention. To illustrate, our scan building tactic infers an operator wishes to scan that building which is closest to the icon. Tactics make similar inferences on sketch parameter input. Our overhead scan tactic, for example, operates on the closest explore area sketch (an input parameter type discussed below). For this reason, an operator may drag their tactic icons through the environment in order to precisely assign position.

Once invoked, tactic status is encoded into the icon's color, being one of the following: pending (black), in progress (blue), failed (red), or successfully completed (green). We also add a tactic button to the left panel in order to provide a quick enumeration of ongoing work as well as provide access to additional developer and swarm heath engineer data. This includes access to children tactic and primitive data and agent waypoint lists when applicable. As the list grow long, it becomes difficult to correlate specific tactic buttons with ongoing field work. However, utilizing spatial memory, one may also return to the tactic icon within the scene view at any time in order to access the same information via a pop window. To cancel a tactic as well as its children tactics and primitives, one may simply scribble-erase the icon. 

\subsubsection{Tactic Chaining}

\begin{figure}
    \footnotesize
    \centering
    \includegraphics[width=.50\textwidth]{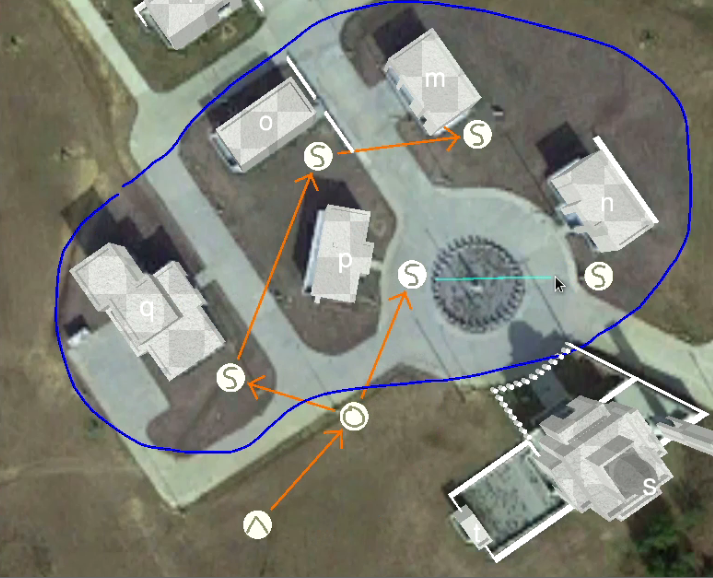}~
    \includegraphics[width=.49\textwidth]{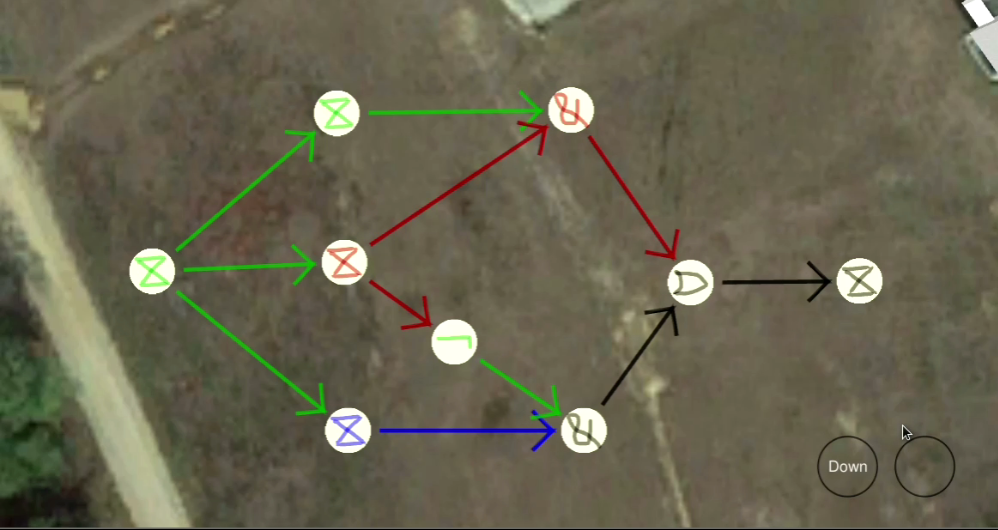}~\\
                ~\\


\begin{tabularx}{\textwidth}{cXcXcXcX}
    
    \multicolumn{8}{c}{\textbf{Gesture Legend}}\\
    \midrule
	
    \includegraphics[width=.025\textwidth]{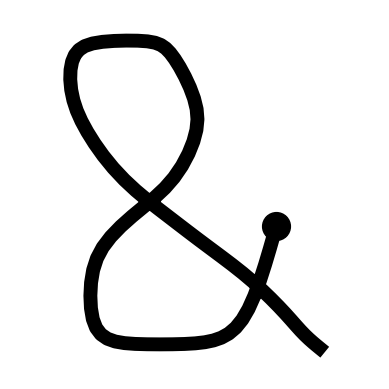} &Conjunction
    &\includegraphics[width=.025\textwidth]{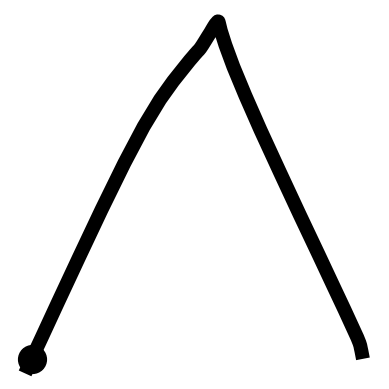} &Create agents
    &\includegraphics[width=.025\textwidth]{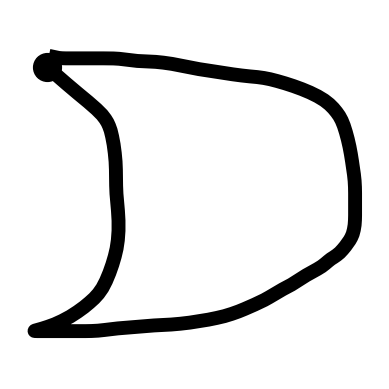} &Disjunction
    &\includegraphics[width=.025\textwidth]{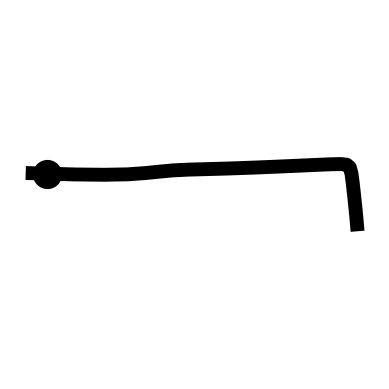} &Negation\\
    
    \includegraphics[width=.025\textwidth]{images/ui/gestures/circle.png} &Overhead scan
    &\includegraphics[width=.025\textwidth]{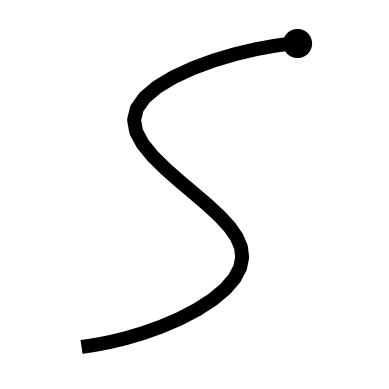} &Scan building
    &\includegraphics[width=.025\textwidth]{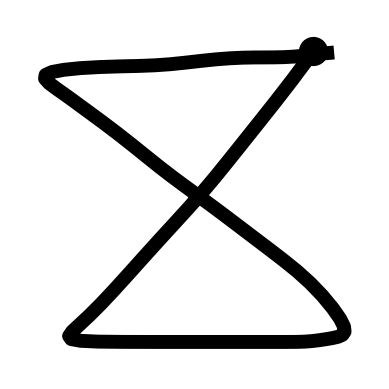} &Timer\\
        
    \bottomrule
\end{tabularx}
    \captionof{figure}{Tactic chaining. Left: Operator drawing a link between two nodes to form a tactic chain link. Right: Example tactic chain including logic gate-like nodes. Tactic icons and links are color coded where blue indicates the tactic is in progress, black means pending, green denotes successful completion, and red signifies failures.}
    \label{fig:chaining}
\end{figure}

An operator rarely performs only a single tactic. Rather, he or she deploys multiple tactics in a specific order to advance particular mission objectives. The coordination of multiple tactics is therefore an important feature that we support via tactic chaining. As illustrated in \cref{fig:chaining}, an operator is able to link two tactic icons together via a sketch gesture. The combination of multiple links forms a directed acyclic graph. When an operator issues the root node tactic, C2 submits the entire graph structure to PyC2 for further processing. By default, children tactics execute only after their parent tactics successfully complete. In other words, parent node status propagates to children nodes, and if any parent tactic fails, the child tactic similarly fails. However, this propagation behavior can be modified via the use of logic gate tactic nodes. 

Logic gate tactic nodes are identical to standard tactics in implementation. One difference is that they override the default propagation behavior (see \cref{sec:rise_tactics}). PyC2 presently implements negation, conjunction, and disjunction. The negation gate inverts the status of its parent tactic, which can be useful for planning contingencies. Disjunction gates output a success response if any parent tactic is successful, whereas conjunction gates requires that all parent tactics are successful. Logic gate tactic nodes enable an operator to express sophisticated mission plans in sketch input form. 

\subsubsection{Tactic Parameters}
Tactics require input data that specify their precise behavior on invocation, \emph{e.g.}, a persistent surveillance loitering altitude  or safe building standoff distance. Most parameters are typically determined empirically or resolved through automation, though during development or in special cases, an operator may need to specify alternative values. For this reason, certain input data can be modified by an operator through the popup window interface. Data types we found to be suitable for the popup window include numeric, text, and boolean input. However, other data types are more easily described via sketch-based interactions. Throughout the course of the DARPA OFFSET program in which we developed and integrated numerous primitives and tactics, we encountered a consistent demand for only three sketch interaction types: selection groups, points, and polylines.

\begin{itemize}
\item \textbf{Selection Groups}: It is sometimes necessary, especially for developers and swarm health engineers, to be able to select individual agents or a group of agents upon which subsequent commands are assigned. To support this operation, we implement a lasso selection technique, whereby the operator may draw a stroke around those objects he or she wishes to group. The operator may continuously lasso select objects until satisfied. Those objects within each new stroke are combined with previous groups using disjunctive union logic, thereby allowing deselection. Any tactic issued by the operator will then be restricted to just those agents in the final selection group. Should the specified tactic generate children tactics, those tactics too will adhere to the same restriction. 

\item \textbf{Points}: Points are geospatial location markers that specify a certain position within the environment, \emph{e.g.}, points of interest, rendezvous points, and breach points. Points are mapped to gestures such that when drawn, the associated point appears at the gesture's centroid within the scene view. The point can then be drag around to a precise location in three-dimensional space, though through customization, axes can be locked to reduce error. For instance, it is common practice to lock points to the ground plane. 

\item \textbf{Polylines}: A polyline is a curve specified by an ordered sequence of geospatial locations that may be open, or closed to form a loop. Sketch strokes serve as the mechanism an operator uses to input polyline data, as strokes are a natural and fluid form of communication. One can therefore use polylines to specify trajectories including preferred routes and boundaries (no-go zones, explore areas, and deployment zones). An operator enters a gesture into the scene view to select a polyline type. The operator then inputs a stroke that defines the initial polyline, after which he or she can modify by sketching directly onto the polyline. The modification stroke acts as a magnetic tool that pulls the polyline in the direction of the cursor. When the polyline type is a closed loop, self loops and intersections are removed. Example routes and boundaries are shown in \cref{fig:fx4_mission_plan}.
\end{itemize}

To aid with perception and comprehension, tactic developers may customize point and polyline types with unique gesture invocation and rendering properties. See \cref{sec:rise_tactics} for additional information.

\subsection{Situation Awareness}
C2 provides several tools to aid in the perception and comprehension of swarm intelligence. Swarm commanders can use this information to estimate future status and decide next steps.  

\begin{figure}
    \centering
    \includegraphics[width=.8\textwidth]{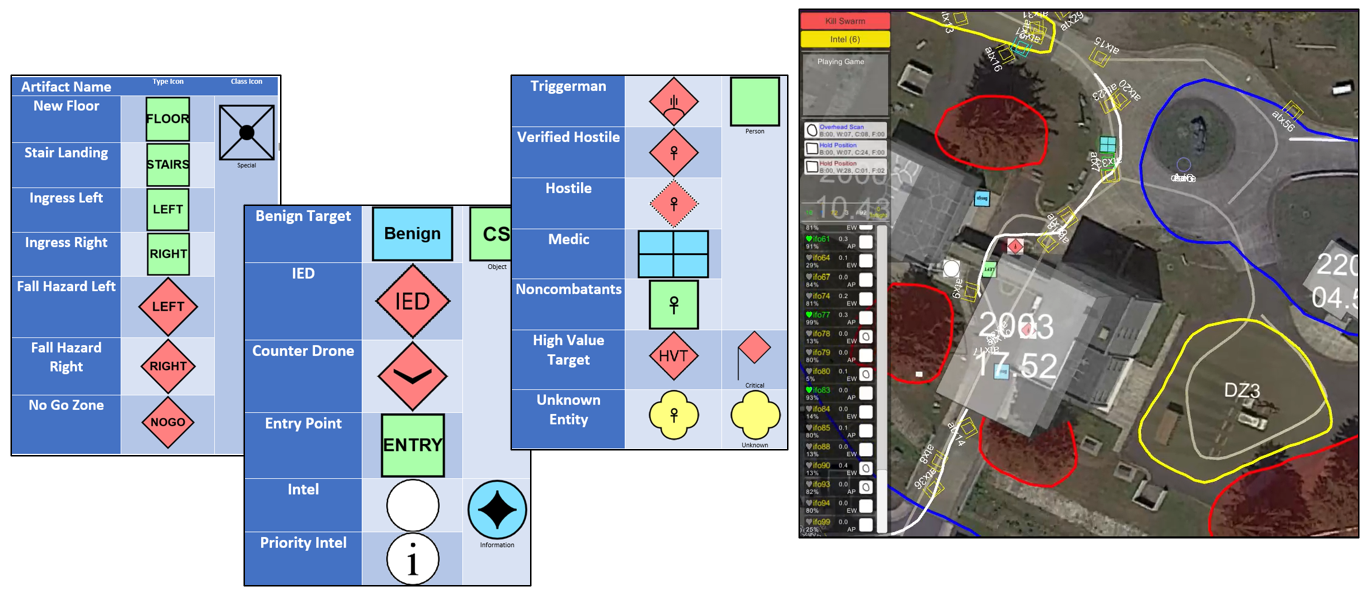}~
    \captionof{figure}{Example iconography used to communicate artifact information during the DARPA OFFSET Joint Base Lewis-McChord field experiment.}
    \label{fig:sa_icons}
    ~\\

    \centering
    \includegraphics[width=.3\textwidth]{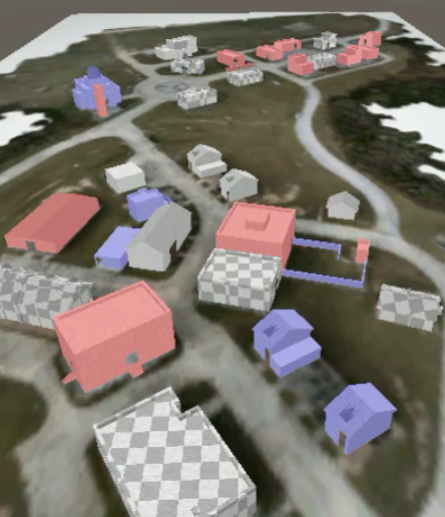}
    \captionof{figure}{Buildings are color coded in accordance with swarm gathered intelligence information. We render buildings in a checkerboard pattern when the building is known to exist via a priori knowledge, which translations to solid color when discovered by the swarm. A building known to contain threats inside is represented by a red shade, and buildings known to contain intelligence information as blue.}
    \label{fig:building_intel}
    ~\\~\\
    \centering
    \includegraphics[width=.445\textwidth]{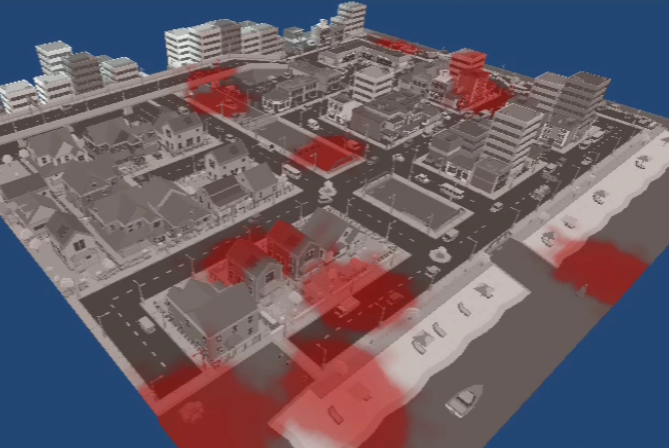}~
    \includegraphics[width=.43\textwidth]{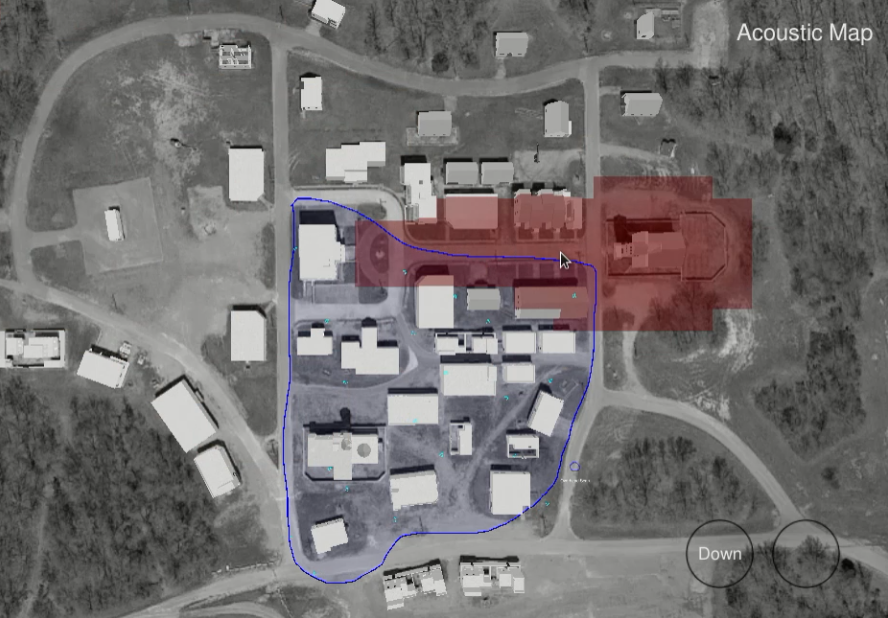}
    \captionof{figure}{Example grid data visualizations. Left: Threat distribution map. Right: Accoustic zone map correlated with the opreato's cursor position. }
    \label{fig:grid_data}
\end{figure}

\begin{figure}[t]
    \centering
    \includegraphics[width=.31\textwidth]{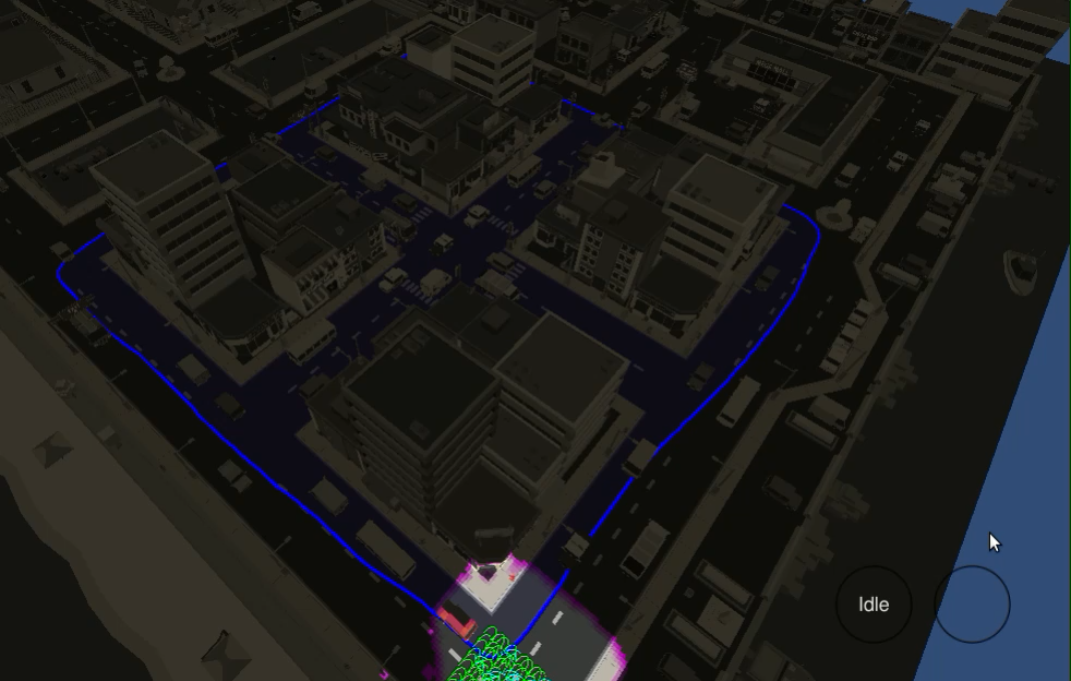}~
    \includegraphics[width=.32\textwidth]{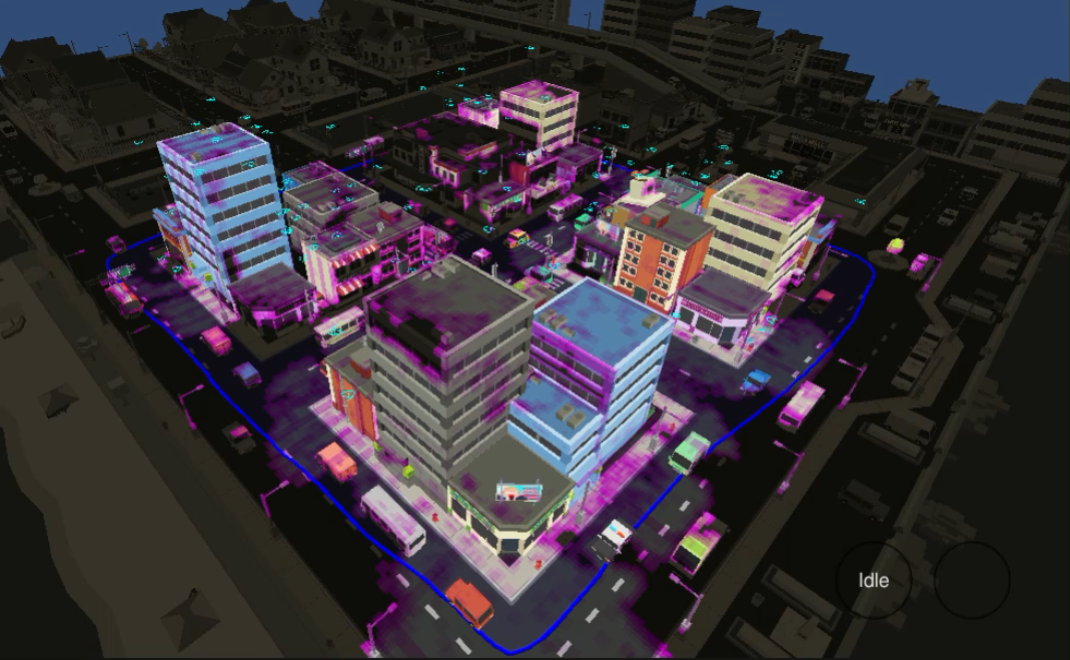}~
    \includegraphics[width=.32\textwidth]{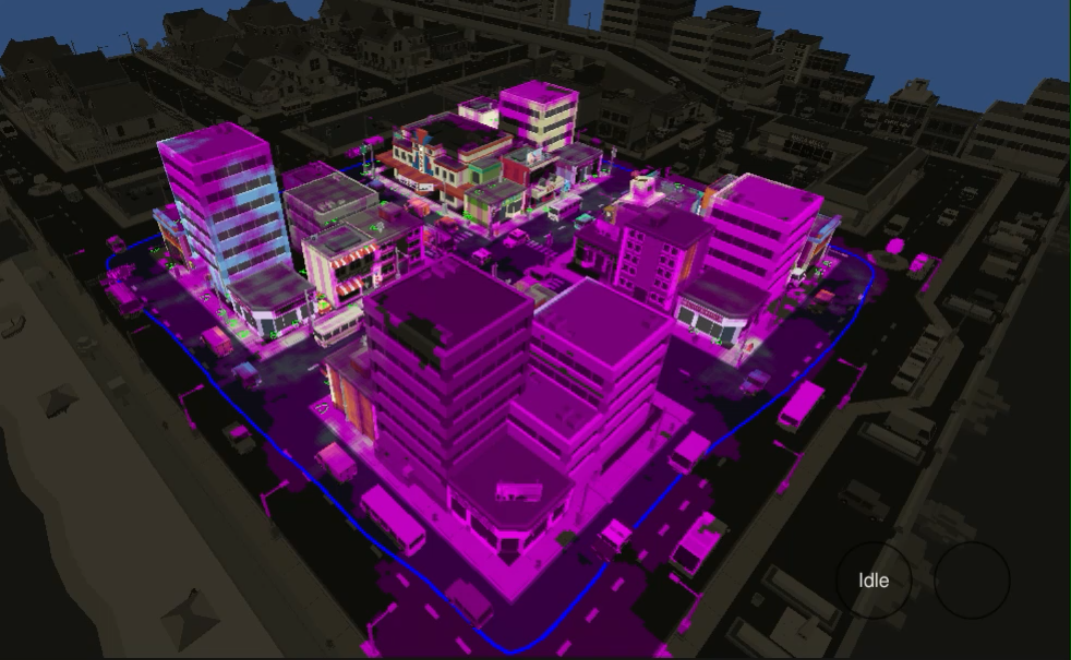}
    \captionof{figure}{Temporal coverage visualization as time progresses from left to right. Unexamined areas initially render in a dark sepia tone. As agents navigation through the environment, explored scene geometry comes into full color. Finally, as time progresses and we lose situation awareness in particular areas, color fades to magenta.}
    \label{fig:coverage}
    ~\\~\\
    \centering
    \includegraphics[width=.75\textwidth]{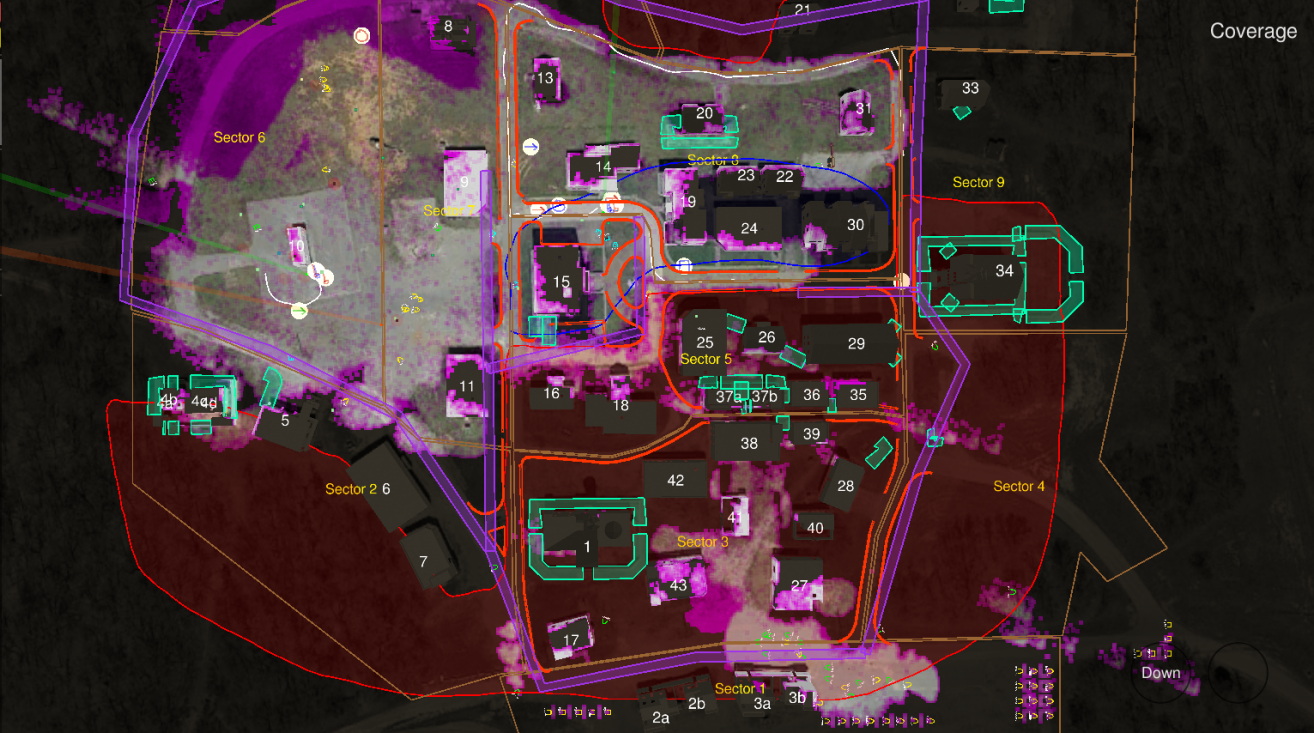}
    \captionof{figure}{Temporal coverage visualization captured in the course of a joint Northrop Grumman and BBN exercise during the DARPA OFFSET sixth field experiment at Fort Campbell. Dark areas have not yet been examined. Full color areas have current situation awareness. Magenta-colored areas have been previously examined, but the situation data is now stale.}
    \label{fig:joint_coverage}
\end{figure}

\subsubsection{Artifacts}
\label{sec:artifacts}
We define an artifact as an object of interest. C2 visualizes relevant artifacts detected by the swarm using military iconography as shown in \cref{fig:sa_icons}. We use standardized iconography to aid with comprehension because of their widespread use and ability to communicate essential information. Two DCRI levels \cite{self2005acquisition} are supported, where C2 renders recognized artifacts using general class icons and identified artifacts using specific icons, \emph{e.g.}, person versus noncombat, medic, or hostile. To gather more information, an operator may view an agent's video stream or command the swarm to surveil the artifact. We further render a red sphere around artifacts that pose a threat. Once neutralized, we remove the sphere.

\subsubsection{Intelligence Information}
As an alternative to the tactic and agent list view, a commander may opt to use our intelligence messages view. In this mode, swarm intelligence messages are listed in the left information panel. Unread messages are organized by priority level and require commander feedback to ensure they have been read. Messages referencing scene view data include hyperlinks that move the camera to the associated location. For example, ``HVT reported in building 10" embeds a clickable link that moves the camera over the building.

\subsubsection{Building Visualizations}

C2 visualizes swarm intelligence gathered on building state information as shown in \cref{fig:building_intel}. Specifically, we render a priori known building geometry in a checkerboard pattern. This indicates that the swarm expects to find a building, although its status is unconfirmed. Once the swarm verifies its state, we render the building solid. Additionally, buildings that house threats or intelligence are shaded with an oscillating red or blue tint, respectfully, and a quick glance of the scene view quickly reveals areas where additional attention may be required. 

\subsubsection{Grid Data Visualization}

Swarms may track data that is well represented as a grid overlay. For this reason, we provide a mechanism by which swarm entities (any networked components including agents or third-party programs) may communicate custom grid data to C2. \cref{fig:grid_data} illustrates two examples born out of our collaborations with Michigan Tech Research Institute. The left grid visualization is a probability threat distribution where denser probabilities (brighter red) correlate with larger threats. Initially, initial uniform distribution evolves into a multimodal distribution as the swarm gathers intelligence and red forces are isolated. The right grid visualization illustrates an acoustic map. This is the area where deceptive firefight sounds would be heard if projected from under the operator's cursor. One can see how grid data visualizations may be useful in several contexts; for example, instead of visualizing acoustic zone data, the swarm could report on network communication ranges, among other data. 

\subsubsection{Temporal Coverage}

In order to visualize coverage over time, RISE C2 tracks agents as they move through the environment. Specifically, for each agent, across each frame, we cast rays from each camera into the virtual environment. The resulting collection of hit points yield a coverage estimate for the given frame. All hit points are then cached in a GPU-based voxel hash table representation of the environment that a custom temporal coverage shader uses to render the environment. Each entry stores a voxel ID and timestamp. When the said shader renders a fragment, it queries the hash table to determine when the voxel was last seen by an agent. As shown in \cref{fig:coverage}, if the associated voxel is absent from the hash table, we assume the voxel has never been observed, and so we render the fragment in a dark sepia tone. Otherwise, the fragment is colored based on a linear interpolation between full color to magenta, based on the time passed since the voxel was observed\footnote{We fade to full magenta after twenty minutes.}.

This visualization has several practical applications. Practitioners, for example, can use temporal coverage data to quickly validate reconnaissance and persistent surveillance tactics work according to expectation. Swarm operators can maintain situation awareness to verify where within the mission area knowledge is absent or outdated.  \cref{fig:joint_coverage} shows the temporal coverage visualization in use at the DARPA OFFSET sixth field experiment during a Northrop Grumman and BBN joint run scenario. The swarm commander was able to quickly measure progress against mission objectives and modify the mission plan accordingly to increase coverage. 

\subsection{Supported Environments}
C2 supports three environment types of varying complexity. The simplest environment we support is an image based top-down map view that C2 loads from a public repository based on given GPS coordinates. This environment is useful for ad hoc testing that may occur during travel or when 3D models are unavailable. Second, C2 can load GeoJSON encoded data, which can be useful for randomized as well as custom scenario testing. However, when more complexity is required, developers can export custom Unity scenes that C2 can load. Finally, a number of custom scenes are also built into RISE, namely the Joint Base Lewis-McChord, Camp Shelby, and Fort Campbell CACTF sites.

\section{Tactics Development Interface}
\label{sec:rise_tactics}

PyC2 provides a straightforward interface for rapid tactic development. To implement a new tactic, one derives a new class from the PyC2 Tactic base class and specifies a tactic definition as shown in \cref{listing:tactic_definition}. All information required by C2 to invoke the tactic is included in this definition, including the command gesture, required input parameters, as well as human-readable descriptions of the tactic and its parameters. When PyC2 discovers a new C2 device via its heartbeat, we transmit all tactic definitions over the network. When C2 receives a new tactic definition, it will retrain the gesture recognizer (collecting new samples if required) so that an operator can immediately invoke the tactics. The associated tactic popup window (see \cref{sec:tactics_ui}) is automatically populated with numeric, boolean, and text parameter data entries. This design enables one to focus on core tactic logic without having to concern oneself with user interface integration, \emph{e.g.}, GUI layout design, arrangement, and embeddings within hierarchical menus.  

\begin{figure}

    \input{listings/poi.py}
    \captionof{lstlisting}{Dictionary entry for a point of interest type in PyC2. The type, id, gesture command name, and rendering properties of this custom point type are automatically encoded and transmitted over the network whenever PyC2 recieves new C2 heartbeat. }
    \label{listing:custom_params}
    ~\\
    
    \input{listings/tactic_definitions.py}
    \captionof{lstlisting}{Tactic definition for Overhead Scan, which subdivides a given area into a number of smaller regions based on the input parameters and generates one primitive move-to command for each region. A tactic definition comprises a tactic name, short human-readable description, gesture, context (which can be an explore area \emph{or} a sector in this example), and parameter list where each entry specifies a human-readable name, description, data type, and default value. Each tactic definition is encoded and broadcast over the network whenever PyC2 receives a new C2 heartbeat.}
    \label{listing:tactic_definition}

\end{figure}
 
When C2 invokes a tactic, PyC2 receives a network message with all non-sketch parameter data encoded into it. PyC2 parses the message, verifies data validity, checks for errors, and resolves context. That is, since one can write a tactic that works under different contexts, PyC2 will additionally examine all available context options and select that which is closest to the gesture's position. For example, our overhead scan tactic operates over either an explore area or sector. If the closest explore area polyline point is less than that of the closest sector point, Pyc2 will select the explore area. Errors are reported back to C2, otherwise, all parameter data including lasso-selected agents are copied into a task object. PyC2 then instantiates the associated tactic class, passing in the task object as its only parameter. In this way, all boilerplate processing is handled automatically, and again one can focus primarily on tactic logic. 

Any tactics may invoke other tactics or robot primitives, each of which are handled as individual children tasks. Therefore, each child task encodes a tactic or primitive name, along with its required input parameter values. PyC2 implements a wrapper task class for each primitive that contains parameter data selection, manipulation, and transformation routines specific to the primitive. In most cases, specialized logic is not required because the encoded primitive is a direct copy of key-value (parameter-value) pairs. However, in more complex cases, primitive encoding may be situation-dependent. Each child task is queued via the tactic base class's queuing mechanism, and PyC2's task management system handles task bidding, assignment, and invocation as described in \cref{sec:bidding}. Appendix A presents a complete tactic example. In the remainder of this section, we discuss how tactic behaviors can be customized and what tools PyC2 provides to enable efficient tactic development. 

\subsection{Tactic Execution Customization}
The simplest tactics are those that generate children tasks and nothing more. Our tactic base class provides default behaviors for all task processing stages and error handling. However, there are often situations where one must override default tactic behaviors in order to extend its capabilities. A tactic may wish to issue several primitives to a particular agent over time, such as to move to a particular location and then secure a threat. In another case, a tactic may encode certain parameters based on which agent won a bid. Or a tactic may provide custom error handling logic, and so forth. For this reason, one may override the default bid completion, bidding failure, task completion, task cancelled, task failure, and tactic completion callback handlers. 

One may also override a tactic's prerequisite check. As described in \cref{sec:tactics_ui}, an operator may chain commands together using gate-like tactic nodes. By default, we only invoke a child tactic after all parent tactics on which the child depends have completed successfully. Otherwise, we fail the child tactic. When this behavior is inappropriate, one can employ a custom prerequisite check, as we have done with the conjunction, disjunction, and negation tactics. 

\subsubsection{Sketch Input Customization}

As illustrated in \cref{listing:custom_params}, one may generate new custom sketch input types based on C2's generic point and polyline classes. In defining a new type, one specifies its underlying type name, unique identifier, command gesture, and render properties. Once defined, custom parameter types can be used as context that PyC2 resolves when a tactic is invoked. Two benefits arise from this approach. First, by having unique parameters types, users can more easily distinguish their presence and purpose in C2. Second, it helps resolve ambiguity when a gesture is close to multiple different sketch types---there are fewer chances for operator error.

Custom parameters are also quick to implement and put into practice. For example, during our final field experiment at Fort Campbell, we decided that agents should return to a recovery point to facilitate field operations. Our safe land tactic previously landed agents at unobstructed locations when their battery level fell below a certain threshold. Thus, we created a new input parameter type and updated our tactic such that if a recovery point was specified somewhere within the scene, we would instead command agents to a location near the recovery point. This tactic was implemented and validated in simulation within two hours and fielded the same day using more than seventy air agents.

\subsection{The Tactic Development Toolkit}

PyC2 comes equipped with a variety of features and utilities that enable intelligent tactic design. The most relevant tools follow:

\begin{itemize}
\item \textbf{Scene Geometry}: C2 transmits a priori scene geometry to PyC2 during initialization. This information includes GeoJSON encoded building label, boundary, wall, and height data. Tactics can use this information to inform reconnaissance, surveillance, and other maneuvers that interact with buildings. Tactics can also use building data as context, \emph{e.g.}, to select which building an operator intends to scan. 

\item \textbf{Artifact Support}: PyC2 tracks artifacts with which the swarm may interact, along with basic state information such as whether the artifact has been recognized or detected, and whether the artifact is a threat or has been neutralized. Like sketch input and buildings, tactics may use artifacts as context. Secure artifact is one example tactic where the operator may manually command the swarm to neutralize a threat. In this case, the tactic chooses that artifact which is closest to the operator's secure tactic gesture.

\item \textbf{Path Planner}: We include a jump point search (JPS) \cite{harabor2011online} based path planner implemented as a multilevel grid. The underlying grid structure encodes known obstacles, including buildings and no-go zone sketches. A tactic can use the JPS planner to recommend optimal paths through the environment for air and ground agents. An agent who receives this path information as a primitive input parameter may optionally follow the prescribed path to a specified destination using its own local path planner to avoid dynamic obstacles while coordinating airspace maneuvers with other agents. 

\item \textbf{Sketch Input}: All operator sketch parameters are transmitted over the network when they are created and as they are modified. PyC2 stores each parameter in a local database with which tactics may interact. Utilities for querying sketches based on type and distance, analyzing geometric properties such as length and winding, and transforming sketches by spatial resampling and line simplification are included. These tools enable developers to efficiently analyze context and generate work based on free form sketch-based user input. 

\item \textbf{Sketch Generation}: We also provide mechanism that enable developers to generate and publish new sketch data. In advance of the DARPA OFFSET sixth field experiment, the Naval Information Warfare Center Pacific (NIWC) provided curb, wall, and power line footprint data. We encoded this information as custom sketch data that PyC2 treated as no-go zones. As such, our path planner was able to route around these obstacles.

\end{itemize}

\section{Primitives Development Interface}
\label{sec:rise_primitives}

All tactics are made up of primitives, just as all words are made up of letters. Taken as a group, primitives can spell out a vastly more complex action. In RISE, we define primitives as basic actions, individual to a specific agent. They are combined into tactics, which encompass the entire swarm (see \cref{sec:rise_tactics}). A single tactic may involve running multiple primitives on the same agent, or even running different primitives on separate agents simultaneously.

Primitives combine high-level individual agent logic with low-level robotic sensing and control algorithms to effect desired behaviors.

\begin{figure}
    \centering
    \includegraphics[width=.925\textwidth]{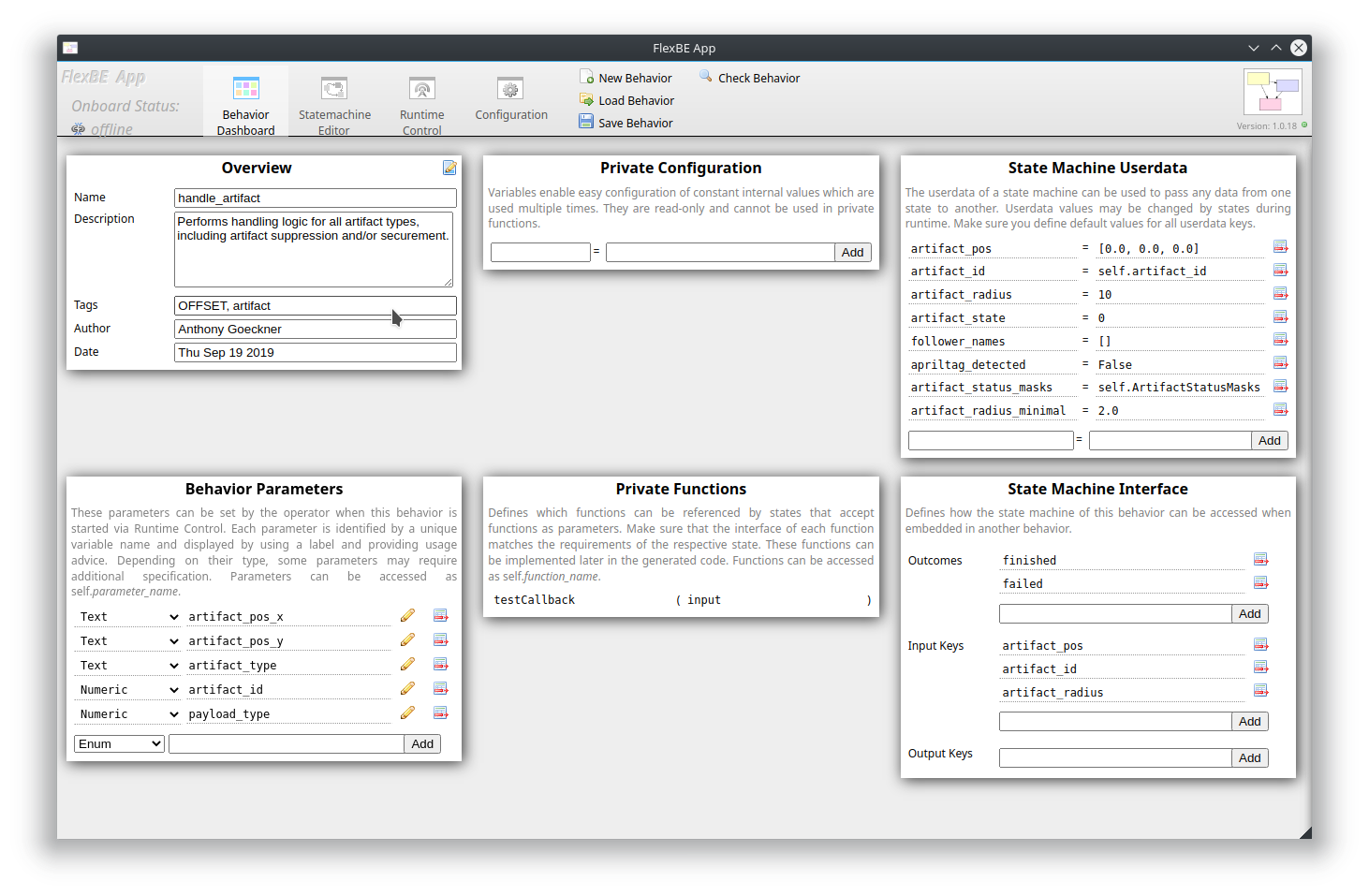}~\\~\\
    \includegraphics[width=1\textwidth]{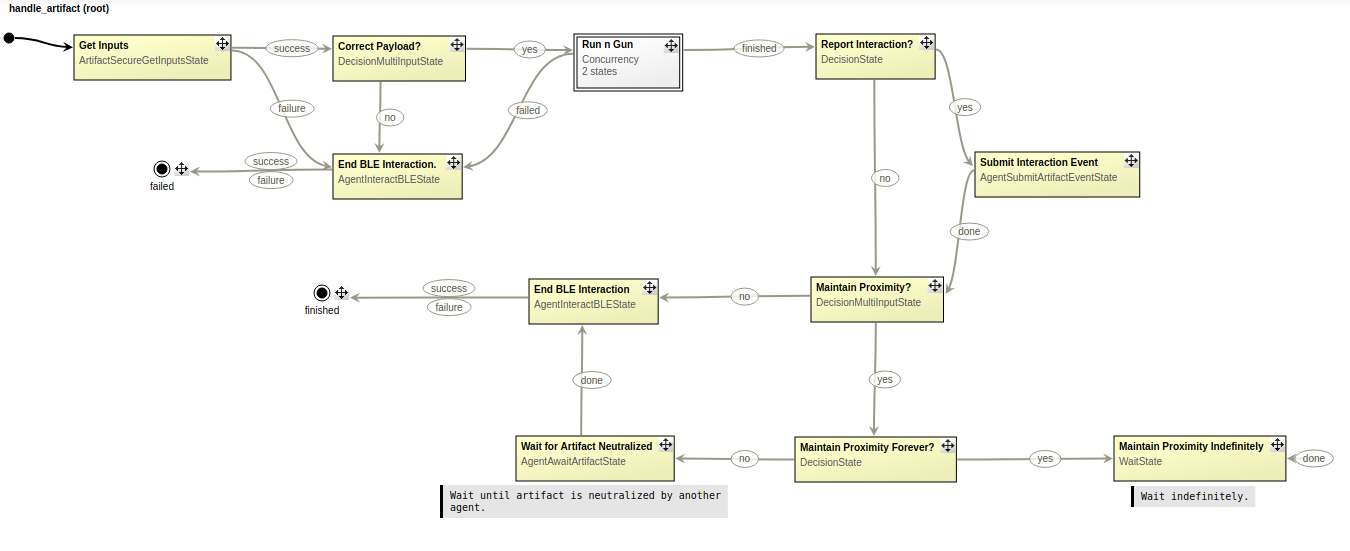}
    \captionof{figure}{Images of the FlexBE App utilized to create agent primitives. At bottom, a high-level mission logic primitive handles interaction with artifacts found in the environment.}
    \label{fig:FlexBEApp}
\end{figure}

We provide a highly modular system for integrating new primitives by making use of the FlexBE behavior engine. FlexBE\footnote{http://philserver.bplaced.net/fbe/index.php} is a ROS package providing state machine creation and execution capabilities, with simple ROS integration and an easy-to-use visual programming interface \cite{schillinger2016FlexBE}. See \cref{fig:FlexBEApp} for an example of primitive creation using the FlexBE state machine editor. These FlexBE state machines are tightly interwoven with the basic operation of each agent, providing many core functionalities. This is important, since it allows primitives to make use of existing agent behaviors using FlexBE's state machine nesting and concurrency features. Rather than rewriting code to, for example, move the agent to a waypoint, the existing waypoint movement primitive may be nested in the new primitive. In this way, we are able to quickly integrate new primitives with the agent codebase and unlock new behaviors. Each agent's unique set of primitives are contained within platform specific FlexBE State Libraries.

While the agent primitive system is designed to allow for easy integration of third-party primitives, we provide a number of primitives with RISE. These include basic movement, mission interaction, building breach and entry, building interior exploration, and more.

\section{Algorithms Development Interface}
\label{sec:rise_algorithms} 

Low-level robotic algorithms form the basis of agent capabilities. We implement many of these from the ROS ecosystem, but a number are our own creations for RISE. The majority of these algorithms are implemented as ROS nodes on each agent's main compute board, although some reside on vehicle flight control units (FCUs), which are based on the Pixhawk standard. 
\subsection{Sensing}

Each agent in the swarm collects data about its local environment. Depending on the class of agent, different sensors might be available. For example, IFO vehicles have a forward-facing stereo camera, downward-facing camera, downward-facing single-point LIDAR, Global Positioning System (GPS) receiver, magnetometer, altimeter, and accelerometer. ATX ground vehicles do not have downward-facing sensors, but instead have a 360-degree circular LIDAR.

We implement drivers to translate raw sensor data into useable ROS messages, which are made available to other ROS nodes running on the agent.Processing of magnetometer, altimeter, accelerometer, and GPS data occurs on the vehicles' FCUs, using both the PX4 and Ardupilot systems depending on the platform. However, processing of camera and LIDAR data occurs on the main compute board of the vehicle and is implemented in ROS nodes. We use a simple speckle filter to remove strands of grass and other non-obstacles from the LIDAR data. For stereo cameras, the driver calculates a disparity image, which it makes available to other ROS nodes for perception.

\subsection{Perception}

To implement visual obstacle detection on ground vehicles, we take advantage of our stereo cameras to perform binocular disparity-based detection. Our implementation is based on the work of \citeA{vboatsInspiration} and is highly effective at detecting obstacles while ignoring environmental artifacts such as stems of grass which do not pose a collision risk to the vehicle.

For aerial vehicles, we implement the PX4 Avoidance system, which detects and then avoids obstacles using the forward-facing depth camera (see \cref{sec:motionPlanning} below) \cite{px4Avoidance}.

In addition to local data, our agents may also use the telemetry information of other vehicles to avoid collisions. Telemetry information is shared between all vehicles using the MANET. Future work may extend this feature to include prediction and make it more useful.
\subsection{Motion Planning and Control}
\label{sec:motionPlanning}

We use data described in the Sensing and Perception sections above to form two costmaps for each ground agent. A two-dimensional local costmap stores detailed information about obstacles close to the agent. This costmap features a decay function such that obstacles not seen for a certain period of time will be removed from the costmap. This allows for accurate tracking of dynamic obstacles. We also maintain a second, global costmap which is used to store obstacle information for the entire mission area. This costmap has no decay function, meaning that obstacles will remain in the costmap until updated by new local sensor data. This two-dimensional global costmap is used for global path planning.

Ground vehicle movement planning is performed by ROS's move\_base system. This involves a global planner for long-distance goals and a local planner which is reactive to obstacles blocking the global path. The move\_base system also handles high-level actuator control, issuing velocity commands to the FCU.

For IFO vehicles, PX4 Avoidance controls all obstacle avoidance using its own internal costmap. This involves direct control of vehicle velocity and heading. We have not made significant modifications to the publicly-available PX4 Avoidance library, other than to allow the avoidance function to be switched off temporarily. This allows agent primitives to control the vehicle heading for in-place rotation, something impossible with standard PX4 Avoidance.

\subsection{Agent Logic}

To enable agent primitives that further mission objectives, we provide algorithms for high-level agent logic. These include automatic detection and recognition of AprilTags using the ROS AprilTag library \cite{wang2016iros}, interaction with the environment and mission scenario using Bluetooth Low-Energy beacons, and more. See \cref{sec:field_experimentation} for more information on how we enable agents to interact with the environment.

\section{Live, Virtual, \& Constructive}
\label{sec:Simulation}

We treat agents as logical entities whose underlying components may be centralized or distributed, as well as physical or virtual. This is possible because ROS implements a publish–subscribe messaging pattern over network-based named buses for inter-process communication (see \cref{sec:networking}). As such, ROS packages and libraries that use ROS are self-contained in a way that enables robotics modules to run as independent processes. By exploiting this feature, it is possible for one module to publish sensor data while another module analyzes the data and generates locomotion commands that a third module actuates; and although all processes belong to the same logical entity, they may in fact be distributed across disparate systems. We therefore think of a logical entity as being the set of all namespace topics that effectuate a robot instance. 

As OFFSET progressed, it became apparent we would need to design a method to allow for varying levels of fidelity within our simulation environment. Typical robotic simulation environments, such as Gazebo \cite{Aguero-2015-VRC}, provide a robust development environment, but typically only for single agent use. While we initially attempted to utilize Gazebo for swarm development, it became clear that the system would be computationally constrained at around ten platforms. We observed similar issues across other various high level fidelity simulation environments. This pushed us to develop the concept that we call in our system Fluid Fidelity (\cref{sec:fluid-fidelity}). 

Another implementation choice that we determined to be critical was the ability to have virtual agents working with physical ones. From a logistical perspective alone, it often becomes unfeasible to transport hundreds of platforms to various real world environments for intermittent testing. So while we wanted to test out a new swarm tactic on physical platforms, we needed a way to have a few physical agents work alongside virtual agents to quickly evaluate new capabilities without bringing the whole swarm with us. Due to the utilization of a singular application for simulation and command and control (\cref{sec:architecture}) as well as our ROS interfacing (\cref{sec:ROS}), we were able to integrate this capability.

\subsection{High Fidelity Simulation}
\label{sec:level3}
For our high fidelity simulation environment, we took inspiration from many industry standard simulations, such as Gazebo and AirSim \cite{shah2017airsim}. Still, our high fidelity simulation does have several characteristics that cause it to vary from many common simulations. Two of the key differentiators is that there is no active physics in the high fidelity simulation and agent bodies consist of simple relative sized cube objects. In place of a physics environment, each platform type has a geometry based PID controller to actuate motion. Agents send actuation commands via ROS Twist messages or setpoint position commands just like on the real platforms. Comparable to other simulators, individual agents receive sensory information from ROS messages. All ROS messages pertaining to sensor sources that would typically be received from a physical sensor are instead generated from the Unity Game Environment. The ROS nodes simulated are TF information, MAVROS data, LIDAR, RGB, and depth, among others. Each of the sensor streams per agent are organized through ROS namespaces. All data are generated as ROS 2 messages, but since the agent codebases expect this information as ROS messaging, we utilized the ROS 2 bridge to translate between them. This ROS 2 translation and a centralized ROS master is handled through an application we call the ROS Simulation Server. The ROS Simulation Server allows us to run multiple high fidelity agents on a single system and has pairing with the simulation system to allow for Fluid Fidelity (\cref{sec:fluid-fidelity}). Although Unity now has official support for ROS integration \footnote{https://github.com/Unity-Technologies/ROS-TCP-Connector}, we needed ROS support before this system existed, and we instead implemented our own method utilizing our ZMQ message system (see \cref{sec:networking}).

\begin{figure}
    \centering
    \includegraphics[width=.5\textwidth]{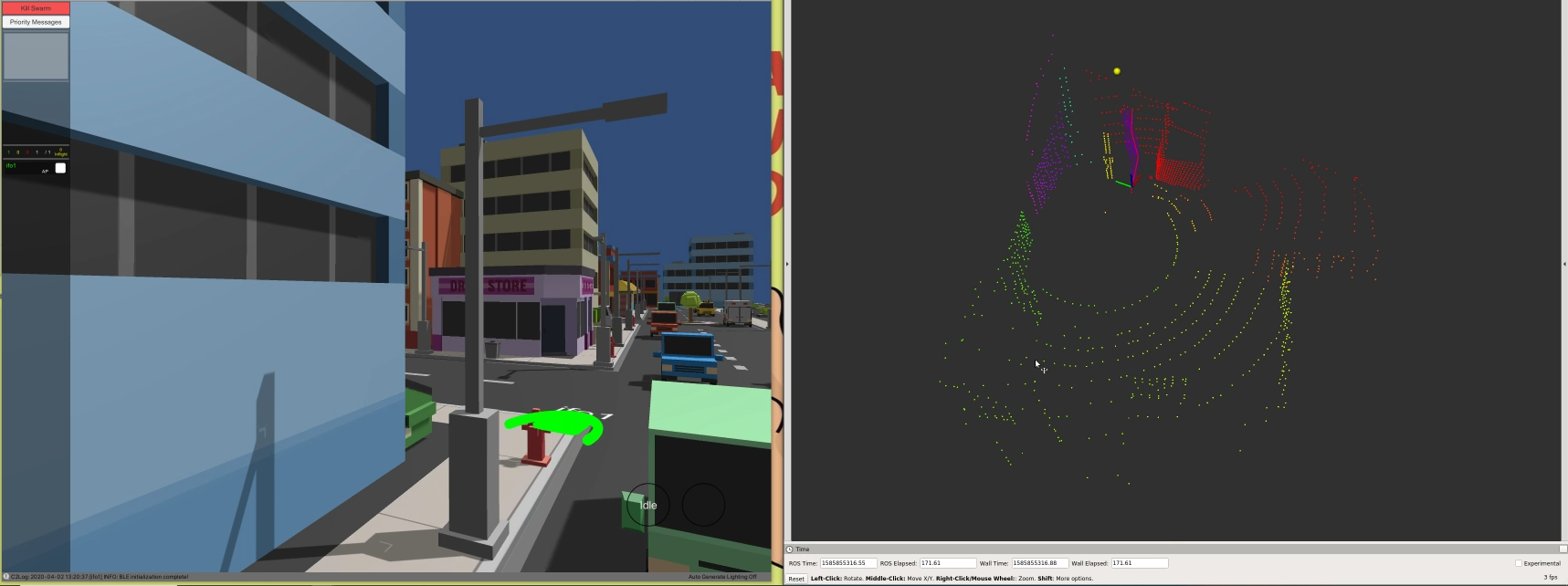}~
    \includegraphics[width=.5\textwidth]{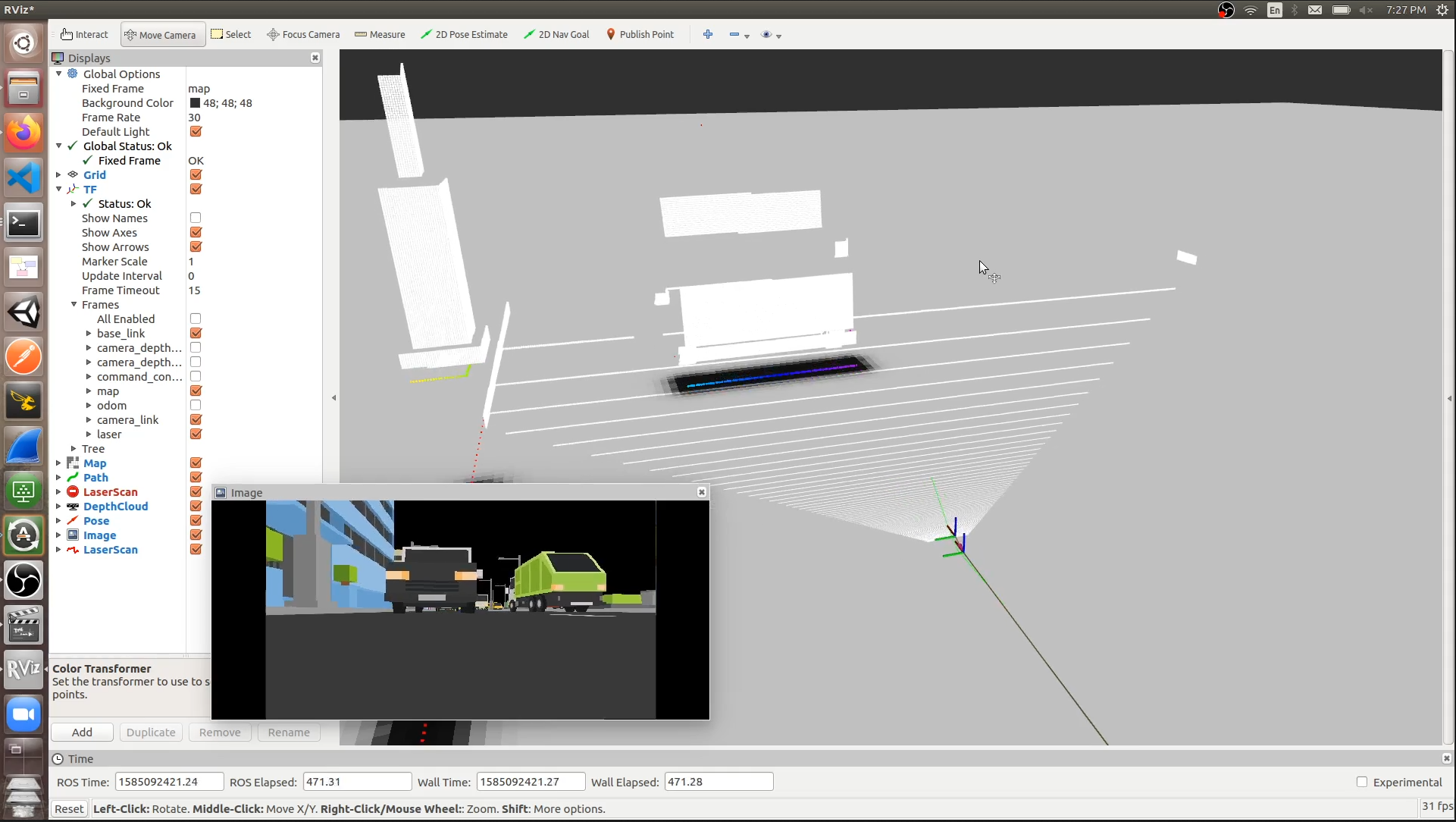}~
    \captionof{figure}{Left: UAV navigating around a street lamp in simulation and its associated sensor feed as visualized in RViz. Right: UGV navigating a simulation environment and its associated depth camera sensor feed, also as it appears in RViz.}
    \label{fig:sim}
\end{figure}

With the data streams being facilitated, actual agent primitives and algorithms are tested through the simulated Docker containers. The underlying codebase is exactly identical to that which the real swarm agents are running when utilizing high fidelity simulation. There are only a few ROS nodes that need minor adjustments, which is determined via a runtime ROS parameter to make networking configuration changes. ROS nodes that interact directly with the hardware sensors are disabled, as they are no longer needed given the data is being published from Unity. Each individual agent has a corresponding Docker instance, differentiated only by the hostname of the container and the mounted file directory codebase. \cref{fig:sim} showcases both a high fidelity simulated IFO and ATX utilizing this system and rendering their sensor streams in RViz as their navigation stacks run.

\subsection{Low Fidelity Simulation}
\label{sec:level1}
Our Low Fidelity simulation capability is vastly different from standard robotic simulators. There are no robotic algorithms running in this configuration, and the whole concept revolves around mimicry. We found when using our high fidelity simulator that around a maximum of ten agents was all that could be achieved on a single machine. While distributing the computational load across multiple systems is possible with our high fidelity simulation, we found it to be not feasible to achieve the scale we were looking for, especially since we lacked cloud compute resources. Tactic development is also sometimes performed rapidly in the field, where additional compute resources or an internet connection are not always guaranteed. These limitations led to the creation of our low fidelity simulator. In this setup, there are no simulated agent Docker containers running, and the configuration consists solely of the Unity Base and PyC2 application (see \cref{fig:architecture}). Unity is rendering the individual game objects for the simulated agents, similar to the high fidelity simulator, but it is not generating any simulated sensor sources except for positional information. In place of generating the raw sensor information, there are components within our Unity environment that are performing rough estimations for things such as agent attrition or agent detection, essentially mimicking the response of the full sensor pipeline in the high fidelity simulation. Another aspect is PyC2 mimicking agent primitives. Since there are no actual agent primitives or algorithms running, PyC2 is in charge of running agent instances that run mock algorithms that simulate the relative agent responses for certain tactic executions. These mock algorithms still send commands via the setpoint position or twist API just like the high fidelity simulation environment. It would be unmaintainable to have PyC2 have mock duplicates of every single agent primitive, so it's important to note that only the most common primitives have direct low fidelity mappings. To handle the rest, that is where the concept of Fluid Fidelity comes in (see \cref{sec:fluid-fidelity}).

\begin{figure}
    \centering
    \includegraphics[width=.8\textwidth]{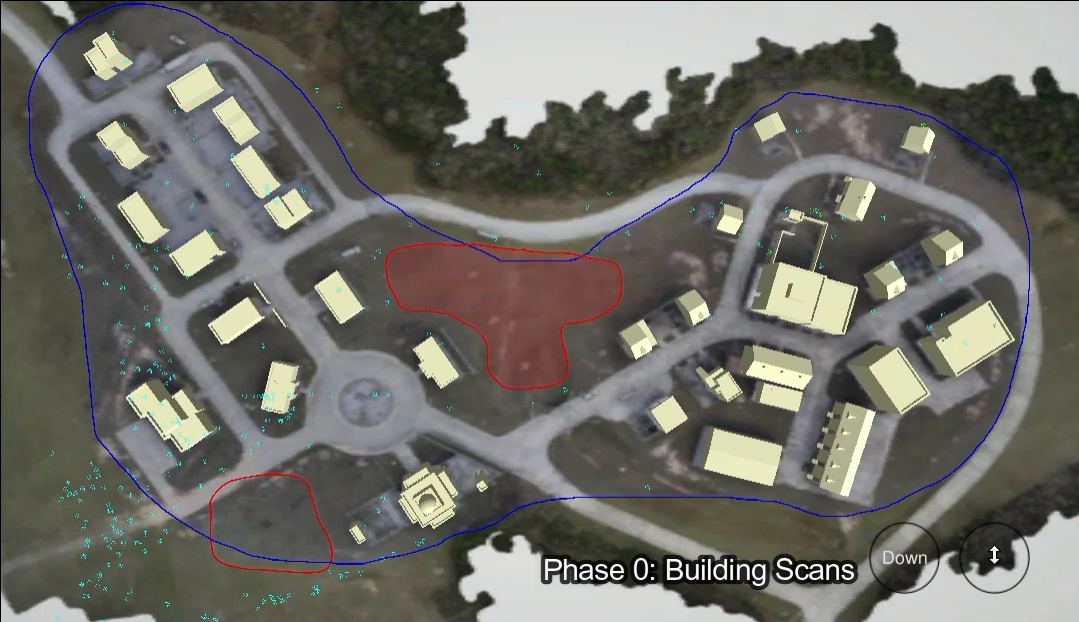}
    \captionof{figure}{Low fidelity simulation running over 100 agents in the Camp Shelby, MS environment, using a single machine.}
    \label{fig:level1}
\end{figure}

\cref{fig:level1} showcases 100+ low fidelity agents running on a single machine. It is worth noting that regardless of fidelity level or whether the agent is real or simulated, the interface to the end user is always the same. This low fidelity simulation is not particularly useful when doing agent specific development, but is very powerful when performing tactic development (see \cref{sec:rise_tactics}). Without the creation of low fidelity, not nearly as many tactics would exist. This lower fidelity also allows for further concepts such as faster than real-time performance, which allows for fast tactic analysis and opens the door to concepts such as optimal course of action generation.

\subsection{Fluid Fidelity}
\label{sec:fluid-fidelity}
As low fidelity simulated agents are unable to simulate some primitives required by certain tactics, we need high fidelity simulated agents to be able to step in and execute those primitives. As described in \cref{sec:level1}, low fidelity agents merely have mock algorithms that mimic actual agent primitives, but the collection is not complete. This is where the concept of fluid fidelity comes in. As an operator is running a mission, there may arise situations where tactics are called that low fidelity agents cannot perform. We needed a way to perform a seamless transition to convert a low fidelity agent to a high fidelity agent, so that tactic could still be executed without restarting the runtime environment. The heartbeat message, as described in \cref{sec:api}, contains two properties called ``fidelity level" and ``config hash". Both of these fields are utilized by our simulation environment to realize what kind of agent should be spawned, and the agent's onboard sensors. By adjusting the values in these fields from both the low level simulation and the high level simulation, we are able to make on the fly changes as to what level of agent is running.

\subsection{Cross Simulation Support}
\label{sec:cross-sim}
\begin{figure}
    \centering
    \includegraphics[width=.8\textwidth]{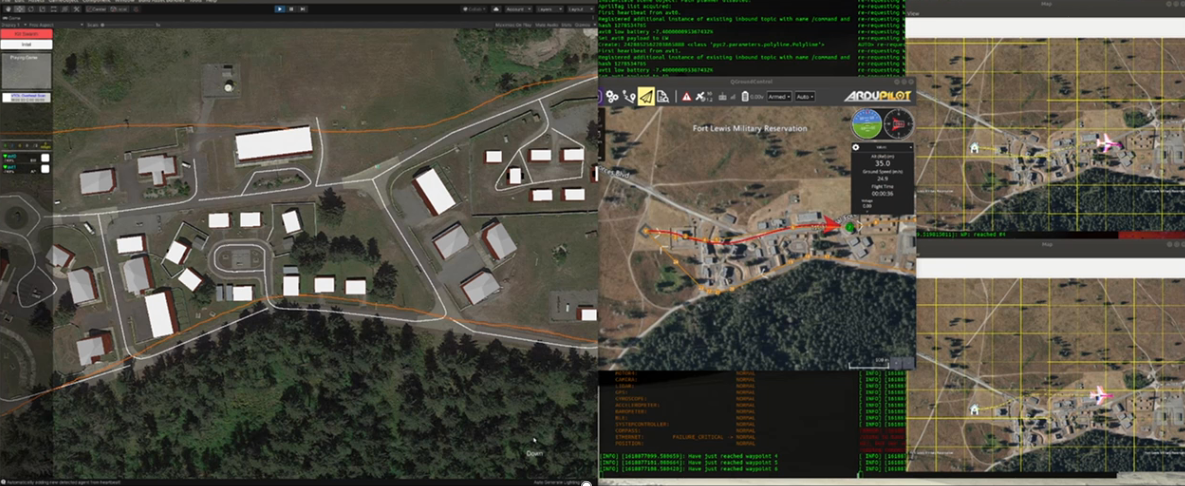}
    \captionof{figure}{Swarm Engine running alongside the Ardupilot simulator with two simulated AVTs}
    \label{fig:cross-sim}
\end{figure}

Over the course of our development, we realized that there are unique situations in which our simulation was missing a required feature we needed. Most of the time, this arises during the creation of very specific robotic algorithms. Whether we needed a detailed physics model, higher quality visuals, or a unique platform with accurate model description, these are gaps within our simulation. Thankfully, there are many other simulation environments that specialize in these areas. Due to the nature of our minimum API required for agent integration \cref{sec:api}, we are able to run Swarm Engine, simulated Docker instance(s), and a separate simulation at the same time. In this configuration, Swarm Engine interprets the simulated agent as a physical platform instead and does not generate any simulated sensor sources. Instead, the separate simulation instance is in charge of generating the simulated sensor sources via ROS messaging. The simulated Docker instance is expecting to receive ROS information, so as long as the simulator is configured with correct topic names and namespacing, we are able to perform this swap. Both Swarm Engine and the separate simulation environment must be able to localize within the same environment, however, most simulators have the ability to configure GPS information. We utilized this configuration to quickly spin up our recently integrated VTOL platforms, the AVTs. \cref{fig:cross-sim} showcases an example where we had two simulated AVTs being tasked by Swarm Engine but utilizing the ArduPilot simulation for sensory information. This allowed us to quickly develop and verify the AVT primitives and algorithms that we used on the real platforms without the need of supporting an additional vehicle type in our simulation environment. This type of simulated environment setup was also utilized with Gazebo during integration with the Hive (see \cref{sec:hive}).

\section{RISE Extensibility: Third Party Integration}
\label{sec:integrations}

In this section, we highlight various third party capabilities that were integrated into the RISE system as part of the DARPA OFFSET program. These third parties, known as ``sprinters," comprise small business and university research groups that proposed capabilities for specific thrust areas that were developed and integrated during a ``sprint." DARPA scheduled five sprints during the program's life that aligned with potential showcases of sprinter capabilities at the regular Field Experiments, or ``FX" (see \cref{sec:field_experimentation}). 
\subsection{Heron Distributed Task Assignment System}
\label{sec:heron_integration}
Heron Systems focused on developing an autonomous task assignment framework via decentralized asynchronous auctioneers to handle the distributed task assignment problem. This would function in place of the existing bidding task allocation system and provide an alternative that offered additional features and capabilities. We closely collaborated with Heron to integrate their capabilities into RISE and leveraged the ROS-based design to handle data passed between Heron's core modules and RISE. Heron's task assignment framework was fully integrated and was demonstrated at a field experiment in Ft. Benning, GA. This capability performed in place of the prior existing bidding system and allowed for a decentralized solution that allowed agents to take on tasks, even if they were not directly in communication with RISE C2 but were communicating with other agents. This provided the ability for tasking assurance and mission persistence, utilizing indirect communication between the intended swarm agent and C2. The modularity in the RISE design and its foundation in ROS allowed for a critical component in the overall swarm operation to be replaced with ease.

\subsection{MTRI-SoarTech Synthetic Scan and Dismount Detection}
\label{sec:mtri_integration}
Michigan Tech Research Institute (MTRI) participated in a number of sprints on the DARPA OFFSET program and were seasoned in integrating with RISE. In Sprint 4, MTRI focused on the virtual testbed thrust area to utilize a synthetic technology that may not exist in the real world but could be simulated in a virtual environment. MTRI, in collaboration with SoarTech, aimed to develop the Structure Situation Awareness for Swarms (SSAS) capability, which provided synthetic through-wall floor plan generation and dismount detection via software that was integrated with RISE. We worked with both MTRI and SoarTech to fully integrate SSAS and showcased this capability using physical platforms at the field experiment at Joint Base Lewis-McChord, WA. Using the RISE Swarm Engine C2, we were able to task UAVs to perform flight tactics around buildings and physically perform scanning operations. The physical scans employed the SSAS behaviors to perform a data collection on simulated info of those real physical buildings to then generate the through wall floor plan of that building and also detect simulated dismounted soldiers. The data product they produced included uncertainty information that would closely resemble real life data collects if this technology existed in reality. From the physical flight and the generated data product, MTRI was able to produce a visualization of that data product and show the generated floor plan and dismount detections performed. By integrating with RISE, a virtual synthetic technology and application was demonstrated using actual physical platforms. RISE was able to support the entire workflow without requiring MTRI and SoarTech to develop the whole end-to-end pipeline, allowing them to focus on developing out their specific capabilities while also being able to apply those capabilities in a larger robotics system. 

\subsection{SoarTech Interior Building Clearing}
\label{sec:soartech_integration}
SoarTech is a seasoned Sprinter that has participated in numerous sprints and has worked closely with us on integrating various capabilities into RISE. During Sprint 3, SoarTech's proposed capability was focused on the indoor environment and the aspect of searching and clearing an interior of a building. Their proposed approach would provide an increase in effectiveness and efficiency for clearing a building floor's interior by implementing a continuous and simultaneous search approach that leverages the multi-agent feature of the swarm. SoarTech developed this capability and wrapped it into a FlexBE state that was integrated into our existing building exploration tactic. As their focus was solely indoors, they leveraged RISE's existing outdoor navigation capabilities to initially maneuver to the buildings. SoarTech ran their own execution environment as a standalone servlet within the RISE agent Docker containers to perform their capability when it was executed in the FlexBE behavior. SoarTech utilized RISE's network to share generated building maps between agents. They maintained communication in large buildings by use of dynamically assigned ``relay'' agents automatically placed at strategic locations to ensure that data could be routed using RISE's MANET (see \cref{sec:manet}) out of the building and to the rest of the swarm. Through extensive collaborative development and iterative testing, we were able to conduct a full integration of their capability with RISE and our ground vehicles. We demonstrated the execution of their capabilities during FX-3 at Camp Shelby, MS. 

\subsection{HIVE}
\label{sec:hive}

\begin{figure}[t]
    \centering
    \includegraphics[width=.5\textwidth]{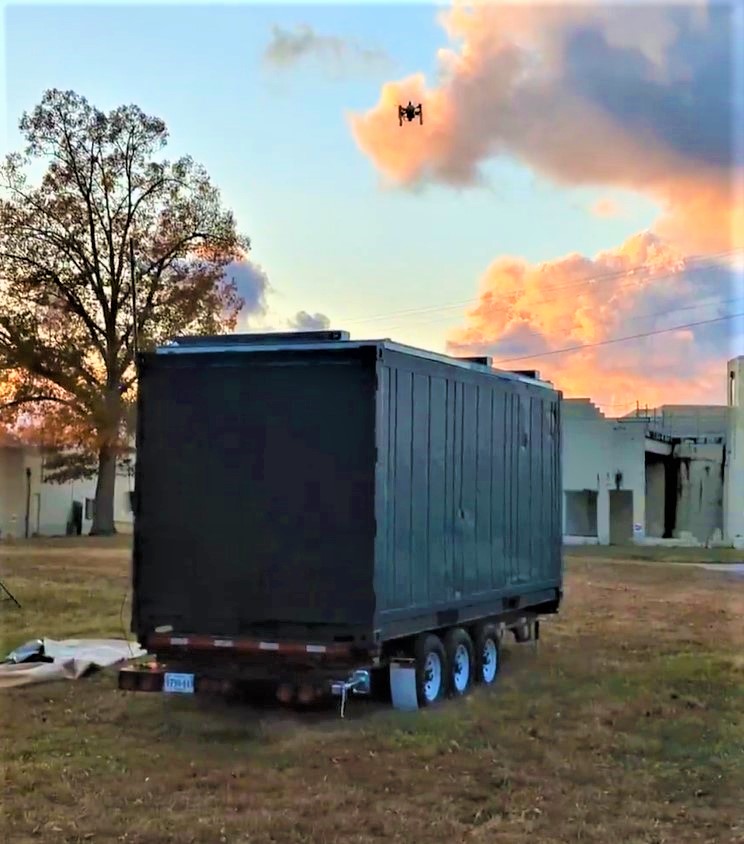}
    \captionof{figure}{RISE UAV agent landing onto HiveISO at Field Experiment 6}
    \label{fig:hive_ifo}
\end{figure}

For the hardware thrust area of Sprint 5, Sentien Robotics proposed an automated UAV ground management system that provided storage, charging capabilities, and launch/recovery logistics for a larger fleet of agents than their previously developed systems. This system is called Hive, and they developed two variations: HiveXL and HiveISO, which were essentially the same but were just aesthetically disparate due to their different manufacturing processes. Their system was fully housed in a trailer that contained storage bays for a large capacity of drones with a retrieval system that can transport an integrated UAV to and from its charging bay to one of two rooftop launch/landing pads. The system also consisted of maneuvering gantry arms for each launch/landing pad to manipulate the UAV to be in the necessary position for take off and landing retrieval. RISE and the Sentien team worked closely with each other to fully integrate a UAV operation workflow in which  RISE C2 could task agents in the Hive, and the Hive would retrieve those agents from their charging bays and bring them to the rooftop launch pad for takeoff. Upon completing their flight operations, the Hive UAVs would return to the Hive and perform a precision landing onto one of the two bays (See \cref{fig:hive_ifo}). Once landed, the Hive would reposition the UAV to the appropriate orientation and then place the UAV onto a charging tray through a bay door. The orientation is important as there is a charging mechanism that connects the drone to the tray for charging in its bay that does require a specific orientation for proper interfacing. Once the vehicle is on the tray, the interior retrieval system will return the vehicle to a charging bay, thus completing a full flight operations cycle. RISE and Sentien were able to demonstrate this full flight operations cycle at the final field experiment at Ft. Campbell, TN. The Hive does provide a significant logistical solution for transportation, charging, deployment, and retrieval of swarm agents but does encounter some limitations in supporting simultaneous launch of numerous platforms. It is better suited for persistent operations of lower numbers of UAV or more consecutive launches, as it would start to encounter congestion of the launch and landing pads. The Hive integration with RISE demonstrates the ability of RISE to incorporate other technologies aside from robotic platforms that support swarm operations, in this case a logistical technology. 

\subsection{Integration Summary}
\label{integration_summary}
The integrations of capabilities from Heron, MTRI, SoarTech, and Sentien Robotics serve as case studies representing diverse examples of some sprinter integration efforts from the DARPA OFFSET program. The various sprints with their different focus areas attracted a large range of capabilities. We were able to work with  various sprinters to establish integration paths for their capabilities into RISE and provide a swarming robotics framework to realize the sprinters' capabilities, which ranged from swarm autonomy, swarm tactics, robotics hardware, human-machine interfaces, AI, and more.

\section{Field Experimentation}
\label{sec:field_experimentation}

As part of the DARPA OFFSET program, field experiments were conducted to test and demonstrate the program's technological progress. These were conducted approximately every six months and were orchestrated by an experimentation team from Naval Information Warfare Center Pacific (NIWC). In total, RISE was tested iteratively at five urban training ranges over four years. \cref{table:FXevents} summarizes all OFFSET field experiments. Whereas FX-0 was an indoor only event, the remaining field experiments took place outside. FX-1---FX-3 gave us the opportunity to test our system in a representative environment, and lessons learned from these experiments informed the design of our architecture as well as our user interface. FX-3 gave us additional insights into scalability issues stemming from our original network solution (see section \cref{sec:netprototypes}), allowing us to achieve greater scale and operational capabilities in subsequent experiments.
Upon reaching FX-4 and FX-6, we had gained the relevant experience and information to consistently field large scale swarm deployments based on mission level objectives.

\newcommand{\ra}[1]{\renewcommand{\arraystretch}{#1}}
\begin{table}
\ra{2.0}
\small
\begin{tabularx}{\linewidth}{ lll X } 
\toprule
\textbf{Experiment} & \textbf{Date} & \textbf{Location} & \textbf{RISE Objectives} \\ 
\midrule
\textbf{FX-0} & March 2018 & FDNY, NYC & Test of robotics codebase and architecture. Also, rehearsal for field robotics. \\
\textbf{FX-1} & October 2018 & Camp Roberts, CA & Implementation of multi-agent framework and command and control interface \\
\textbf{FX-2} & June 2019 & Ft. Benning, GA & Swarming operations for primitive environmental Intelligence, Surveillance, and Reconnaissance (ISR) using next C2 iteration, mobile commanding with tablet C2 modality, and a Virtual Reality (VR) interface\\
\textbf{FX-3} & December 2019 & Camp Shelby, MS & Large scale swarm networking and operations with the next-generation gesture-based iteration of the C2\\
\textbf{FX-4} & June 2020 & Joint Base Lewis-McChord, WA & Enhanced swarm tactic execution and improved C2 with Augmented Reality (AR) modality\\
\textbf{FX-5} & Cancelled & N/A & N/A\\
\textbf{FX-6} & November 2021 & Ft. Campbell, TN & Full scale swarm testing with latest evolution of C2 to include Live, Virtual and Constructive feature and improved AR interface to include commanding \\
\bottomrule
\end{tabularx}
\caption{RISE has been tested at one indoor and five outdoor field experiments over the last four years. Note, FX-5 was cancelled due to COVID-19.}
\label{table:FXevents}
\end{table}

\begin{figure}
    \centering
    \includegraphics[width=\textwidth]{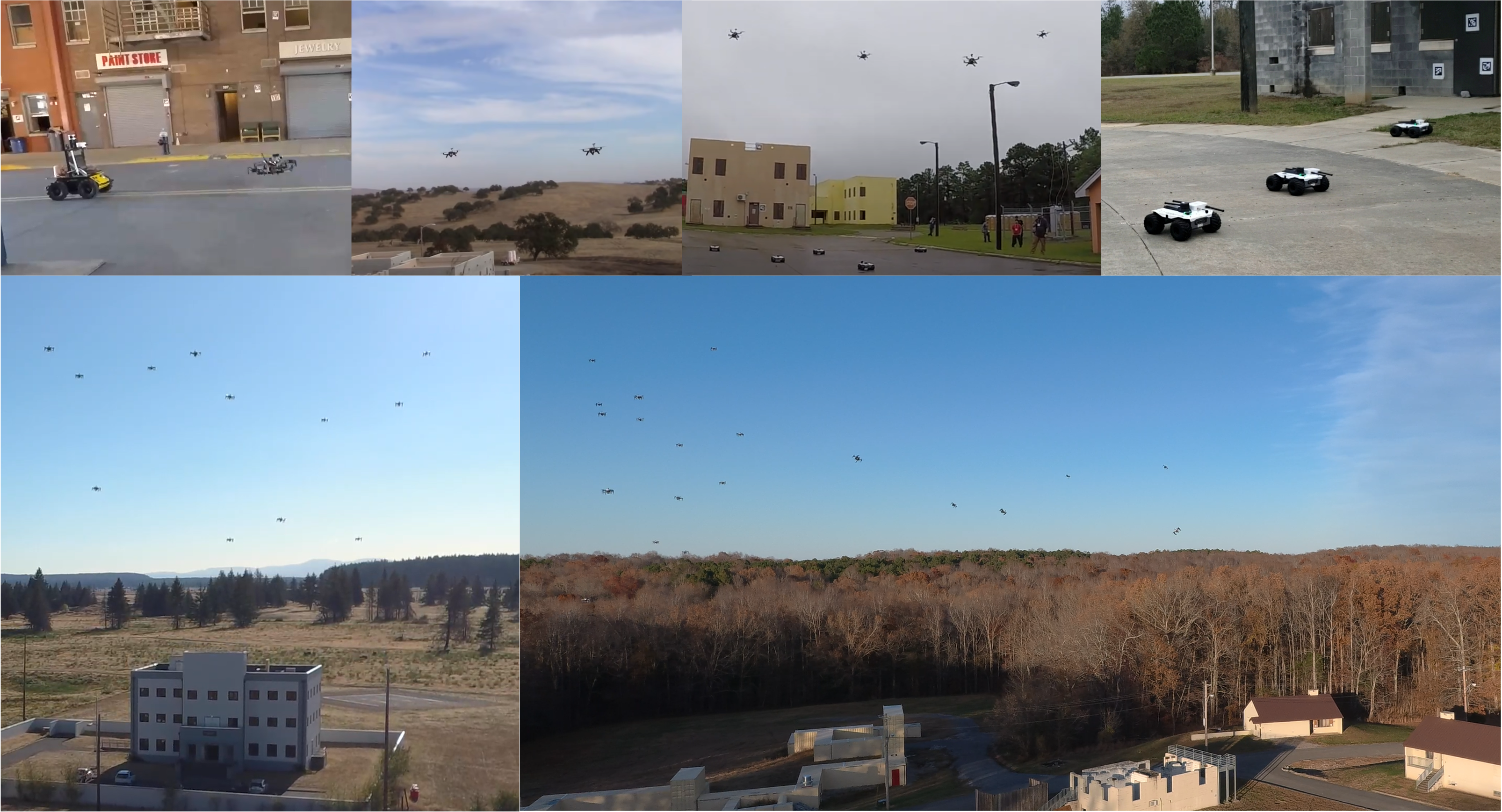}
        \captionof{figure}{The six field experiment locations coorespoding with \cref{table:FXevents}. Top: FX-0 (Left)---FX-3. Bottom: FX-4 and FX-6.}
    \label{fig:collage}
    ~\\~\\
    \centering
    \includegraphics[width=\textwidth]{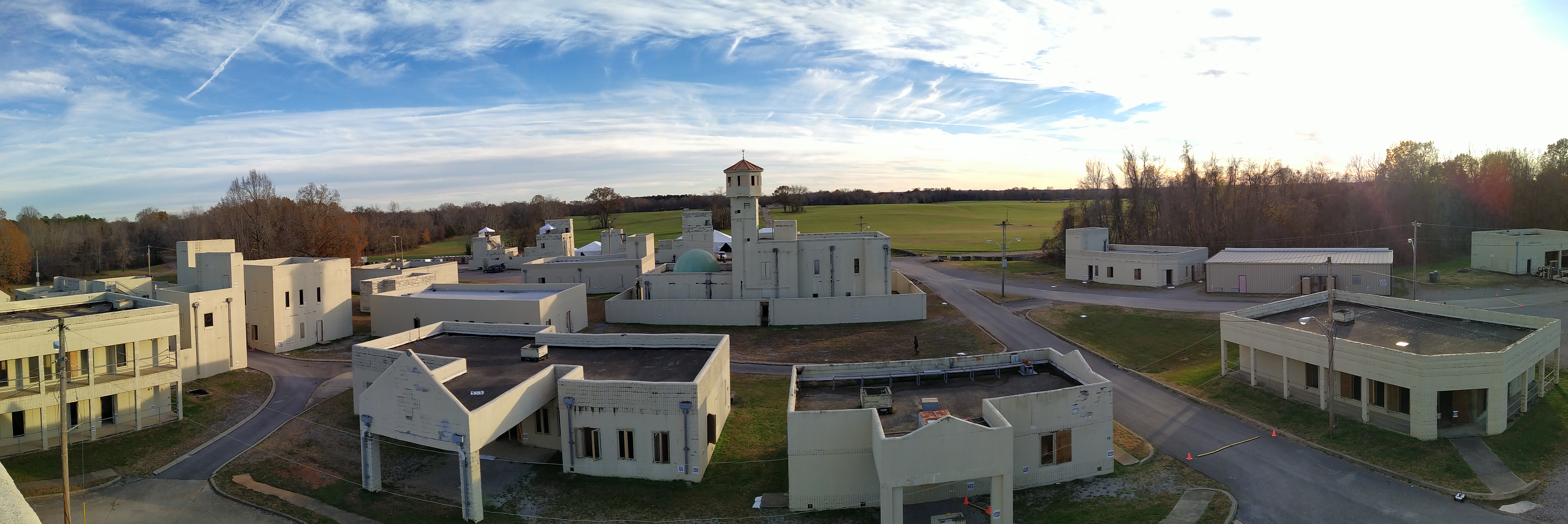}
    \captionof{figure}{Rooftop view of Cassidy CACTF in Ft. Campbell, TN.}
    \label{fig:fx6_cassidyCACTF}
\end{figure}

In the remainder of this section, we discuss the penultimate field experiment, FX-6. This experiment enables us to answer the question: can a multi-robot ecosystem designed for rapid integration and one-to-many control can be fielded, where a single operator commands 150+ autonomous vehicles in tactical maneuvers?

\subsection{Participants}
The Northrop Grumman RISE team filled all primary roles including that of one swarm commander, one swarm operator, one health engineer, and one field operations officer. These individuals were all RISE experts with advanced knowledge of the C2 interface and platforms used in the experiment. Further, all except for the health engineer had participated in prior field experiments. The core team was supported by 8--9 field support personnel who assisted with logistics, deployment, and safety. Finally, a Rajant field engineer participating as an auxiliary team member assisted with network infrastructure. 

 \subsection{Facilities and Equipment}

\subsubsection{Environment}
FX-6 was conducted at Fort Campbell in Tennessee on the Cassidy CACTF, see \cref{fig:fx6_cassidyCACTF}. Cassidy is approximately \SI{305}{\metre} in diameter and comprises an urban road network connecting approximately forty-five buildings, including a church, embassy, town homes, and mosque, among others. In addition to being densely packed, power lines, curbs, trees, and other obstacles permeate the CACTF. 

NIWC augmented the CACTF to support experimentation by distributing scenario-based artifacts throughout the environment. These artifacts are AprilTags \cite{wang2016iros} (a type of fiducial marker) that represent complex objects. For example, AprilTags are used to represent high value targets, hostiles, improvised explosive devices (IEDs), medics, benign objects, building labels, and priority intelligence information, among others. Certain AprilTags are coupled with field nodes that house a Raspberry Pi and Bluetooth Low Energy (BLE) device that enable interactions between robots and artifacts, e.g., an IED can disable a robot via a Bluetooth interaction when the robot comes within range of the IED. The specific AprilTag family used for this experimentation was 48h12. This tag family was used for its nested tag capabilities. The outer tag could be detected by agents from a further distance and typically only contained generalized information such as person or object. Only the inner tag revealed the true nature of a particular fiducial, revealing whether it was benign, hostile, or friendly. Within RISE, most of the inner versus outer tag interaction was handled autonomously by onboard agent logic (see \cref{sec:rise_primitives} ). In total, over \SI{8000}{\metre} of Ethernet cables, \SI{300}{\metre} of optical fiber, 1901 unique AprilTags (26 dynamic), 136 Raspberry Pi nodes, and 46 network switches were employed to generate the test scenario.

\subsubsection{RISE Swarm}
\label{sec:fx6swarm}

The mission scenario and the venue selected aimed at necessitating a larger swarm from one field experiment to the next. Therefore, by FX-6, we not only required a large swarm capable of sustaining a 3.5 hour mission, but one with diverse capabilities to meet varying mission objectives. With this in mind, we brought a large heterogeneous swarm of 274 platforms as shown in \cref{fig:robots}. This composition enabled us to support three operational levels: ground level with ATX platforms, low to mid-altitude air with IFO platforms, and high altitude overhead with AVT platforms. Additional information about each platform type follows:

\begin{figure}[htbp]
    \centering
    \includegraphics[width=.7\textwidth]{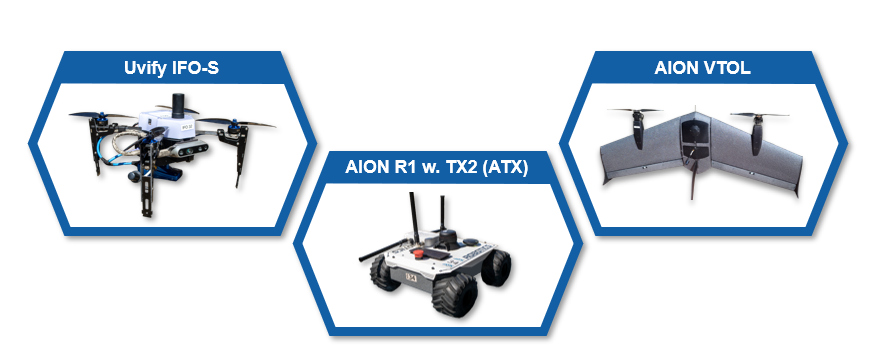}
    \captionof{figure}{The 274 robotics platforms utilized in FX-6 from left to right: 158 Uvify IFO-S (IFO), 97 Aion R1 (ATX), and 19 Aion (AVT) vertical take-off and landing (VTOL) platforms.}
    \label{fig:robots}
\end{figure}

\begin{itemize}
    \item \textbf{Ground}: Ground vehicles provide long mission endurance and are more power-efficient than their aerial counterparts. They can support more payloads and additional hardware without major impacts to their overall performance. They provide a ground-level operational advantage with decreased risk of damage during operations. They can better withstand collisions and/or harsher operations in comparison with aerial platforms. These platforms also have an additional sensor which is an omnidirectional range scanning lidar (RPLidar A2) which was used for mapping and localization. The lidar along with wheel encoders allow these platforms to also perform in GPS denied or restricted environments unlike the other platforms. However, they have limited traverse-ability and must avoid more obstacles than aerial vehicles, which increases the time it takes for them to reach their destination. We initially used Kobuki Turtlebots and custom-built skid-steer rovers. However, to support larger operations, we integrated the AION R1 rover with our system, where it became the primary ground asset. The R1 rovers use Jetson TX2 companion boards and are referred to as ``ATX" (see \cref{fig:robots}).

    \item \textbf{Quadcopter}: Quadcopters provide more targeted control of aerial vehicles and do not require lateral movement to maintain flight. They are highly maneuverable and can hover as needed, allowing them to operate in lower altitudes where there are more obstacles. Quadcopters are capable of traversing the environment in three dimensions, with an ability to go over as well as under various obstacles, and can navigate to goal positions faster than ground vehicles. However, there are limitations to the platform, such as the importance of its power:weight ratio to achieve and sustain lift. Flight time and efficiency are influenced by vehicle weight and payload. There is also an increased risk of damage, especially when flying in dense environments. We previously integrated custom-built platforms, as well as the Intel Aero. To support the OFFSET program and its scope, we integrated and heavily utilized the Uvify IFO-S platform, referred to as the ``IFO" (See \cref{fig:robots}).
        
    \item \textbf{Fixed Wing}: Additionally, we integrated the AION VTOL Tailsitter (AVT) shown in \cref{fig:robots}. While this vehicle is more energy-efficient than quadcopters and thus may remain airborne for longer periods, it must also maintain a minimum airspeed. VTOL type fixed wing aircraft remove the need for launching mechanisms as well as long take-off runways, and offer the hovering capability of a quadcopter, but are more susceptible to wind. Utilization of non-VTOL fixed wing platforms with swarming has led to constraints on the number of platforms one can launch simultaneously \cite{chung_live-fly_2016}. However, while VTOL platforms have limited rotational agility, requiring large open areas to turn, this is countered by their increased flight time as opposed to quadcopters. Therefore, VTOLs are well suited for high altitude operations such as overhead surveillance.

\end{itemize}

Every platform further carried a Bluetooth emitter that was used to simulate a payload. Potential payloads were electronic warfare (EW), anti-personnel (AP), acoustic spoofing (AS), and covering fire (CF). These payload types affected agent capabilities, what field node artifacts they could interact with, and impacted how they were deployed in a mission. 

Given the mission duration, maintaining and updating a large swarm is an intensive effort. Aside from the logistics of maintaining the hardware and performance aspect of the agents, it also requires the ability to update all agents to the latest software before execution of their missions. RISE's architectural design allows for the utilization of Ansible to effectively and efficiently deploy software updates to the swarm. A variety of playbooks were developed for deployments based on update requirements. These updates can be as discrete as deploying an updated Docker container, robot extensions codebase, or other components of the platform codebase. 

\subsubsection{Command and Control}

\begin{figure}[htbp]
    \centering
    \includegraphics[width=\textwidth]{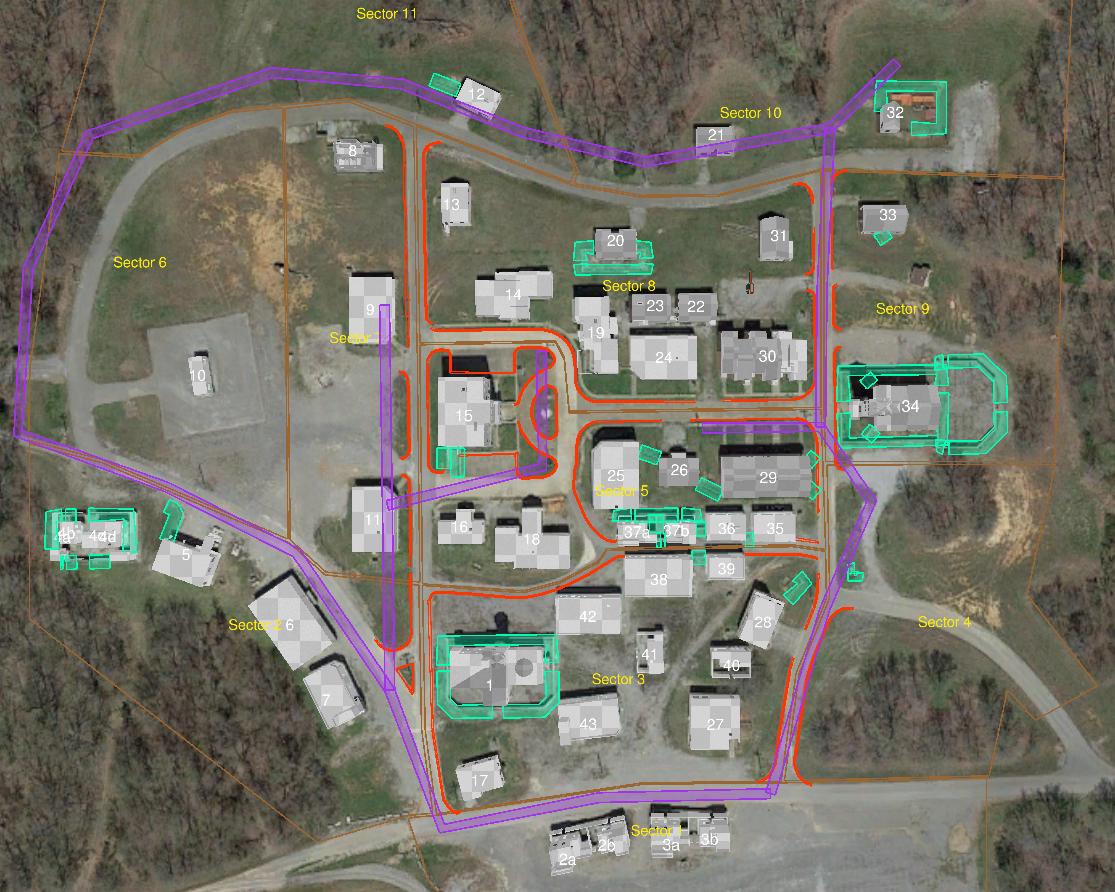}~
    \captionof{figure}{Fort Campbell Cassidy CACTF augmented with a-priori site intelligence provided by NIWC, including curb (red), wall (green), and powerline (purple) obstacle data as well as sector boundary information. Buildings were modeled prior to FX-6 from floor plan data site visit information, whereas obstacle data was loaded at the start of each mission run. Map is oriented so that north is up.}
    \label{fig:fx6-site_intel}
\end{figure}

The swarm operator ran RISE C2 software on an Alienware laptop comprising a \SI{2.20}{\giga\hertz} Intel\copyright~Core\texttrademark{} i7-8750H CPU with \SI{32}{\giga\byte} of \SI{2667}{\mega\hertz} DDR4 memory and an NVIDIA\textsuperscript{\textregistered}~ GeFore\textsuperscript{\textregistered}~ RTX 2080 mobile graphics processor. A mouse was used for all swarm command and control, while keyboard arrow keys were used to toggle between visualization modes. The swarm health engineer and swarm commander used a similar system, though the commander only used C2 for situational awareness. The commander, operator, and engineer operated from the roof of building 3b, shown at the south side (bottom) of \cref{fig:fx6-site_intel}.

\subsection{Protocol}

\subsubsection{Scenario}
\label{sec:FX_scenario}

Throughout the OFFSET program, NIWC developed mission scenarios focused on swarm operations in an urban environment that iteratively evolved since the first field experiment. The mission story typically centers around an oppositional force (OPFOR) having established a hold on an urban environment, with primary and secondary High Value Targets (HVTs) spread throughout and protected by various defensive measures. The objective is then for the blue force swarm (BLUFOR) to enter the environment, overcome the defensive measures, and secure the HVTs. For FX-6, the mission statement given in a scenario and experiment guide distributed by NIWC states: 
\begin{displayquote}
DARPA HQ has received intelligence pointing to the development of a weaponized contagion that is located in the buildings within the Fort Campbell Cassidy CACTF compound. Blue forces have established a foothold on the south end (Sector 1) of the CACTF. The primary assault will take place starting from the south end (Sector 1).
\end{displayquote}

Further, the mission is divided into four phases as follows:
\begin{itemize}
\item Phase 0: Swarm Deployment and Intelligence, Surveillance and Reconnaissance (ISR) 
\item Phase 1: Combat Action and Isolation of Primary Target Objectives 
\item Phase 2: Conduct an Urban Raid
\item Phase 3: Seize Key Urban Terrain
\end{itemize}

The objective of phase 0 is to deploy the swarm, gather intelligence information about the environment, maintain surveillance in the area of operation, and maintain situational awareness. Localization and sensor errors prohibited our ground vehicles from autonomously moving with the precision necessary to enter buildings, and for this reason, we primarily participated only in phase 0, as later phases required breaching capabilities. 

\subsubsection{Mission Preparation}

The BLUFOR team was given approximately one hour to prepare for each mission run. During this time, the field operations personnel situated the swarm in their initial deployment zones according to mission requirements. Since assaults came from sector one per the established scenario, IFO and ATX platforms were distributed along the south road, using all available space. For safety, AVT fixed wing platforms were set in an open area east of building 3b, away from most personnel. During this time, the swarm commander worked with the swarm operator to develop a course of action based on mission requirements and platform limitations. Once the plan was finalized and platforms placed, systems were powered on for a communications check. Network issues, when they occurred, were resolved and the swarm was powered off. Thereafter, the swarm commander briefed the experiment team, event organizers, and VIP guests on the mission plan and expected outcomes while the swarm operator separately input the final initial course of action in preparation for the start of the mission. 

\subsubsection{Mission Execution}

After the mission briefing, field operations personnel returned to the deployment zone or took safety spotting positions throughout the CACTF. Those robots required to execute the initial COA were powered on, and once a sufficient percentage of the swarm was operational, the swarm commander began the assault. Throughout a mission run, the swarm commander would monitor the situation, react to intelligence information, and decide on subsequent courses of action, which the swarm operator effectuated through C2. The mission continued until the swarm commander called mission end or the allotted 3.5 hours were exhausted.

\subsection{Results}
\label{sec:fx6results}

\begin{figure}[htbp]
    \centering
    \includegraphics[width=\textwidth]{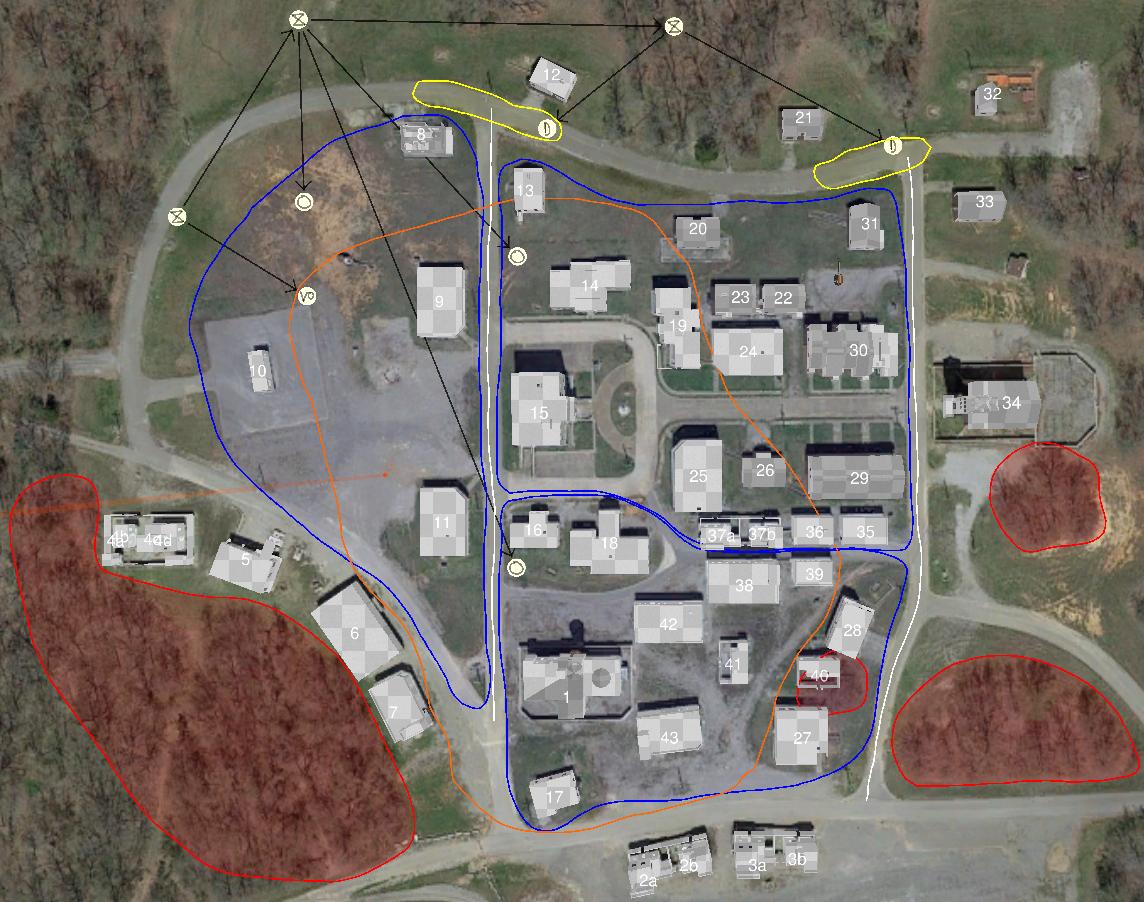}~
    \captionof{figure}{Example starting course of action used at FX-6 with obstacles (curbs, walls, and powerlines) removed for clarity. Details are discussed in \cref{sec:fx6results}.}
    \label{fig:fx6-mission_plan}
\end{figure}

With the field experiments scenario infrastructure in place, RISE was able to field its capabilities and technological innovations in a manner that provided insight into robotic swarm operations. 

\subsubsection{Command and Control}
\cref{fig:fx6-mission_plan} illustrates one example course of action used at FX-6 used to start a mission run. Since the Cassidy CACTF possessed trees not included in our virtual environment representation, the swarm operator manually setup no-go zones to ensure PyC2's path planner routed around these obstacles. The swarm operator further used a tactic chain to sequence three events. First, a timer (far left hour glass symbol) with zero delay initiates a VTOL scan using three AVTs over the orange sketch area. Next, after a one-minute delay, a second timer initiates three overhead scans, one per blue explore area, which utilizes the IFO quadcopter assets. Finally, after another minute passes, the third timer initiates two deploy tactics intended to move ATX ground agents into the northern yellow deploy zones. Since each deploy zone is connected to a route (white south-to-north sketch lines), agents will traverse the specified path rather than use a possibly shorter path through the CACTF. A recovery point (orange point with beacon near building 11) was also set, identifying where IFO agents should fly to and land when their batteries are low. With this approach, the operator was able to deploy all available assets and achieve high coverage using only 14 tactic sketch parameters (tactical control measures) and 9 tactics. 

As the mission progressed, the swarm operator was typically directed by the commander to continue scan operations, including overhead and building scans depending on the situation. Attempts were made to engage hostiles and breach buildings with ground agents, though localization and sensor errors often prohibited these advancements. 

In total, 13 unique tactics were used throughout the entire event, two of which employed virtual hardware sensors. These tactics being operationalized with sketch-based tactic parameters provided enough flexibility to serve our needs given our platforms' autonomy limitations. With greater autonomy in obstacle avoidance and navigation, more tactics would have been required as we would have put the swarm to other uses, especially in coordinating assaults and securing buildings.

\subsubsection{Operational Sustainment}
At FX-6 , RISE was able to maintain operations throughout the allotted 3.5 hours per trial run, with the ability to extend beyond that. Given the nature of field experimentation with low-cost robotics in a dense urban environment, attrition such as sensor and system malfunctions, abrupt degradation in flight conditions, prolonged platform life cycle, and other physical inhibitors is to be expected. Yet because of operational prudence and our approach to tactics design, the swarm of 274 assets brought to FX-6 were appropriately provisioned for the two-week event, and despite occasional vehicle collisions with the environment, onsite repairs and revivals of assets allowed for not only swarm sustainment but growth. With each exercise shift, RISE provisioned a greater percentage of its assets toward the mission run such that by the end of the event, RISE was able to utilize 174 assets simultaneously with additional assets on standby to ensure mission sustainment. A key question of the program was if a single operator could control 150+ drones, and this final result proved that it is possible. 
\subsubsection{Mission Coverage}
\begin{figure}[t]
    \centering
    \includegraphics[width=1\textwidth]{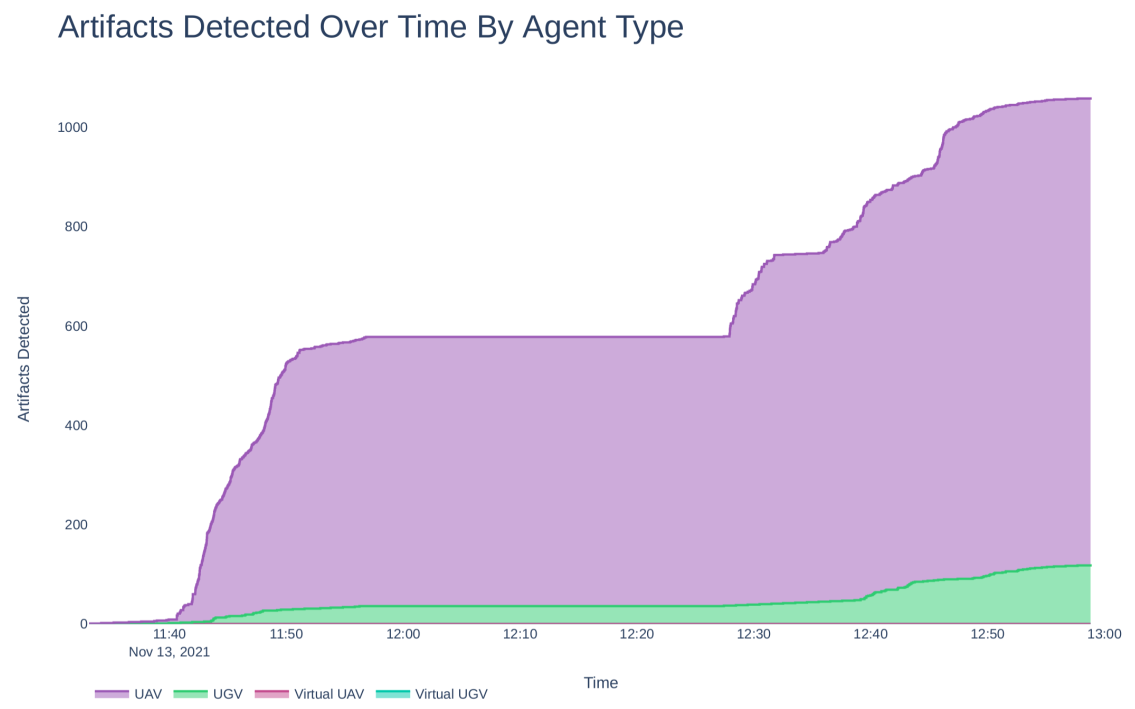}
    \captionof{figure}{Unique artifact detection events recorded by NIWC's orchestration software over the first part of 100 agent run at FX-6.}
    \label{fig:detections}
\end{figure}

\begin{figure}
    \centering
    \includegraphics[width=.73\textwidth]{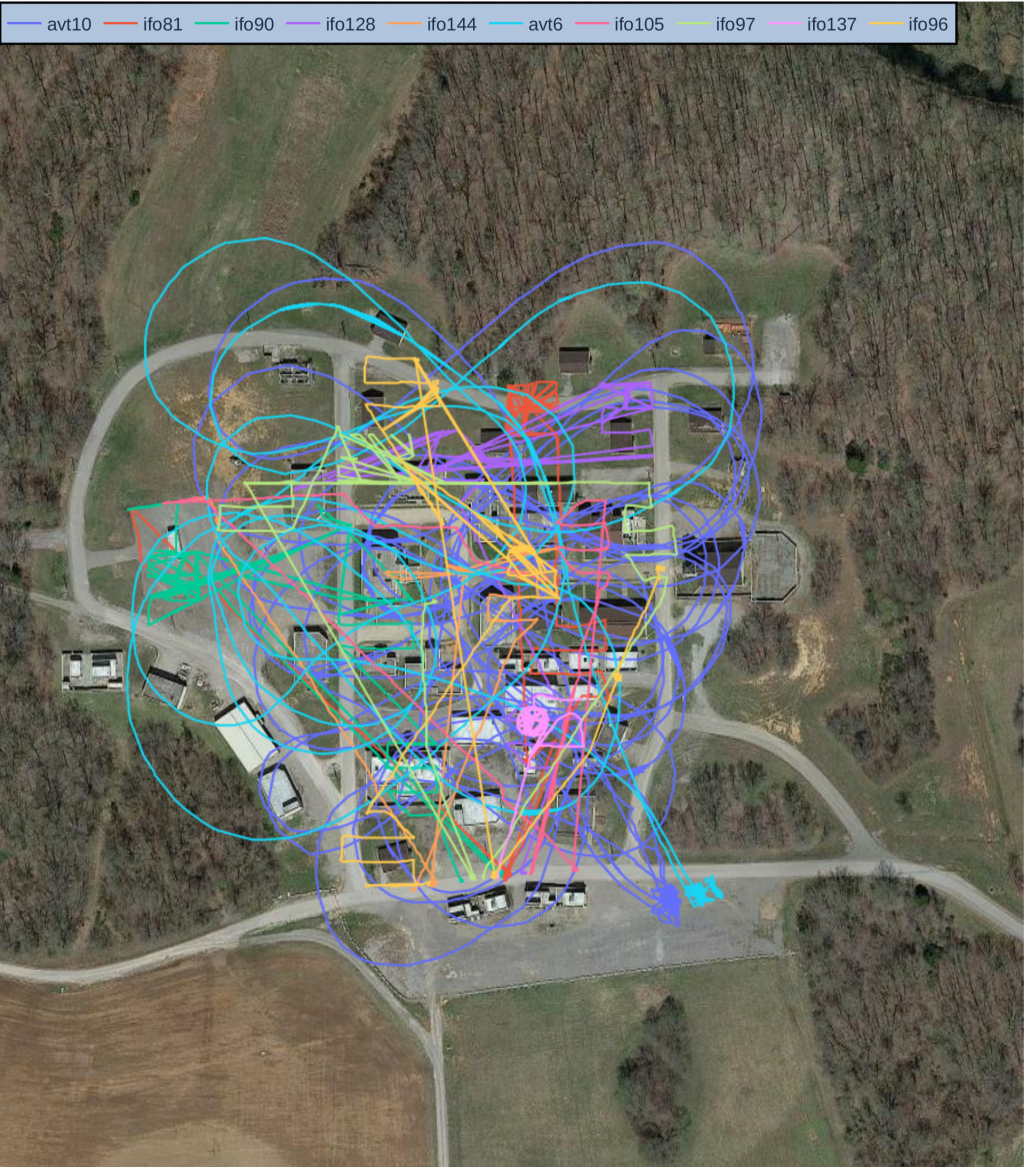}~\\~\\
    \includegraphics[width=.73\textwidth]{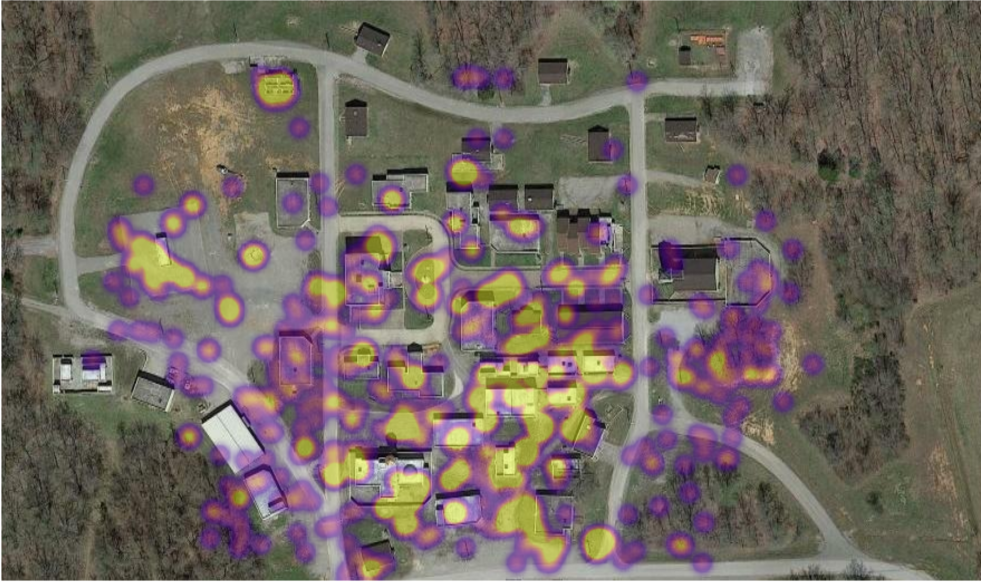}~
    \captionof{figure}{Agent trajectories (top) and coverage (bottom) recorded by NIWC's orchestration software over the course of run at Fort Campbell during FX-6.}
    \label{fig:heatmap}
\end{figure}

\cref{fig:detections} shows the detections for approximately the first hour and a half of a run using 100 agents. Within approximately the first 20 minutes, the swarm was able to detect 600 artifacts and as the mission went on, even more were detected. \cref{fig:heatmap} shows the mission area coverage over the course of one run, where the top image illustrates the flight paths of a subset of ten swarm agents and the bottom image visualizes swarm activity across the environment. These results highlight not only the full area coverage provided by the swarm, but also that the swarm activities coincide with the building density of the CACTF. Though these charts and images are only a snapshot of a single run of the RISE swarm during FX-6, they are representative of the consistent ability of RISE to deploy its swarm to cover the entire operational area and perform its mission.  

\section{Discussion, Lessons Learned, and Future Work}

With RISE we aimed to demonstrate that a system designed for rapid integration of new technologies and ideas can be used by a single operator to control a large heterogeneous swarm of autonomous agents. To this end, we designed an ecosystem informed by our experiences as an integrator who worked with third parties focused on key features rather than holistic systems. Distilled down to the set of requirements defined in \cref{sec:requirements}, these experiences combined with OFFSET program requirements led us to develop a solution that separates concerns into four core modules---those pertaining to the user interface, tactics, behaviors, and algorithms. In particular, the separation of tactics into a module that facilities coordination between agents at a level of abstraction above the agent, in combination with a sketch-based interface where operations are described in a sketch language were key in enabling rapid integration and one-to-many swarm control. With PyC2, tactics development time dropped from weeks down to hours. PyC2 also enabled tactics developers to focus on how an operator will operationalize a tactic through the user interface (without being a UI developer), which can then be put into relation with how individual robots are coordinated to serve a specific tactical objective. Once all modules were in place, we were able to quickly write tactics that enabled an operator to input a single command that invoked numerous agents, and the operator did not need to concern their self with individual robot operations, owing in part to behavior level autonomy. As such, RISE successfully enabled a single operator to simultaneously control 174 platforms in the real world. This may effectively increase a Soldier’s span of control in the future to 1 Soldier to 150+ platforms, in contrast to the current paradigm today of 1:1. Swarm control for operators is enabled through our C2 interface and tactics development tools.

In contrast to most prior work in swarming, RISE has been operated on hardware in real environments since the very beginning. We have fielded and tested at 5 military ranges, with more field work planned in the future. Field robotics at the scale of swarming brings many challenges, but swarm logistics is also a ripe area for new research.

In this discussion, we highlight the topics of Swarm Engine C2, tactics development, and field robotics, offering lessons learned and future work for each topic.

\subsection{Swarm C2}
We designed Swarm Engine C2 so that a single operator would be able to command a large scale swarm. We demonstrated through field experimentation that it was possible for an operator to sustain continuous operations over an extended period of time, while maintaining sufficient situation awareness. Difficulties that arose during testing were primarily due to hardware errors and insufficient automation. In these situations, the operator was unable to generate new swarm level work and agents were left idle. These circumstances were due to the nature of the program and experimentation, as they do not reflect normal operations as they would occur in practice. Robot swarm hardware improvements (rather than our RISE platform) in combination with new tactics already under development will overcome these issues, thereby allowing the operator to work as intended. Once these tactics are available, a full simulation-based user study will be warranted.

Our user interface was designed to be customizable so that operators could select memorable gestures based on their personal experiences and associations. We found, however, that our team used those gestures that were defined by the engineer who implemented the associated feature. Because it is known that individually defined developer gestures are less favorable than those proposed by large groups \cite{morris2010understanding}, there were sometimes memorability issues. That is, as expected, individuals are able to learn our interface commands, but the gesture set was not optimal. And although customization enables rapid prototype in the short term, over a longer period of time, as swarm tactics mature and become standardized, we anticipate a need for periodic elicitation studies and agreement analysis \cite{vatavu2015formalizing}. As a result, swarm sketch grammars will become codified in the same way that mathematical, circuit diagram, and military iconography have all been standardized.

\subsection{Tactics Development Through PyC2}
PyC2 was a late development that came about during the second phase of the OFFSET program. Prior to PyC2, tactics were written partly in C2 and partly in robotics code, which proved to be cumbersome, time-consuming, and limiting. Thus, PyC2 was born out of a need to accelerate tactics integration and enable non-roboticists with the ability to develop new tactics (see Requirement \ref{req:tacticsAbstraction}). Once in place, our team rapidly developed and iterated numerous tactics, including those that led to a substantial performance improvement between the third and fourth field experiments. Being able to write new tactics without having to modify C2 and with the aid of sketches and context proved to be a valuable tool. Tactics development only slowed down due to platform limitation issues and the need for our team to turn its attention toward other priorities. With PyC2, we also saw an increase in sprinter integrations, as outlined in \cref{sec:integrations}.

One key issue we plan to address near term is that PyC2 is a fracture critical component. This means that if PyC2 fails or is unable to communicate with an agent, we are unable to command the swarm. Our solution is to enable PyC2 to run directly on robot platforms with an improved bidding protocol that better enables collaborative tasking. In this way, PyC2 tactics, integration, and development are unaffected, yet PyC2 becomes a distributed system. 

\subsection{Field Robotics}

RISE is developed with a bottom-up approach, so all overall tactic performance depends upon agent abilities to reliably complete tasks. When agents are unable to perform tasks reliably, this leads to a cascade problem of additional agents having to step in to take the tasks that failed or forces the swarm commander to re-think their approach to a mission. Most task failures were caused by individual agents' inability to accurately localize and navigate through the environment. Navigation and localization is an ongoing research area of robotics, especially in urban environments. Pushing forward in these areas will further increase the swarm's effectiveness and allow for even more complex agent behavior in the future.

With the scale of the RISE swarm, logistics and overall platform maintenance (both hardware and software) is a large problem. For example, over time, damage occurs to agents' sensors or motors and requires troubleshooting and/or repair. From a software perspective, every platform is running a unique release of its given codebase as well as particular firmware for its flight controller. Both of these systems needed to be up-to-date on all platforms in use to ensure consistent swarm operations between platforms. Also, the platform's FCU (e.g., the Pixhawk) requires calibration, and though this is not necessarily an intensive process, it becomes a high level of effort when it comes to calibrating the numerous platforms in the swarm. Additionally, even though on paper all the platforms subgroups were identical, we noted a lack of consistency from platform to platform, which added complications to platform development. By utilizing Ansible and Docker, we were able to create manageable platform update procedures for agent codebases and could deploy to ~50 assets in ~5-15min, depending on the amount of update required. Firmware updates and calibration were still an ongoing pain-point over the course of the program. Future work would potentially involve either more direct involvement in firmware codebases, removing our dependency on these subcomponents or working more closely with manufacturers to help manage these aspects.

By leveraging FlexBE, we established a strong foundation given our system's bottom-up approach to tactic development. The reusability of state development and simple drag and drop primitive creation based on underlying algorithms or states allowed for rapid creation of new agent behaviors. In the end, however, it seems as though there has been a large push beyond hierarchical state machines and into behavior trees. While FlexBE could theoretically support this capability, it does not currently exist within the system. As our system and architecture has evolved over the last 4 years, we discovered that FlexBE and PyC2 have potential to be duplicative in the future. For example, it is still an open question as to whether certain logic flow should exist within the agent codebases in primitives or at the tactic level within PyC2.

Utilizing ROS was an absolute necessity when designing RISE, and it would be our team's choice again. We do however wish that the ROS 2 transition occurred earlier. We were often a little too early to adopt each new ROS 2 feature. We believe that as ROS 2 matures, the future decentralized concept of no ROS master will be a powerful tool in future swarm algorithm development and swarm development in general.

\section{Conclusion}

Our Rapid Integration Swarming Ecosystem (RISE) provides a platform for future multi-agent research and deployment. Using both physical and simulated swarms, we demonstrated RISE's rapid integration of third-party swarm tactics and behaviors. Our physical testbed, composed of more than 250 networked heterogeneous agents, has been extensively tested in mock warfare scenarios at five CACTF sites. With our live, virtual, constructive (LVC) simulation capabilities, RISE allows the use of both virtual and physical agents simultaneously. Other simulation advances such as our ``fluid fidelity'' concept and super real-time simulation enable rapid swarm tactic prototyping. Our gesture-based interface enables an operator ratio greater than 1:150, making RISE a massive force multiplier. Together, these feature allow RISE to translate mission needs to robot actuation and swarm operation with unprecedented ease and flexibility.

\section*{Acknowledgments}

This work was supported in part by DARPA under Contract N66001-17-C-4066. We thank OFFSET program manager Dr. Timothy Chung for his expert oversight and insightful feedback over the program's four years as well as our collaborators at Northrop Grumman, University of Central Florida, and Intelligent Automation, Inc. We also want to thank the global ROS community and all open source robotic developers.

\bibliographystyle{apacite}
\bibliography{refs, refs_goeckner}
\begin{appendices}
\crefalias{section}{appendix}
\section{Full Tactic Example}
\label{app:example_tactic}
\cref{listing:full_tactic} presents the full Hold Position tactic. An operator uses this tactic to move a number of agents onto a sketched path surrounding an object or structure of interest, with all agents facing inward. Detailed comments are provided for comprehension.

\begin{figure}[htbp]
\input{listings/tactic_example.py}
\captionof{lstlisting}{Full tactic implemented in PyC2.}
\label{listing:full_tactic}
\end{figure}

\section{Network Parameters}
\label{app:network}
We changed several PGM parameters from their default values to allow for more efficient use of our limited network resources. First, we reduced the ambient SPM rate to $\frac{1}{30}$ Hz, and only transmit a single end-of-transmission ``heartbeat'' SPM 30 seconds after the last data transmission. Most importantly, we increased the NAK back-off interval to two seconds, the NAK repeat interval to 10 seconds, and reduced the allowed NAK retry attempts to 3. We further increased the NAK repair data interval to 800 ms and set a maximum repair data rate of 1 KB/s. Together, these adjustments prevent flooding and network breakdown in the case of serious communications disturbances, while still allowing for an acceptable measure of reliability.

\end{appendices}

\end{document}